\documentclass[lettersize,journal,doublecolumn]{IEEEtran}
\usepackage{PerlazaTIT2023}
\begin{document}
 
%\title{The Dependence of Generalization on Training Datasets in Statistical Machine Learning}
\title{The Method of Gaps: \\Exact Expressions for the Generalization Error of Supervised Learning Algorithms}
 
\author{Samir M. Perlaza{\orcidlink{0000-0002-1887-9215}} and Xinying~Zou{\orcidlink{0009-0001-9952-1783}}
\thanks{
Samir M. Perlaza and Xinying Zou are with INRIA, Centre Inria d'Universit\'e C\^ote d'Azur, 06902 Sophia Antipolis, France (e-mail: samir.perlaza@inria.fr, xinying.zou@inria.fr). }
\thanks{Samir M. Perlaza is also with the Department of Electrical and Computer Engineering, Princeton University, Princeton, NJ 08544 USA; and also with the GAATI Mathematics Laboratory, University of French Polynesia, 98702 Faaa, French Polynesia.}
}


 \maketitle 

\begin{abstract} 
In this paper, the method of gaps, a technique for deriving closed-form expressions in terms of information measures for the generalization error of supervised machine learning algorithms is introduced. The method relies on the notion of \emph{gaps}, which characterize the variation of the expected empirical risk (when either the model or dataset is kept fixed) with respect to changes in the probability measure on the varying parameter (either the dataset or the model, respectively). 
This distinction results in  two classes of gaps: Algorithm-driven gaps (fixed dataset) and data-driven gaps (fixed model). 
In general, the method relies on two central observations: 
$(i)$~The generalization error is the expectation of an algorithm-driven gap or a data-driven gap. In the first case, the expectation is with respect to a measure on the datasets; and in the second case, with respect to a measure on the models.
$(ii)$~Both algorithm-driven gaps and data-driven gaps exhibit closed-form expressions in terms of relative entropies. In particular, algorithm-driven gaps involve a Gibbs probability measure on the set of models, which represents a supervised Gibbs algorithm. Alternatively, data-driven gaps involve a worst-case data-generating (WCDG) probability measure on the set of data points, which is also a Gibbs probability measure. Interestingly, such Gibbs measures, which are exogenous to the analysis of generalization, place both the supervised Gibbs algorithm and the WCDG probability measure as natural references for the analysis of supervised learning algorithms.
New exact expressions and all existing exact expressions for the generalization error of supervised  learning algorithms can be obtained with the proposed method. 
Such new expressions are intended as structural and conceptual characterizations, rather than computational shortcuts. Finally, these expressions unveil strong connections among  generalization, hypothesis testing, information measures, and Pythagorean identities.   

\end{abstract}


\section{Introduction}

The \emph{generalization error} of a supervised machine learning algorithm is the expectation (w.r.t. the joint probability distribution of datasets and models) of  the difference between its population risk and its empirical risk \cite{bousquet2002stability, aminian2021exact,  aminian2024information, russo2019much, xu2017information, aminian2021jensen, aminian2022tighter, Perlaza-ISIT-2022, PerlazaTIT2024,  bu2020tightening, esposito2021generalization}. 
The generalization error has become a key metric for assessing an algorithm’s performance and its ability to generalize beyond training data.   
Traditional approaches for analyzing the generalization error generally fall into two categories. The first category derives upper bounds on the generalization error based on properties of the hypothesis class, such as Rademacher complexity, VC dimension \cite{vapnik2006estimation, alon1997scale, mohri2008rademacher, shalev2014understanding, cherkassky1999model,sachs2023generalization,bartlett2002rademacher,koltchinskii2006local,bartlett2002localized};  covering numbers, or metric entropy \cite{duchi2021learning,haussler1992decision,williamson2000entropy}. 
However, for many architectures of deep neural networks, the VC dimension is extremely large (even infinite), which limits the relevance of generalization analyses based on such a metric~\cite{bartlett2021deep,nagarajan2019ucfail}.
The second category examines the dependency between learning algorithms and data distributions through various information-theoretic metrics \cite{jiao2017dependence, wang2019information, issa2019strengthened, asadi2018chaining, lopez2018generalization,  asadi2020chaining, hafez2020conditioning, haghifam2020sharpened, rodriguez2021tighter}. 
The initial characterizations of the generalization error using information measures were introduced for the Gibbs algorithm in \cite{aminian2021exact, aminian2024information}. Later, such a result was extended for the case of Gibbs algorithms using $\sigma$-finite reference measures \cite{PerlazaTIT2024}.
For learning algorithms beyond Gibbs, two main types of guarantees are provided under the information-theoretic framework. The first type is often referred to as \emph{guarantees in probability}, which assert that with high probability over the probability measure of the datasets, the expected risk induced by the learning algorithm is bounded above by a certain threshold. This type of guarantees is referred to as  PAC-Bayesian generalization bounds as in \cite{catoni2007pac, viallard2024learning, AlquierPACjournal2024,shawe1997pac, guedj2019free, mcallester2003pac, haddouche2020pacbayes,dziugaite2017nonvacuous}. 
In parallel, \emph{guarantees in expectation} provide bounds on the generalization error. To account for the dependence between the learning algorithm and the data distribution, mutual information is a common metric in these analyses. 
Such an information measure was  first introduced in \cite{pmlr-v51-russo16} and later in \cite{xu2017information}. Chaining techniques were incorporated in \cite{asadi2018chaining, asadi2020chaining} to derive upper bounds, with similar bounds appearing in \cite{bu2020tightening, rodriguez2021upper}. Alternative metrics such as Wasserstein distance \cite{aminian2022tighter, lopez2018generalization, wang2019information, rodriguez2021tighter}, maximal leakage \cite{issa2019strengthened, esposito2020robust}, mutual $f$-information \cite{masiha2023f}, and Jensen-Shannon divergence \cite{aminian2021jensen} have also been used to derive generalization bounds. Comprehensive literature reviews of these guarantees can be found in \cite{hellstrom2023generalization} and, more recently, in \cite{rodriguez2024information}. Nonetheless, for any specific use, the tightness of many of these bounds cannot be guaranteed, as noted in~\cite{Jiang2020Fantastic} and \cite{gastpar2023fantastic,zhang2017understanding,belkin2019reconciling}. 
This highlights the need for exact characterizations of the generalization error for supervised machine  learning algorithms, as those presented in \cite{zouJSAIT2024} and \cite{InriaRR9539}. These new exact expressions involve several information measures, e.g., relative entropy, mutual information and lautum information \cite{palomar2008lautum}, and significantly motivated this work.  

The present work is dedicated to deriving closed-form expressions for the generalization error of machine learning algorithms through the lens of information theory. The main contribution is a general method for such a purpose, which has been coined \emph{the method of gaps}. Interestingly, all existing exact  expressions for the generalization error of machine learning algorithms can be obtained with the proposed method. 
Many new expressions can also be obtained. Nonetheless, it is important to clarify that these expressions are not intended to provide computational shortcuts for estimating the generalization error, nor do they directly yield new surrogate objectives for training supervised learning algorithms. Their role is instead structural and conceptual. They provide exact representations of the generalization error in terms of information-theoretic quantities and reference probability measures that have a precise interpretation in supervised learning. In this sense, the contribution of the paper is to reveal how the generalization error can be decomposed into, compared to, and interpreted through canonical objects such as the Gibbs algorithm, the WCDG probability measure, relative entropy, mutual information, and lautum information. The usefulness of these identities therefore lies in the conceptual understanding they provide and in the connections they establish with hypothesis testing, information geometry, and Pythagorean relations involving information measures.

\subsection{Gaps and the Method of Gaps}
The \emph{method of gaps} owes its name to the notion of \emph{gap}, which can be defined by highlighting that the empirical risk is a function whose arguments are a model and a dataset. A gap characterizes the variation of the expectation on the empirical risk, when either the model or the dataset is kept fixed, and the probability measure of the varying parameter changes from an initial measure to a final measure.
From this perspective, two classes of gaps might be considered. 
When the dataset is kept fixed and the probability distribution on the set of models is let to vary, an algorithm-driven gap is obtained. 
Alternatively, when the model is kept fixed and the probability distribution on the set of datasets varies, a data-driven gap is obtained. 
Hence, a gap depends on three parameters: Either a model or a dataset; an initial probability measure; and a final probability measure.
Certain algorithm-driven and data-driven gaps have been shown to have closed-form expressions in terms of relative entropies. The first closed-form expressions of an algorithm-driven gap were presented in \cite[Theorem~$1$]{Perlaza-ISIT2023b} to quantify the variation of the expected empirical risk due to a deviation from the Gibbs algorithm to an alternative algorithm. This foundational result serves as a key building block in this work for deriving closed-form expressions for algorithm-driven gaps due to transitions between arbitrary algorithms. See for instance, Theorem~\ref{TheoremRGgeneral} in Section~\ref{SecSensitivityArbitraryA}.
Alternatively, the first closed-form expression for a data-driven gap due to a change from a worst-case data-generating (WCDG) probability measure to an arbitrary measure was presented in \cite[Theorem~$5.2$]{zou2024Generalization}. This crucial result formed the basis for deriving the closed-form expressions for data-driven gaps due to a change between arbitrary probability measures in \cite[Theorem~$8$]{zouJSAIT2024}.
In a nutshell, exact expressions for algorithm-driven gaps involve the Gibbs algorithm studied in~\cite{PerlazaTIT2024}, while exact expressions for data-driven gaps involve the WCDG probability measure introduced in~\cite{zou2024Generalization}. In both cases, these expressions are surprisingly simple and flexible, as they contain free parameters that can be chosen subject to few constraints. 

From this perspective, two variants of the method of gaps can be identified. 
The former, which is based on algorithm-driven gaps, involves two steps:~$(i)$ To express the generalization error as the expectation (with respect to a probability measure on the datasets) of a particular algorithm-driven gap (see Lemma~\ref{LemmaJun4at9h32in2024} in Section~\ref{SecCharAlgorithmDrivenGaps}); and~$(ii)$ To exploit the closed-form expressions for algorithm-driven gaps presented in Section~\ref{SecSensitivityArbitraryA} to obtain expressions for the generalization error. 
The method of gaps via algorithm-driven gaps was implicitly used to obtain the exact expression of the generalization error of the Gibbs algorithm in \cite[Theorem~$17$]{PerlazaTIT2024}\footnote{Such an exact expression was first presented in \cite{aminian2021exact}. Nonetheless, the method of proof used therein cannot be related to the method of gaps}. Although no particular method is mentioned, \cite[Theorem~$16$]{PerlazaTIT2024}  can be associated with \emph{step one} of the method of gaps; and \cite[Theorem~$17$]{PerlazaTIT2024} to \emph{step two}. Unfortunately, the closed-form expressions for algorithm-driven gaps developed in \cite{Perlaza-ISIT-2022, PerlazaTIT2024} are limited to transitions from the Gibbs algorithm to alternative algorithms. This limits such a theoretical framework to obtain exact expressions for the generalization error of arbitrary algorithms. Nonetheless, these findings are essential for obtaining the results presented in this work.

The second variant of the method of gaps, which is based on data-driven gaps, also consists of two  steps:~$(i)$ To express the generalization error as the expectation (with respect to a probability measure on the models) of a particular data-driven gap (see \cite[Equation~$(75)$]{zouJSAIT2024}); and~$(ii)$ To exploit the closed-form expressions for data-driven gaps presented in Section~\ref{SecDataDrivenGaps} to obtain expressions for the generalization error. 
This second variant was the method of proof used in~\cite{zouJSAIT2024} to obtain the first exact expression  \cite[Theorem~$10$]{zouJSAIT2024} for the generalization error of arbitrary machine learning algorithms. Using the same method of proof, other closed-form expressions  \cite[Theorem~$11$ and Theorem~$13$]{InriaRR9539} have been obtained. 
The expressions obtained via this second variant of the method of gaps are surprisingly flexible, as they depend on parameters that can be arbitrarily chosen subject to mild conditions. However, these expressions are developed under the assumption that datasets consist of data points that are independent and identically distributed.
Interestingly, this assumption has been relaxed in~\cite{InriaRR9599} while adapting the method of gaps, as presented in this work,  to the case of federated supervised machine learning algorithms. 

Each variant of the method of gaps leads to different equivalent expressions for the generalization error of supervised machine learning algorithms. 
Interestingly, these new expressions lead to new perspectives from which the generalization error can be studied. In particular, strong connections with mismatched hypothesis testing~\cite{boroumand2022mismatched} and classical hypothesis testing~\cite{lehmann2005testing} are identified. Moreover, as a byproduct, several Pythagorean identities involving the generalization error and relative entropy are established. 

\subsection{Notation}
Sets are denoted by calligraphic uppercase letters, except sets of probability measures. Given a set~$\set{M}$ and a sigma-field~$\mathscr{F}$ on~$\set{M}$, the set of all probability measures that can be defined on the measurable space~$\left(\set{M},\mathscr{F}\right)$ is denoted by~$\triangle\left(\set{M},\mathscr{F}\right)$ or simply~$\triangle\left(\set{M}\right)$, when the sigma-field is fixed in the analysis.
If~$\set{M} \subseteq \reals^d$, then the Borel sigma-field on~$\set{M}$ is denoted by~$\BorSigma{\set{M}}$.
The subset of measures in~$\triangle\left(\set{M}\right)$ that are absolutely continuous with respect to a probability measure~$Q$ is denoted by~$\triangle_{Q}\left(\set{M}\right)$. 
Given a measure~$P \in \triangle_{Q}\left(\set{M}\right)$, the Radon-Nikodym derivative of~$P$ with respect to~$Q$ is denoted by~$\frac{\mathrm{d}P}{\mathrm{d}Q}$.
Given a set~$\set{N}$, the set of all probability measures defined on~$\left(\set{M},\mathscr{F}\right)$ conditioned on an element of~$\set{N}$ is denoted by~$\triangle\left( \set{M} | \set{N} \right)$. More specifically, given a measure~$P_{M | N} \in \triangle\left( \set{M} | \set{N} \right)$ and~$n \in \set{N}$, the measure~$P_{M | N = n}$ is in~$\triangle\left( \set{M} \right)$.

%%%
The \emph{relative entropy} is defined below as the extension to~$\sigma$-finite measures of the relative entropy usually defined for probability measures.
\begin{definition}[Relative Entropy]\label{DefRelEntropy}
	Given two~$\sigma$-finite measures~$P_1$ and~$P_2$ on the same measurable space, such that~$P_2$ is absolutely continuous with respect to~$P_1$,  the relative entropy of~$P_2$ with respect to~$P_1$ is
	\begin{equation}
\label{EqNovember18at10h45in2024SophiaAntipolis}	
		\KL{P_2}{P_1} = \int \frac{\mathrm{d}P_2}{\mathrm{d}P_1}(x)  \log\left( \frac{\mathrm{d}P_2}{\mathrm{d}P_1}(x)\right)  \mathrm{d}P_1(x),
	\end{equation}
	where the function~$\frac{\mathrm{d}P_2}{\mathrm{d}P_1}$ is the Radon-Nikodym derivative of~$P_2$ with respect to~$P_1$.
\end{definition}

Given a probability measure~$P_{N} \in \triangle\left(\set{N}\right)$, the mutual information (see \cite{Shannon-1948a} and \cite{Shannon-1948b}) and lautum information (see \cite{palomar2008lautum}) induced by the measures~$P_{M | N}$ and~$P_{N}$ are denoted respectively by
\begin{IEEEeqnarray}{rcl}
	\label{Eqmutual}
	I\left( P_{M | N}; P_{N} \right) & = & \int \KL{P_{M|N = n}}{P_{M}} \mathrm{d} P_{N}(n), \mbox{ and }\\
	\label{Eqlautum}
	L\left( P_{M | N}; P_{N} \right) & = & \int \KL{P_{M}}{P_{M|N = n}} \mathrm{d} P_{N}(n),
\end{IEEEeqnarray}
with~$P_{M} \in \triangle\left(\set{M}\right)$ being the marginal measure induced by~$P_{M | N}$ and~$P_{N}$; and~$\KL{\cdot}{\cdot}$ being the relative entropy in~\eqref{EqNovember18at10h45in2024SophiaAntipolis}.

\section{Statistical Supervised Machine Learning}\label{SubSecSML}

Let~$\set{M}$,~$\set{X}$ and~$\set{Y}$, with~$\set{M} \subseteq \reals^{d}$, be sets of \emph{models}, \emph{patterns}, and \emph{labels}, respectively.  %Let also~$\left( \set{X} \times \set{Y} \right) = \set{X} \times \set{Y}$.
A pair~$(x,y) \in \set{X} \times \set{Y}$ is referred to as a \emph{labeled pattern} or as a \emph{data point}.
%%
%Let the probability measure
%\begin{IEEEeqnarray}{rcl}\label{EqGTDG}
%	P_{Z} \in \triangle\left( \set{X} \times \set{Y}\right)
%\end{IEEEeqnarray}
%be the one from which data points are indepe ndently sampled.
%%
A training dataset~$\vect{z} \in \left( \set{X} \times \set{Y}\right)^{n}$ is a tuple of~$n$ data points of the form:
\begin{equation}\label{EqTheDataSet}
\vect{z} = \big(\left(x_1, y_1 \right), \left(x_2, y_2 \right), \ldots, \left(x_n, y_n \right)\big)  \in \left( \set{X} \times \set{Y}\right)^n.
\end{equation}  
Let the function~$f: \set{M} \times \set{X} \rightarrow \set{Y}$ be such that the label assigned to the pattern~$x$ according to the model~$\vect{\theta} \in \set{M}$ is~$f(\vect{\theta}, x)$.
The function~$f(\vect{\theta}, \cdot) : \set{X} \to \set{Y}$ might for instance represent a neural network whose  weights are vectorized into~$\vect{\theta}$.  
Let the function 
\begin{equation}\label{EqEllHat}
\hat{\ell}: \set{Y} \times \set{Y} \rightarrow [0, +\infty],
\end{equation} 
be the risk or loss function. Such a function is often of the form~$\hat{\ell}(\mu, \nu) = \abs{\mu - \nu}^p$, with~$p \geqslant 1$; or~$\hat{\ell}(\mu, \nu) = \mathds{1}_{\lbrace \mu \neq \nu \rbrace}$.
In the following, the risk function~$\hat{\ell}$ is assumed to be a nonnegative function and  for all~$y \in \set{Y}$,~$\hat{\ell}\left( y , y\right)~=~0$. 

In general,  given a labeled pattern~$(x, y) \in \set{X}\times\set{Y}$, the  risk induced by a model~$\vect{\theta} \in \set{M}$ is~$\hat{\ell}\left( f(\vect{\theta}, x), y \right)$.  
Nonetheless, in order to capture the fact that a model $\vect{\theta}$ does not necessarily lead to a deterministic label for a given pattern $x$, the analysis follows by considering the function~$\ell: \set{X} \times \set{Y} \times \set{M} \to [0, +\infty]$. When such a deterministic mapping exists, then, 
\begin{IEEEeqnarray}{rcl}
\label{EqEll}
\ell(x,y,\vect{\theta}) & = & \hat{\ell}\left( f(\vect{\theta}, x), y \right).
\end{IEEEeqnarray} 
The problem of supervised machine learning boils down to devising an algorithm that takes as input a training dataset, as~$\vect{z}$ in~\eqref{EqTheDataSet}, and outputs a model~$\vect{\theta}$, with certain probability, to satisfy some performance metric.
The notion of \emph{machine learning algorithm} is formally introduced hereunder.
\begin{definition}[Algorithm]\label{DefSMSA}
A conditional probability measure~$P_{\vect{\Theta} | \vect{Z}} \in  \triangle\left(\set{M}| \left( \set{X} \times \set{Y} \right)^{n} \right)$ is said to represent a supervised machine learning algorithm that can be trained with $n$ data points.
\end{definition}

Let~$P_{\vect{\Theta} | \vect{Z}} \in  \triangle\left(\set{M}| \left( \set{X} \times \set{Y} \right)^{n} \right)$ be an algorithm. Hence, the instance of such an algorithm trained upon the dataset~$\vect{z}$ in~\eqref{EqTheDataSet} is denoted by~$P_{\vect{\Theta} | \vect{Z} = \vect{z}}$, which is simply a probability measure in~$\triangle\left( \set{M}\right)$.

A central performance metric for a fixed model is the empirical risk it induces with respect to a given dataset. Such a metric is defined as follows.
\begin{definition}[Empirical Risk]\label{DefEmpiricalRisk}
The \emph{empirical risk} induced by a model~$\vect{\theta} \in \set{M}$ with respect to the dataset~$\vect{z}$ in~\eqref{EqTheDataSet} is determined by the  function~$\mathsf{L}: \left( \set{X} \times \set{Y} \right)^n \times \set{M} \rightarrow [0, +\infty ]$, which satisfies  
\begin{IEEEeqnarray}{rcl}
\label{EqLxy}
\mathsf{L} \left(\vect{z}, \vect{\theta} \right)  & = & 
\frac{1}{n}\sum_{i=1}^{n}  \ell\left( x_i, y_i, \vect{\theta} \right).
\end{IEEEeqnarray}
\end{definition}
The expectation of the empirical risk~$\mathsf{L}\left(\vect{z}, \vect{\theta} \right)$ in~\eqref{EqLxy} requires particular notations depending on whether such an expectation is with respect to a measure on the models or the datasets. 
More specifically, given a model~$\vect{\theta}$, which is kept fixed, the expectation of~$\mathsf{L}\left(\vect{z}, \vect{\theta} \right)$ under the assumption that the dataset~$\vect{z}$ is drawn from a probability measure~$P \in \triangle \left( \left(\set{X} \times \set{Y} \right)^n \right)$ is denoted using the functional~$\mathsf{R}_{\vect{\theta}}: \triangle\left( \left(\set{X} \times \set{Y} \right)^n \right) \to \reals$, which satisfies
\begin{IEEEeqnarray}{rCl}
\label{EqRModel}
\mathsf{R}_{\vect{\theta}} \left( P \right) & = & \int \mathsf{L}\left(\vect{z}, \vect{\theta} \right) \mathrm{d}P (\vect{z}).
\end{IEEEeqnarray}

Alternatively, given a dataset~$\vect{z}$, which is kept fixed, the expectation of~$\mathsf{L}\left(\vect{z}, \vect{\theta} \right)$ under the assumption that the model~$\vect{\theta}$ is drawn from a probability measure~$P \in \triangle \left( \set{M} \right)$ is denoted using the functional~$\mathsf{R}_{\vect{z}}: \triangle\left(\set{M} \right) \to \reals$, which satisfies
\begin{equation}
\label{EqRxy}
\mathsf{R}_{\vect{z}}\left( P  \right) = \int \mathsf{L} \left(\vect{z} , \vect{\theta} \right)  \mathrm{d} P(\vect{\theta}).
\end{equation}
The following section introduces the generalization error. Later,  it is shown that the generalization error of any supervised machine learning algorithm can be characterized by studying the sensitivity of the functionals~$\mathsf{R}_{\vect{\theta}}$ in~\eqref{EqRModel} and~$\mathsf{R}_{\vect{z}}$ in~\eqref{EqRxy}.

\section{Generalization Error}
 
 In a practical setting, the generalization capability of a supervised machine learning algorithm is evaluated via \emph{test datasets}, which are different from the \emph{training dataset} \cite{Perlaza-ISIT2023b}.
In particular, given an algorithm~$P_{\vect{\Theta} | \vect{Z}} \in  \triangle\left(\set{M}| \left( \set{X} \times \set{Y} \right)^n \right)$, a test dataset~$\vect{u}  \in \left( \set{X} \times \set{Y} \right)^n$, and a training dataset~$\vect{z} \in \left( \set{X} \times \set{Y} \right)^n$, a typical evaluation of the generalization capability of the instance~$P_{\vect{\Theta} | \vect{Z} = \vect{z}}$ of such an algorithm consists in calculating the difference between the expected empirical risk it  induces on the test dataset~$\vect{u}$ and on the training dataset~$\vect{z}$. That is, the difference
\begin{IEEEeqnarray}{rcl}
\label{EqJun4At8h36in2024}
\mathsf{R}_{\vect{u}} \left( P_{\vect{\Theta} | \vect{Z} = \vect{z}} \right) - 
\mathsf{R}_{\vect{z}}\left( P_{\vect{\Theta} | \vect{Z} = \vect{z}} \right),
\end{IEEEeqnarray}
where the functionals~$\mathsf{R}_{\vect{u}}$ and~$\mathsf{R}_{\vect{z}}$ are both defined as in~\eqref{EqRxy}.
In a nutshell, the evaluation consists in calculating the variation of the expected empirical risk induced by the algorithm~$P_{\vect{\Theta} | \vect{Z}= \vect{z} }$ with respect to a dataset that changes from the training dataset~$\vect{z}$ to the test dataset~$\vect{u}$.
A small variation is interpreted as a good capability of the algorithm~$P_{\vect{\Theta} | \vect{Z}}$ to generalize the training dataset~$\vect{z}$ into the test dataset~$\vect{u}$. While the difference in~\eqref{EqJun4At8h36in2024} being smaller than a certain threshold for certain test datasets is not a guarantee of good generalization into other test datasets, this is the most popular tool used by  practitioners for evaluating machine learning algorithms. This is in part, because in most cases, test and training datasets are the only data actually available. 
 
The analysis of the generalization capabilities of an algorithm~$P_{\vect{\Theta} | \vect{Z}}$ can be deepened if further assumptions are adopted. For instance, it can be assumed that test and training datasets are both independently drawn from the same probability measure. Indeed, this is one of the most popular assumptions, and the one typically adopted to define the \emph{generalization error}.
In a nutshell, the generalization error induced by an algorithm~$P_{\vect{\Theta} | \vect{Z}}$ is the expectation of the variation in~\eqref{EqJun4At8h36in2024} under the assumption that both test and training datasets are independently sampled from the same probability measure in~$ \triangle\left( \left(\set{X} \times \set{Y}\right)^n \right)$. 
 \begin{definition}[Generalization Error]\label{DefDEGG}
The generalization error induced by the algorithm~$P_{\vect{\Theta} | \vect{Z}} \in  \triangle\left(\set{M}| \left( \set{X}\times\set{Y} \right)^{n} \right)$ under the assumption that  training and test datasets are independently sampled from a probability measure~$P_{\vect{Z}} \in \triangle\left( \left(\set{X} \times \set{Y}\right)^n \right)$, which is denoted by~$\overline{\overline{\mathsf{G}}} \left(P_{\vect{\Theta} | \vect{Z}}, P_{\vect{Z}} \right)$, is 
\begin{IEEEeqnarray}{rcl}
\nonumber
& & \overline{\overline{\mathsf{G}}} \left(P_{\vect{\Theta} | \vect{Z}}, P_{\vect{Z}} \right) \\
\label{EqJun4at9h02in2024}
& \triangleq & \iint  \left(  
\mathsf{R}_{\vect{u}} \left( P_{\vect{\Theta} | \vect{Z} = \vect{z}} \right) - 
\mathsf{R}_{\vect{z}}\left( P_{\vect{\Theta} | \vect{Z} = \vect{z}} \right)
 \right) \mathrm{d}P_{\vect{Z}} \left( \vect{u} \right) \mathrm{d}P_{\vect{Z}} \left( \vect{z} \right).\Dsupersqueezeequ\IEEEeqnarraynumspace
\end{IEEEeqnarray}
\end{definition}
Often, the generalization error~$\overline{\overline{\mathsf{G}}} \left(P_{\vect{\Theta} | \vect{Z}}, P_{\vect{Z}} \right)$ is defined as the expectation of the difference between the \emph{population error} and the \emph{empirical risk}, cf.,~\cite[Equation~$2$]{aminian2021exact} and~\cite[Equation~$4$]{xu2017information}.   
The population error, induced by a model~$\vect{\theta} \in \set{M}$ under the assumption that datasets are drawn from the probability measure~$P_{\vect{Z}}$, is~$\mathsf{R}_{\vect{\theta}} \left( P_{\vect{Z}} \right)$, where the functional~$\mathsf{R}_{\vect{\theta}}$ is defined in~\eqref{EqRModel}. 
Alternatively, the training error induced by the model~$\vect{\theta}$ with respect to a training dataset~$\vect{z} \in \left( \set{X} \times \set{Y} \right)^n$ is~$\mathsf{L}\left( \vect{z} , \vect{\theta} \right)$, where the function~$\mathsf{L}$ is defined in~\eqref{EqLxy}.
Using these elements, the following lemma introduces an alternative expression for the generalization error~$\overline{\overline{\mathsf{G}}} \left(P_{\vect{\Theta} | \vect{Z}}, P_{\vect{Z}} \right)$ in terms of the \emph{population error} and the \emph{empirical risk}.
 
\begin{lemma}\label{LemmaNovember5at8h38in2024SophiaAntipolis}
The generalization error~$\overline{\overline{\mathsf{G}}} \left(P_{\vect{\Theta} | \vect{Z}}, P_{\vect{Z}} \right)$ in~\eqref{EqJun4at9h02in2024} satisfies
\begin{IEEEeqnarray}{rcl}
\nonumber
& &  \overline{\overline{\mathsf{G}}} \left(P_{\vect{\Theta} | \vect{Z}}, P_{\vect{Z}} \right) \\
\label{EqJune5at15h05in2026SophiaAntipolis}
& = &
\iint \bigg( \mathsf{R}_{\vect{\theta}} \left( P_{\vect{Z}} \right) - \mathsf{L}\left( \vect{z} , \vect{\theta} \right)\bigg)\mathrm{d} P_{\vect{\Theta} | \vect{Z} = \vect{z}} \left( \vect{\theta} \right) \mathrm{d}P_{\vect{Z}} \left( \vect{z} \right), \spnum
\end{IEEEeqnarray}
where the functional~$\mathsf{R}_{\vect{\theta}}$ is defined in~\eqref{EqRModel} and the function~$\mathsf{L}$ is defined in~\eqref{EqLxy}.
\end{lemma}
\begin{IEEEproof}
The proof follows from~\eqref{EqJun4at9h02in2024} by observing the following equalities:
\begin{IEEEeqnarray}{rcl}
\nonumber
& & \iint \mathsf{R}_{\vect{u}} \left( P_{\vect{\Theta} | \vect{Z} = \vect{z}} \right)   \mathrm{d}P_{\vect{Z}} \left( \vect{u} \right) \mathrm{d}P_{\vect{Z}} \left( \vect{z} \right)\\
\label{EqNovember6at8h49in2024SophiaAntipolis}
& = & \iint \int \mathsf{L} \left(\vect{u},\vect{\theta} \right)  \mathrm{d} P_{\vect{\Theta} | \vect{Z} = \vect{z}} \left(\vect{\theta} \right)  \mathrm{d}P_{\vect{Z}} \left( \vect{u} \right) \mathrm{d}P_{\vect{Z}} \left( \vect{z} \right)  \IEEEeqnarraynumspace\\
\label{EqNovember6at8h49in2024SophiaAntipolisb}
& = & \iint \left( \int \mathsf{L} \left(\vect{u},\vect{\theta} \right) \mathrm{d}P_{\vect{Z}} \left( \vect{u} \right) \right) \mathrm{d} P_{\vect{\Theta} | \vect{Z} = \vect{z}} \left(\vect{\theta} \right)   \mathrm{d}P_{\vect{Z}} \left( \vect{z} \right) \\
\label{EqNovember6at8h49in2024SophiaAntipolisc}
 & = & \iint\mathsf{R}_{\vect{\theta}}\left( P_{\vect{Z}} \right)
 \mathrm{d} P_{\vect{\Theta} | \vect{Z} = \vect{z}} \left(\vect{\theta} \right)   \mathrm{d}P_{\vect{Z}} \left( \vect{z} \right)    ,
 \IEEEeqnarraynumspace
\end{IEEEeqnarray}
where the equality in~\eqref{EqNovember6at8h49in2024SophiaAntipolis} follows from~\eqref{EqRxy};
the equality in~\eqref{EqNovember6at8h49in2024SophiaAntipolisb} follows from \cite[Theorem~$2.6.6$]{ash2000probability}; and 
the equality in~\eqref{EqNovember6at8h49in2024SophiaAntipolisc} follows from~\eqref{EqRModel}. 
The proof follows by observing that 
\begin{IEEEeqnarray}{rcl}
\nonumber
& & \iint \mathsf{R}_{\vect{z}}\left( P_{\vect{\Theta} | \vect{Z} = \vect{z}} \right) \mathrm{d}P_{\vect{Z}} \left( \vect{u} \right) \mathrm{d}P_{\vect{Z}} \left( \vect{z} \right)\\
\label{EqNovember6at17h41in2024SophiaAntipolisA}
& = & \iint  \mathsf{L}\left( \vect{z} , \vect{\theta} \right) \mathrm{d} P_{\vect{\Theta} | \vect{Z} = \vect{z}} \left( \vect{\theta} \right) \mathrm{d}P_{\vect{Z}} \left( \vect{z} \right),
\end{IEEEeqnarray}
where the equality in~\eqref{EqNovember6at17h41in2024SophiaAntipolisA} follows from~\eqref{EqRxy}.
Finally, using~\eqref{EqNovember6at8h49in2024SophiaAntipolisc} and~\eqref{EqNovember6at17h41in2024SophiaAntipolisA} in~\eqref{EqJun4at9h02in2024} leads to
\begin{IEEEeqnarray}{rcl}
\nonumber
& & \overline{\overline{\mathsf{G}}} \left(P_{\vect{\Theta} | \vect{Z}}, P_{\vect{Z}} \right) \\
\label{EqNovember6at18h05in2024}
& = & \iint \bigg( \mathsf{R}_{\vect{\theta}} \left( P_{\vect{Z}} \right) - \mathsf{L}\left( \vect{z} , \vect{\theta} \right)\bigg)\mathrm{d} P_{\vect{\Theta} | \vect{Z} = \vect{z}} \left( \vect{\theta} \right) \mathrm{d}P_{\vect{Z}} \left( \vect{z} \right), \spnum
\end{IEEEeqnarray}
which completes the proof.
\end{IEEEproof}
The expression in~\eqref{EqJune5at15h05in2026SophiaAntipolis}, despite the equivalence, is far more popular than the one in~\eqref{EqJun4at9h02in2024}. Nonetheless, the latter is significantly more insightful and thus, the one used in the remainder of this work. 
This said, the main contribution of this manuscript is a method that allows constructing  explicit expressions for the generalization error in~\eqref{EqJun4at9h02in2024}. In the following, such method is referred to as \emph{the method of gaps} and is introduced in the next section. 

\section{The Method of Gaps}

The method of gaps is a two-step technique to construct explicit expressions for the generalization error~$\overline{\overline{\mathsf{G}}} \left(P_{\vect{\Theta} | \vect{Z}}, P_{\vect{Z}} \right)$ in~\eqref{EqJun4at9h02in2024} in terms of information measures. 
The method is based on the analysis of the variations  of the functionals~$\mathsf{R}_{\vect{\theta}}$ in~\eqref{EqRModel} and~$\mathsf{R}_{\vect{z}}$ in~\eqref{EqRxy}.  More specifically, given two probability measures~$P_1$ and~$P_2$ in~$\triangle\left( \set{M} \right)$ and two probability measures~$Q_1$ and~$Q_2$ in~$\triangle\left(\left( \set{X} \times \set{Y} \right)^n\right)$, the differences~$\mathsf{R}_{\vect{z}}\left( P_1 \right) -  \mathsf{R}_{\vect{z}}\left( P_2 \right)$ and~$\mathsf{R}_{\vect{\theta}}\left( Q_{1} \right)  -  \mathsf{R}_{\vect{\theta}}\left( Q_{2} \right)$  capture the variation of the  functionals~$\mathsf{R}_{\vect{z}}$ and~$\mathsf{R}_{\vect{\theta}}$  due to the change of measure from~$P_2$ to~$P_1$, and from~$Q_2$ to~$Q_1$, respectively. These differences are referred to as \emph{gaps}.  
In a nutshell, the first step of the method of gaps consists in expressing the generalization error as either:~$(a)$~an expectation (with respect to some probability measure~$Q \in \triangle\left( \left(\set{X} \times \set{Y} \right)^n\right)$) of the gap~$\mathsf{R}_{\vect{z}}\left( P_1 \right) -  \mathsf{R}_{\vect{z}}\left( P_2 \right)$, for some specific measures~$P_1$ and~$P_2$; or~$(b)$~an expectation  (with respect to some probability measure~$P \in \triangle\left( \set{M} \right)$) of the gap~$\mathsf{R}_{\vect{\theta}}\left( Q_1 \right) -  \mathsf{R}_{\vect{\theta}}\left( Q_2 \right)$, for some specific measures~$Q_1$ and~$Q_2$.
The second step consists in leveraging the properties of such gaps for expressing the expectations mentioned above  in terms of information measures.
To formalize the presentation of the method, the notion of gap is developed hereunder. 

\subsection{Gaps}

The variations of the functional~$\mathsf{R}_{\vect{\theta}}$ in~\eqref{EqRModel}, for some fixed model~$\vect{\theta} \in \set{M}$; and those of the functional~$\mathsf{R}_{\vect{z}}$ in~\eqref{EqRxy}, for some fixed dataset~$\vect{z} \in \left( \set{X} \times \set{Y} \right)^n$, due to changes in their arguments, are referred to as \emph{gaps}. These gaps are studied via the following functionals:
\begin{IEEEeqnarray}{rCl}
\label{EqDemasiadoA}
\mathsf{G}: \left( \set{X}\times \set{Y}\right)^n \times \triangle\left( \set{M} \right) \times \triangle\left( \set{M} \right) \to \reals; \mbox{ and }\\
\label{EqDemasiadoB}
\mathsf{G}: \set{M}  \times \triangle\left( \set{X}\times \set{Y} \right) \times \triangle\left( \set{X}\times \set{Y}\right)  \to \reals,
\end{IEEEeqnarray}
where the former satisfies
\begin{IEEEeqnarray}{rCl}
\label{EqGZeta}
\mathsf{G}\left( \vect{z}, P_1, P_2 \right) & = & \mathsf{R}_{\vect{z}}\left( P_1 \right) -  \mathsf{R}_{\vect{z}}\left( P_2 \right),
\end{IEEEeqnarray}
and the latter satisfies
\begin{IEEEeqnarray}{rCl}
\nonumber
& &\mathsf{G}\left( \vect{\theta}, Q_1, Q_2 \right) \\
\label{EqGTheta}
& = & \int  \ell\left( x, y, \vect{\theta} \right) \mathrm{d}Q_{1}\left( x, y \right) - \int  \ell\left( x, y, \vect{\theta} \right) \mathrm{d}Q_{2}\left( x, y \right),\IEEEeqnarraynumspace
\end{IEEEeqnarray}
with the function~$\ell$ defined in~\eqref{EqEll}. 
Note that the measures~$P_1$ and~$P_2$ in~\eqref{EqGZeta} are in~$\triangle\left( \set{M} \right)$, while the measures~$Q_1$ and~$Q_2$ in~\eqref{EqGTheta} are in~$\triangle\left( \set{X}\times\set{Y} \right)$. 
Despite the same notation~$\mathsf{G}$ for the functionals in~\eqref{EqGZeta} and in~\eqref{EqGTheta}, their distinction is immediate from their arguments, and thus, there is no ambiguity.  

Given a fixed dataset~$\vect{z}$; and  two algorithms~$P_{\vect{\Theta} | \vect{Z}} \in \triangle\left( \set{M} | \left( \set{X}  \times \set{Y} \right)^n \right)$ and~$Q_{\vect{\Theta} | \vect{Z}} \in \triangle\left( \set{M} | \left( \set{X}  \times \set{Y} \right)^n \right)$ respectively trained upon some datasets~$\vect{z}_1$ and~$\vect{z}_2$, the gap~$\mathsf{G}\left( \vect{z},P_{\vect{\Theta} | \vect{Z}= \vect{z}_1}, Q_{\vect{\Theta} | \vect{Z} = \vect{z}_{2}}\right)$ characterizes the variation of the expected empirical risk~$\mathsf{R}_{\vect{z}}$ when the algorithm changes from~$Q_{\vect{\Theta} | \vect{Z}= \vect{z}_2}$ to~$P_{\vect{\Theta} | \vect{Z}= \vect{z}_1}$.
This justifies coining the gap~$\mathsf{G}\left( \vect{z}, P_1, P_2 \right)$ in~\eqref{EqGZeta} \emph{algorithm-driven gap}, as~$P_1$ and~$P_2$ can be seen as instances of particular algorithms.

Similarly, the gap~$\mathsf{G}\left(\vect{\theta}, Q_1, Q_2 \right)$ in~\eqref{EqGTheta} characterizes the variation of the expected empirical risk~$\mathsf{R}_{\vect{\theta}}$  in~\eqref{EqRModel} when datasets are formed by~$n$ data points  independently sampled from~$Q_1$ instead of~$Q_2$.  This observation becomes clearer by considering two product probability measures~$P_{\vect{Z}} \in \triangle\left( \left( \set{X} \times \set{Y} \right)^n\right)$ and~$Q_{\vect{Z}} \in \triangle\left( \left( \set{X} \times \set{Y} \right)^n\right)$ formed by the measures~$P_{Z} \in \triangle\left( \set{X} \times \set{Y} \right)$ and~$Q_{Z} \in \triangle\left( \set{X} \times \set{Y} \right)$. More specifically, for all measurable subsets of~$\left( \set{X} \times \set{Y} \right)^n$ of the form~$\set{A}_1 \times \set{A}_2 \times \ldots \times \set{A}_n$,  it holds that
\begin{IEEEeqnarray}{rCl}
	\label{EqSaturdayMai20at13h20in2024}
P_{\vect{Z}} \left( \set{A}_1 \times \set{A}_2 \times \ldots \times \set{A}_n \right) & =&  \prod_{t =1}^{n} P_{Z} \left( \set{A}_t \right), \mbox{ and }\\
	\label{EqFridayAugust2at16h24in2024}
Q_{\vect{Z}} \left( \set{A}_1 \times \set{A}_2 \times \ldots \times \set{A}_n \right) & =&  \prod_{t =1}^{n} Q_{Z} \left( \set{A}_t \right).
\end{IEEEeqnarray}
Under this assumption, the following lemma presents an important property of the functional~$\Rloss$ in~\eqref{EqRModel}. 
\begin{lemma}
Consider a probability measure~$P_{\vect{Z}} \in \triangle\left( \left( \set{X} \times \set{Y} \right)^n \right)$ that satisfies~\eqref{EqSaturdayMai20at13h20in2024} for some~$P_{Z} \in \triangle\left( \set{X} \times \set{Y}\right)$. Then, the functional~$\Rloss$ in~\eqref{EqRModel} satisfies
\begin{IEEEeqnarray}{rCl}
	\Rloss \left(P_{\vect{Z}}\right) 
	&=& \int \ell\left(x,y,\vect{\theta}\right) \mathrm{d}P_Z(x,y),
\end{IEEEeqnarray}
where the function~$\ell$ is defined in~\eqref{EqEll}.
\end{lemma}\label{LemmaNovember4at21h56in2024Nice}
\begin{IEEEproof}
	Note that from~\eqref{EqRModel}, it follows that
	\begin{IEEEeqnarray}{rCl}
		\nonumber
		 \Rloss\left(P_{\vect{Z}}\right) &=&	\int \mathsf{L}\left(\vect{z},\vect{\theta}\right) \mathrm{d}P_{\vect{Z}}\left(\vect{z}\right)\\
\label{EqNovember4at21h45in2024Nice}		 
		 	&=&\int \frac{1}{n}\sum_{t=1}^{n} \ell\left( x_t, y_t,\vect{\theta}\right) \mathrm{d}P_{\vect{Z}}\left(\vect{z}\right)\\
\label{EqNovember4at21h45in2024Niceb}
		&=& \int \ell\left(x,y,\vect{\theta}\right) \mathrm{d}P_Z(x,y),
	\end{IEEEeqnarray}
where the equality in~\eqref{EqNovember4at21h45in2024Nice} follows from~\eqref{EqLxy}; and the equality in~\eqref{EqNovember4at21h45in2024Niceb} follows from  \cite[Theorem~$1.6.3$]{ash2000probability}.
This completes the proof.
\end{IEEEproof}

Lemma~\ref{LemmaNovember4at21h56in2024Nice}, given a fixed model~$\vect{\theta} \in \set{M}$ and the probability measures~$P_{\vect{Z}}$ and~$Q_{\vect{Z}}$ that respectively satisfy~\eqref{EqSaturdayMai20at13h20in2024} and~\eqref{EqFridayAugust2at16h24in2024}, for some~$P_Z$ and~$Q_Z$ in~$\triangle\left( \set{X} \times \set{Y} \right)$, leads to the following equality:
\begin{IEEEeqnarray}{rCl}
\mathsf{G}\left( \vect{\theta}, P_Z, Q_Z \right)  & = & \mathsf{R}_{\vect{\theta}}\left( P_{\vect{Z}} \right)  -  \mathsf{R}_{\vect{\theta}}\left( Q_{\vect{Z}} \right),
\end{IEEEeqnarray}
which is the variation of the functional~$\mathsf{R}_{\vect{\theta}}$ when its argument changes from~$Q_{\vect{Z}} \in \triangle\left( \left( \set{X} \times \set{Y} \right) ^n \right)$ to~$P_{\vect{Z}}  \in \triangle\left( \left( \set{X} \times \set{Y} \right) ^n \right)$. This justifies coining the gap~$\mathsf{G}\left( \vect{\theta}, Q_1, Q_2 \right)$ in~\eqref{EqGTheta} the \emph{data-driven gap}, as~$P_{\vect{Z}}$ and~$Q_{\vect{Z}}$ are interpreted as statistical descriptions of the datasets.
Note the limitation due to the fact that the analysis of the variations of the functional~$\mathsf{R}_{\vect{\theta}}$ via the functional~$\mathsf{G}$ in~\eqref{EqGTheta}, implicitly assumes that datasets are formed by~$n$ independent and identically distributed data points (see~\eqref{EqSaturdayMai20at13h20in2024}). This limitation is not present while studying the variation of the functional~$\mathsf{R}_{\vect{z}}$ via the functional~$\mathsf{G}$ in~\eqref{EqGZeta}. This induces two different variants of the method of gaps, the first one for the case of algorithm-driven gaps; and a more limited variant for the case of data-driven gaps. 

\subsection{Method of Algorithm-driven Gaps}\label{SecCharAlgorithmDrivenGaps}

\subsubsection{Step One}
The first step of the \emph{method of algorithm-driven gaps} consists in writing the generalization error~$\overline{\overline{\mathsf{G}}} \left(P_{\vect{\Theta} | \vect{Z}}, P_{\vect{Z}} \right)$ in~\eqref{EqJun4at9h02in2024} in terms of the functional~$\mathsf{G}$ in~\eqref{EqGZeta}, as the expectation of an algorithm-driven gap. The following lemma formalizes this first step. 

\begin{lemma}\label{LemmaJun4at9h32in2024}
Consider the generalization error~$\overline{\overline{\mathsf{G}}} \left(P_{\vect{\Theta} | \vect{Z}}, P_{\vect{Z}} \right)$ in~\eqref{EqJun4at9h02in2024} and assume that for all~$\vect{z}$, the probability measure~$P_{\vect{\Theta} | \vect{Z} = \vect{z}}$ is absolutely continuous with respect to a probability measure~$P_{\vect{\Theta}} \in \triangle\left( \set{M} \right)$ that satisfies for all measurable subsets~$\set{C}$ of~$\set{M}$, 
\begin{IEEEeqnarray}{rcl}
\label{EqJune26at16h54in2024}
P_{\vect{\Theta}} \left( \set{C} \right) & = &  \int P_{\vect{\Theta} | \vect{Z} = \vect{z}} \left( \set{C} \right) \mathrm{d} P_{\vect{Z}} \left( \vect{z}\right).
\end{IEEEeqnarray}
Then,
\begin{IEEEeqnarray}{rCl}
\label{EqJun4at9h33in2024}
\overline{\overline{\mathsf{G}}} \left(P_{\vect{\Theta} | \vect{Z}}, P_{\vect{Z}} \right) & = & \int \mathsf{G}\left( \vect{z}, P_{\vect{\Theta}} , P_{\vect{\Theta} | \vect{Z} = \vect{z}} \right)  \mathrm{d} P_{\vect{Z}} \left( \vect{z} \right),
\end{IEEEeqnarray}
where the functional~$\mathsf{G}$ is defined in~\eqref{EqGZeta}.
\end{lemma}
\begin{IEEEproof}
The proof is presented in Appendix~\ref{AppProofOfLemmaJun4at9h32in2024}.
\end{IEEEproof}
Given a training dataset~$\vect{z}$, as the one in~\eqref{EqTheDataSet}, a probability measure~$P_{\vect{\Theta} | \vect{Z} = \vect{z}}$ represents the specific instance of an algorithm~$P_{\vect{\Theta} | \vect{Z}}$ obtained by training such an algorithm on the training dataset~$\vect{z}$. 
From this perspective, each dataset~$\vect{z}$ in~$\left( \set{X} \times \set{Y} \right)^n$ generates an instance of the algorithm~$P_{\vect{\Theta}| \vect{Z}}$.
Hence, the measure $P_{\vect{\Theta}}$  in~\eqref{EqJune26at16h54in2024} can be understood as the barycenter of the algorithm~$P_{\vect{\Theta}| \vect{Z}}$. This is in the sense that the measure~$P_{\vect{\Theta}}$ is the convex combination of all possible instances of the algorithm~$P_{\vect{\Theta} | \vect{Z}}$, where the coefficients of such a combination are determined by the probability measure on the datasets~$P_{\vect{Z}}$.
Within this context, Lemma~\ref{LemmaJun4at9h32in2024} presents the generalization error~$\overline{\overline{\mathsf{G}}} \left(P_{\vect{\Theta} | \vect{Z}}, P_{\vect{Z}} \right)$ in~\eqref{EqJun4at9h02in2024} as the expectation (when~$\vect{z}$ is assumed to be drawn from~$P_{\vect{Z}}$) of the variations of the empirical risk~$\mathsf{R}_{\vect{z}}$ in~\eqref{EqRxy} when the instance~$P_{\vect{\Theta} | \vect{Z} = \vect{z}}$ of the algorithm, is replaced by the algorithm's barycenter~$P_{\vect{\Theta}}$. 

\subsubsection{Step Two}
The second step consists in exploiting the fact that algorithm-driven gaps exhibit closed-form expressions involving the conditional Gibbs probability measure studied in~\cite{daunas2024equivalence, daunas2025asymmetry, PerlazaTIT2024, Perlaza-ISIT2023a, InriaRR9508, InriaRR9521} and 
used in \cite{catoni2004statistical, zdeborova2016statistical, alquier2016properties,zhang2006ep,zhang2006information}. In order to formally describe this step, such a conditional Gibbs probability measure is introduced. 
Consider two parameters: 
$(a)$~A~$\sigma$-finite measure~$Q$ defined on $\set{M}$; and~$(b)$~A positive real~$\lambda$.
These parameters are often referred to as the reference measure; and the temperature coefficient.
A Gibbs probability measure, parametrized by~$Q$ and~$\lambda$, and conditioned on a training dataset~$\vect{z}$ in~$\left( \set{X}  \times \set{Y} \right)^n$, is denoted by~$P^{\left(Q, \lambda\right)}_{\vect{\Theta}| \vect{Z} = \vect{z}}  \in \triangle\left( \set{M} \right)$. 
This notation is chosen to explicitly highlight the parameters~$Q$,~$\lambda$, and the training dataset~$\vect{z}$.
Before introducing an explicit expression for such a measure, two mathematical objects of central importance are introduced. 
First, let the function~$K_{Q,\vect{z}}: \reals \rightarrow \reals \cup \lbrace +\infty\rbrace$ be  such that  
\begin{IEEEeqnarray}{rcl}
\label{EqK}
K_{Q,\vect{z}}\left(t \right) & = &  \log\left( \int \exp\left( t \; \mathsf{L}\left(\vect{z}, \vect{\theta}\right)  \right) \mathrm{d}Q(\vect{\theta}) \right),
\end{IEEEeqnarray} 
where the function~$\mathsf{L}$ is defined in~\eqref{EqLxy}.
The function~$K_{Q,\vect{z}}$, often referred to as the \emph{log-partition function} of the measure~$P^{\left(Q, \lambda\right)}_{\vect{\Theta}| \vect{Z} = \vect{z}}$, is nondecreasing, convex, and differentiable infinitely many times \cite[Lemma~$15$ and Lemma~$16$]{PerlazaTIT2024}. 
Let also the set~$\set{K}_{Q,\vect{z}} \subset (0, +\infty)$ be 
\begin{IEEEeqnarray}{rcl}
\label{EqSetKxy}
\set{K}_{Q,\vect{z}} & \triangleq &\left\lbrace s > 0: \; K_{Q,\vect{z}}\left(-\frac{1}{s} \right)  < +\infty \right\rbrace.
\end{IEEEeqnarray}
The set~$\set{K}_{Q,\vect{z}}$ is either empty or a convex interval including the interval~$\left(0, b \right)$, for some real~$b \in (0, +\infty]$. See for instance, \cite[Lemma~$1$]{PerlazaTIT2024}.  Interestingly, when~$Q$ is a probability measure, it follows that~$\set{K}_{Q,\vect{z}} = (0, +\infty)$.
Let also the set~$\set{K}_{Q}$ be
\begin{IEEEeqnarray}{rCl}
\label{EqSeptember3at18h10in2024}
\set{K}_{Q} & \triangleq & \displaystyle\bigcap_{\vect{z} \in \left( \set{X} \times \set{Y} \right)^n} \set{K}_{Q,\vect{z}}.
\end{IEEEeqnarray}
Using this notation, the Gibbs algorithm is defined as follows.
\begin{definition}[Gibbs Algorithm]\label{DefGibbsAlgorithm}
Let $Q$ be a $\sigma$-finite measure on $\set{M}$ and $\lambda$ be a fixed real in $\set{K}_{Q}$, with~$\set{K}_{Q}$ in~\eqref{EqSeptember3at18h10in2024}.  
An algorithm, denoted by~$P^{\left(Q, \lambda\right)}_{\vect{\Theta}| \vect{Z}} \in \triangle\left( \set{M} | \left( \set{X}  \times \set{Y} \right)^n \right)$, is said to be a Gibbs algorithm with parameters~$Q$ and~$\lambda$,  if the instance of such algorithm obtained by training it upon the dataset~$\vect{z}$ in~\eqref{EqTheDataSet}, denoted by~$P^{\left(Q, \lambda\right)}_{\vect{\Theta}| \vect{Z} = \vect{z}}  \in \triangle\left( \set{M} \right)$, satisfies for all~$\vect{\theta} \in \supp Q$,
\begin{IEEEeqnarray}{rcl}\label{EqGenpdf}%\label{EqRND}
\frac{\mathrm{d}P^{\left(Q, \lambda\right)}_{\vect{\Theta}| \vect{Z} = \vect{z}}}{\mathrm{d}Q} \left( \vect{\theta} \right) 
  & =& \exp\left( - K_{Q,\vect{z}}\left(- \frac{1}{\lambda} \right) - \frac{1}{\lambda} \mathsf{L}\left(\vect{z}, \vect{\theta}\right)\right),
\end{IEEEeqnarray}
where the nonnegative function~$\frac{\mathrm{d}P^{\left(Q, \lambda\right)}_{\vect{\Theta}| \vect{Z} = \vect{z}}}{\mathrm{d}Q}: \set{M} \to (0, +\infty)$ is the Radon-Nikodym derivative of the measure~$P^{\left(Q, \lambda\right)}_{\vect{\Theta}| \vect{Z} = \vect{z}}$ with respect to~$Q$; and  the functions~$\mathsf{L}$ and~$K_{Q,\vect{z}}$ are respectively defined in~\eqref{EqLxy} and~\eqref{EqK}.
\end{definition}
Using this notation, the second step of the method of gaps is formalized as follows.

\begin{lemma}\label{LemmaAugust3at11h10in2024}
Consider the generalization error~$\overline{\overline{\mathsf{G}}} \left(P_{\vect{\Theta} | \vect{Z}}, P_{\vect{Z}} \right)$ in~\eqref{EqJun4at9h02in2024}  and assume that for all~$\vect{z} \in \left( \set{X} \times \set{Y} \right)^n$, the  measures~$P_{\vect{\Theta} | \vect{Z} = \vect{z}}$, $P_{\vect{\Theta}}$ in~\eqref{EqJune26at16h54in2024}, and the~$\sigma$-finite measure~$Q$ in~\eqref{EqGenpdf} satisfy $P_{\vect{\Theta} | \vect{Z} = \vect{z}} \ll P_{\vect{\Theta}} \ll Q$.  Then,
\begin{IEEEeqnarray}{rCl}
\nonumber
& & \overline{\overline{\mathsf{G}}} \left(P_{\vect{\Theta} | \vect{Z}}, P_{\vect{Z}} \right) \\
\nonumber
& = & \lambda\int \bigg( \KL{P_{\vect{\Theta}}}{P^{\left(Q, \lambda\right)}_{\vect{\Theta}| \vect{Z} = \vect{z}}}  - \KL{P_{\vect{\Theta} | \vect{Z} = \vect{z}} }{P^{\left(Q, \lambda\right)}_{\vect{\Theta}| \vect{Z} = \vect{z}}}  
\\ 
\label{EqJune4at11h20in2024}
& & + \KL{P_{\vect{\Theta} | \vect{Z} = \vect{z}} }{Q} - \KL{P_{\vect{\Theta}}}{Q} \bigg) 
\mathrm{d} P_{\vect{Z}} \left( \vect{z} \right),
\end{IEEEeqnarray}
where the measure~$P^{\left(Q, \lambda\right)}_{\vect{\Theta}| \vect{Z} = \vect{z}} \in \triangle\left( \set{M}\right)$ is defined in~\eqref{EqGenpdf}; and~$\lambda \in \set{K}_{Q}$, with~$\set{K}_{Q}$ in~\eqref{EqSeptember3at18h10in2024}.
\end{lemma}

\begin{IEEEproof}
The proof follows immediately from Lemma~\ref{LemmaJun4at9h32in2024} and  Theorem~\ref{TheoremRGgeneral}, which is introduced later in Section~\ref{SecSensitivityArbitraryA}.
\end{IEEEproof}

Note that while the right-hand side of the equality in~\eqref{EqJune4at11h20in2024} depends on the measure~$Q$ and a real~$\lambda$, the left-hand side does not depend upon these parameters. 
This shows the paramount relevance of the equality in~\eqref{EqJune4at11h20in2024} as particular choices of~$Q$ and~$\lambda$ would yield different equivalent expressions for the generalization error~$\overline{\overline{\mathsf{G}}} \left(P_{\vect{\Theta} | \vect{Z}}, P_{\vect{Z}} \right)$ in~\eqref{EqJun4at9h02in2024}. 
Section~\ref{SecExpressionsViaAlgorithmDrivenGaps} explores some of these choices and presents some interesting explicit expressions for~$\overline{\overline{\mathsf{G}}} \left(P_{\vect{\Theta} | \vect{Z}}, P_{\vect{Z}} \right)$  emerging from~\eqref{EqJune4at11h20in2024}.

\subsection{Method of Data-driven Gaps}\label{SecCharDataDrivenGaps}

\subsubsection{Step One}
The first step of the \emph{method of data-driven gaps} consists in writing the generalization error~$\overline{\overline{\mathsf{G}}} \left(P_{\vect{\Theta} | \vect{Z}}, P_{\vect{Z}} \right)$ in~\eqref{EqJun4at9h02in2024} in terms of the functional~$\mathsf{G}$ in~\eqref{EqGTheta}, as the expectation of a data-driven gap. 
To achieve such an objective, the central observation is that the joint probability measure~$P_{\vect{\Theta} | \vect{Z}} P_{\vect{Z}} \in \triangle\left( \set{M} \times \left( \set{X} \times \set{Y} \right) ^n \right)$ shares common properties with the measure~$P_{\vect{Z} | \vect{\Theta}}  P_{\vect{\Theta}}  \in \triangle\left( \left( \set{X} \times \set{Y} \right) ^n \times \set{M}  \right)$, where~$P_{\vect{\Theta}}$ is in~\eqref{EqJune26at16h54in2024} and the conditional probability measure~$P_{\vect{Z} | \vect{\Theta}} \in \triangle\left(  \left( \set{X} \times \set{Y} \right)^n | \set{M}\right)$ satisfies for all measurable subsets~$\set{B}$ of~$\left( \set{X} \times \set{Y} \right)^{n}$,
\begin{IEEEeqnarray}{rCl}
	\label{EqSaturdayJan27in2024b}
	P_{\vect{Z}} \left( \set{B} \right) = \int P_{\vect{Z} | \vect{\Theta} = \vect{\theta}  } \left( \set{B} \right)  \mathrm{d}P_{\vect{\Theta}}\left( \vect{\theta} \right).
\end{IEEEeqnarray}
These common properties are discussed in detail in Appendix~\ref{AppendixMiscellanea}. 
In the following, it is assumed that the probability measure~$P_{\vect{Z}}$ in~\eqref{EqSaturdayJan27in2024b} is a product measure formed by the measure
\begin{IEEEeqnarray}{rCl}\label{EqGTDG}
	P_{Z} \in \triangle\left( \set{X} \times \set{Y}\right).
\end{IEEEeqnarray}
That is, the probability measures~$P_{\vect{Z}}$ and~$P_{Z}$ satisfy the equality in~\eqref{EqSaturdayMai20at13h20in2024}.
For all~$\vect{\theta} \in \set{M}$, the marginal measures of the probability measure~$P_{\vect{Z} | \vect{\Theta} = \vect{\theta}}$, denoted by~$P_{Z_t | \vect{\Theta} = \vect{\theta}} \in \triangle\left( \set{X} \times \set{Y} \right)$ with~$t \in \lbrace 1,2, \ldots, n \rbrace$, are probability measures satisfying the following conditions. 
Given~$m \in \lbrace 1, 2, \ldots, n \rbrace$ and a measurable subset $\set{A}\subseteq \left( \set{X} \times \set{Y} \right)$, let the set~$\set{B}_m \subseteq \left( \set{X} \times \set{Y} \right)^n$ be
\begin{IEEEeqnarray}{rcl}\label{EqNovember25at09h37in2025Sophia}
\set{B}_m & = & \set{A}_{1} \times \set{A}_2 \times \ldots \times \set{A}_m \times \ldots \times \set{A}_n,
\end{IEEEeqnarray}
 where for all~$t \in \lbrace 1, 2, \ldots, n \rbrace \setminus \lbrace m \rbrace$,~$\set{A}_t = \set{X} \times \set{Y}$; and~$\set{A}_m = \set{A}$.
The marginal probability measure~$P_{Z_m | \vect{\Theta} = \vect{\theta}}$ satisfies
\begin{IEEEeqnarray}{rcl}
\label{EqNovember6at10h55in2024SophiaAntipolis}
P_{Z_m | \vect{\Theta} = \vect{\theta}} \left( \set{A} \right) & = & P_{\vect{Z} | \vect{\Theta} = \vect{\theta}} \left( \set{B}_m \right).
\end{IEEEeqnarray}
Under the above assumptions on~$P_{\vect{Z}}$ and~$P_{Z}$, the marginal measures~$P_{Z_1 | \vect{\Theta} = \vect{\theta}}$,~$P_{Z_2 | \vect{\Theta} = \vect{\theta}}$,~$\ldots$,~$P_{Z_n | \vect{\Theta} = \vect{\theta}}$  turn out to be identical to some~$P_{Z | \vect{\Theta} = \vect{\theta}}$, as shown by the following lemma.
\begin{lemma}\label{LemmaNovember5at17h06in2024SophiaAntipolis}
Consider the conditional probability measure~$P_{\vect{Z} | \vect{\Theta}} \in \triangle\left( \left( 
\set{X} \times \set{Y}\right)^n | \set{M}\right)$ and the probability measures~$P_{\vect{\Theta}}$ and~$P_{\vect{Z}}$ in~\eqref{EqSaturdayJan27in2024b}. If~$P_{\vect{Z}}$ satisfies~\eqref{EqSaturdayMai20at13h20in2024} for some probability measure~$P_{Z} \in \triangle\left( \set{X} \times \set{Y}\right)$, then, for all measurable sets~$\set{A}$ of~$\left(\set{X} \times \set{Y}\right)$,  for all~$\vect{\theta}\in \set{M}$,  and for all~$m \in \lbrace 1,2, \ldots, n \rbrace$, the measure~$P_{Z_m | \vect{\Theta} = \vect{\theta}}$ in~\eqref{EqNovember6at10h55in2024SophiaAntipolis} satisfies,
	\begin{IEEEeqnarray}{rCl}
		\label{EqOctober17at15h01in2024}
P_{Z_m | \vect{\Theta} = \vect{\theta}} \left( \set{A}\right) & =& P_{Z | \vect{\Theta} = \vect{\theta}} \left( \set{A}\right), 
	\end{IEEEeqnarray}
for some probability measure~$P_{Z | \vect{\Theta} = \vect{\theta}} \in \triangle\left( \set{X} \times \set{Y} \right)$ that satisfies, 
\begin{IEEEeqnarray}{rCl}
	\label{EqOctober17at14h40in2024}
	P_{Z} \left( \set{A} \right)  & = &  \int P_{Z | \vect{\Theta} = \vect{\theta}} \left( \set{A}\right) \mathrm{d}P_{\vect{\Theta}}\left( \vect{\theta} \right). \spnum
\end{IEEEeqnarray}
\end{lemma}
\begin{IEEEproof}
Consider a fixed $m \inCountK{n}$ and the sets $\set{B}_m$ and $\set{A}_m = \set{A}$ in~\eqref{EqNovember25at09h37in2025Sophia}. Note that
\begin{IEEEeqnarray}{rcl}
\label{EqNovember25at09h46in2025Sophia}
P_{Z}\left( \set{A} \right) & =  & P_{Z}\left( \set{A}_m \right)\\
\label{EqNovember25at09h46in2025SophiaA}
& =  & P_{\vect{Z}}\left( \set{B}_m \right)\\
\label{EqNovember25at09h46in2025SophiaB}
& =  & \int P_{\vect{Z} | \vect{\Theta} = \vect{\theta}} \left( \set{B}_m \right) \mathrm{d}P_{\vect{\Theta}}\left( \vect{\theta} \right) \\
\label{EqNovember25at09h46in2025SophiaC}
& =  & \int P_{Z_m | \vect{\Theta} = \vect{\theta}} \left( \set{A} \right) \mathrm{d}P_{\vect{\Theta}}\left( \vect{\theta} \right),
\end{IEEEeqnarray}
where the equality in \eqref{EqNovember25at09h46in2025SophiaA} follows from the construction in~\eqref{EqNovember25at09h37in2025Sophia} and the assumption in~\eqref{EqSaturdayMai20at13h20in2024};
the equality in \eqref{EqNovember25at09h46in2025SophiaB} follows from~\eqref{EqSaturdayJan27in2024b}; and 
the equality in \eqref{EqNovember25at09h46in2025SophiaC} follows from~\eqref{EqNovember6at10h55in2024SophiaAntipolis}.
Note that the right-hand side of~\eqref{EqNovember25at09h46in2025SophiaC} depends on $m$, while the left-hand side does not. 
Hence, the proof focuses on showing that the conditional measures $P_{Z_1 | \vect{\Theta}}$, $P_{Z_2 | \vect{\Theta}}$, $\ldots$, $P_{Z_n | \vect{\Theta}}$ in~\eqref{EqNovember25at09h46in2025SophiaC} are identical, and thus, independent of the choice of $m$.
For doing so, given the set $\set{A}$ in~\eqref{EqNovember25at09h37in2025Sophia}, let the function $g_{\set{A}}: \set{M} \to [0,1]$ be
\begin{IEEEeqnarray}{rcl}
\label{EqDecember22at15h22in2025Sophia}
g_{\set{A}}\left( \vect{\theta} \right) & = & P_{Z_m | \vect{\Theta} = \vect{\theta}} \left( \set{A} \right),
\end{IEEEeqnarray}
and thus, Borel measurable.
From the equality in~\eqref{EqNovember25at09h46in2025SophiaC}, it follows that 
\begin{IEEEeqnarray}{rcl}
\label{EqNovember25at17h04in2025BusToNice}
P_{Z}\left( \set{A} \right) & =  & \int g_{\set{A}}\left( \vect{\theta} \right) \mathrm{d}P_{\vect{\Theta}}\left( \vect{\theta} \right).
\end{IEEEeqnarray}
Moreover, from \cite[Theorem~$5.3.1$]{ash2000probability}, such a function $g_{\set{A}}$ is unique almost surely with respect to~$P_{\vect{\Theta}}$. The proof is completed by noticing that the choice of $m$ and $\set{A}$ in~\eqref{EqDecember22at15h22in2025Sophia} are irrelevant and thus, the conditional measures $P_{Z_1 | \vect{\Theta}}$, $P_{Z_2 | \vect{\Theta}}$, $\ldots$, $P_{Z_n | \vect{\Theta}}$ are identical to a conditional measure $P_{Z | \vect{\Theta}}$ almost surely with respect to~$P_{\vect{\Theta}}$. That is, $g_{\set{A}}\left( \vect{\theta} \right) = P_{Z | \vect{\Theta} = \vect{\theta}} \left( \set{A} \right)$, with $P_{Z | \vect{\Theta}}$ in~\eqref{EqOctober17at15h01in2024}. 
\end{IEEEproof}
The following lemma formalizes the first step of the method of data-driven gaps. 
\begin{lemma}\label{LemmaOctober17at14h42in2024}
Consider the generalization error~$\overline{\overline{\mathsf{G}}}(P_{\vect{\Theta} | \vect{Z}}, P_{\vect{Z}} )$ in~\eqref{EqJun4at9h02in2024} and assume that 
\begin{itemize}
\item[$(a)$] The probability measure~$P_{\vect{Z}}$ satisfies~\eqref{EqSaturdayMai20at13h20in2024} for some~$P_{Z} \in \triangle\left( \set{X} \times \set{Y}\right)$; 
\item[$(b)$]  For all~$\vect{z} \in \left( \set{X} \times \set{Y} \right)^n$, the probability measures~$P_{\vect{\Theta} | \vect{Z} = \vect{z}}$ and~$P_{\vect{\Theta}}$ in~\eqref{EqJune26at16h54in2024} are mutually absolutely continuous; and
\item[$(c)$]  For all~$\vect{\theta} \in \set{M}$, the probability measures~$P_{\vect{Z} | \vect{\Theta} = \vect{\theta}}$ and~$P_{\vect{Z}}$ in~\eqref{EqSaturdayJan27in2024b} are mutually absolutely continuous.
\end{itemize}
Then,
	\begin{IEEEeqnarray}{rCl}
\label{EqOctober21at11h08in2024Agadir}	
		\overline{\overline{\mathsf{G}}}(P_{\vect{\Theta} | \vect{Z}},P_{\vect{Z}} )  & = &  \int \mathsf{G}\left( \vect{\theta}, P_{Z}, P_{Z | \vect{\Theta} = \vect{\theta}}\right) \mathrm{d} P_{\vect{\Theta}}\left( \vect{\theta}\right),
	\end{IEEEeqnarray}
	where the functional~$\mathsf{G}$ is defined in~\eqref{EqGTheta}; and the conditional probability measure~$P_{Z| \vect{\Theta}}$  satisfies~\eqref{EqOctober17at14h40in2024}.
\end{lemma}
\begin{IEEEproof}
	The proof is presented in Appendix~\ref{AppProofOfLemmaOctober17at14h42in2024}.	
\end{IEEEproof}
From a Bayesian perspective, if the algorithm~$P_{\vect{\Theta} | \vect{Z} = \vect{z}}$ is understood as the posterior of the prior~$P_{\vect{\Theta}}$ after the observation of the training dataset~$\vect{z}$, then, the probability measure~$P_{\vect{Z} | \vect{\Theta} = \vect{\theta}}$ in~\eqref{EqOctober17at15h01in2024} can be understood as the likelihood of the datasets after the observation of the model~$\vect{\theta}$.
From this perspective,  Lemma~\ref{LemmaOctober17at14h42in2024} presents the generalization error~$\overline{\overline{\mathsf{G}}} \left(P_{\vect{\Theta} | \vect{Z}}, P_{\vect{Z}} \right)$ in~\eqref{EqJun4at9h02in2024}  as the expectation (when~$\vect{\theta}$ is assumed to be drawn from~$P_{\vect{\Theta}}$ in~\eqref{EqJune26at16h54in2024}) of the variations of the empirical risk~$\mathsf{R}_{\vect{\theta}}$ in~\eqref{EqRModel} when the measure on the data points changes from  the likelihood~$P_{Z | \vect{\Theta} = \vect{\theta}}$ to the ground-truth probability distribution~$P_{Z}$, which is invariant with respect to the models. 

It is important to highlight the impact on Lemma~\ref{LemmaOctober17at14h42in2024} of the assumption in~\eqref{EqSaturdayMai20at13h20in2024}. In particular, note that the gap $\mathsf{G}\left( \vect{\theta}, P_{Z}, P_{Z | \vect{\Theta} = \vect{\theta}}\right)$ in \eqref{EqOctober21at11h08in2024Agadir} depends exclusively on the measure $P_Z$ (instead of the joint measure $P_{\vect{Z}}$) and the  marginal measure $P_{Z|\vect{\Theta} = \vect{\theta}}$ (instead of the joint measure $P_{\vect{Z}|\vect{\Theta}= \vect{\theta}}$). This simplification follows from Lemma~\ref{LemmaNovember5at17h06in2024SophiaAntipolis} and plays an important role allowing to involve the WCDG probability measure introduced in \cite{zou2024Generalization}.

\subsubsection{Step Two}
The second step consists in exploiting the fact that data-driven gaps exhibit closed-form expressions involving the WCDG probability measure introduced in \cite{zou2024Generalization} and further studied in \cite{zouJSAIT2024}.
The WCDG probability measure is conditioned on a model~$\vect{\theta} \in \set{M}$; and parametrized by a real~$\beta >0$ and a probability measure~$P_S$ in~$\triangle\left( \set{X} \times \set{Y} \right)$. Such a WCDG probability measure is denoted by~$P_{\hat{Z} | \vect{\Theta} = \vect{\theta}}^{\left( P_{S}, \beta \right)} \in \triangle_{P_S}\left( \set{X} \times \set{Y} \right)$. 
Defining an explicit expression for~$P_{\hat{Z} | \vect{\Theta} = \vect{\theta}}^{\left( P_{S}, \beta \right)}$, requires defining the following mathematical objects.     
Let~$\mathsf{J}_{P_S, \vect{\theta}}: \mathbb{R} \rightarrow \mathbb{R}$ be the function
\begin{IEEEeqnarray}{rCl}
	\label{EqOctober17at21h00in2024}
	\mathsf{J}_{P_S, \vect{\theta}}(t) = \log{\left( \int \exp{\left(  t \ell(x,y,\vect{\theta})\right)} \mathrm{d}P_S(x,y) \right)},
\end{IEEEeqnarray}
where the function~$\ell$ is defined in~\eqref{EqEll}. 
The function~$\mathsf{J}_{P_S, \vect{\theta}}$ in~\eqref{EqOctober17at21h00in2024} is convex, nondecreasing, and differentiable infinitely many times in the interior of~$\{t \in\reals: \mathsf{J}_{P_S, \vect{\theta}}\left(t \right)< +\infty\}$ \cite[Lemma~$1$]{zouJSAIT2024}.
Let also the sets~$\set{J}_{P_S, \vect{\theta}} \subseteq \left( 0, +\infty \right)$ and~$\set{J}_{P_S} \subseteq \left( 0, +\infty \right)$ be
\begin{IEEEeqnarray}{rcl}
	\label{EqSetCDFl}
	\set{J}_{P_S, \vect{\theta}} & \triangleq& \left\lbrace t \in (0,+\infty):  \mathsf{J}_{P_S, \vect{\theta}} \left(\frac{1}{t} \right)  < +\infty  \right\rbrace,
\end{IEEEeqnarray}
and
\begin{IEEEeqnarray}{rCl}
	\label{EqOctober17at21h21in2024}
	\set{J}_{P_S} & \triangleq & \displaystyle\bigcap_{\vect{\theta} \in \set{M}} \set{J}_{P_S,\vect{\theta}}.
\end{IEEEeqnarray}
The set~$\set{J}_{P_S, \vect{\theta}}$ is either empty or an interval including the interval~$\left(b, +\infty \right)$, for some real~$b \in (0, +\infty]$. See for instance, \cite[Lemma~$2$]{zouJSAIT2024}.
Using this notation, the WCDG conditional probability measure is defined as follows.
\begin{definition}[The WCDG conditional probability measure]\label{DefWCDG}
A conditional probability measure, denoted by~$P_{\hat{Z} | \vect{\Theta}}^{\left( P_{S}, \beta \right)} \in \triangle\left( \set{X} \times \set{Y} | \set{M} \right)$, is said to be a WCDG conditional probability measure with parameters~$P_S \in \triangle\left( \set{X} \times \set{Y} \right)$ and~$\beta \in \set{J}_{P_S}$, with~$\set{J}_{P_S}$ in~\eqref{EqOctober17at21h21in2024}, if for all~$\vect{\theta} \in \set{M}$ and  for all~$(x,y) \in \supp P_S$,
\begin{IEEEeqnarray}{rCl}
	\label{Eqtheworstgeneral}
	\frac{\mathrm{d}P_{\hat{Z} | \vect{\Theta} = \vect{\theta}}^{\left( P_{S}, \beta \right)} }{\mathrm{d}P_S} (x,y) &&=\exp{\left(\frac{1}{\beta} \ell\left( x,y, \vect{\theta}\right) - \mathsf{J}_{P_S, \vect{\theta}}\left(\frac{1}{\beta}\right)\right)},\IEEEeqnarraynumspace
\end{IEEEeqnarray}
where the function~$\frac{\mathrm{d}P_{\hat{Z} | \vect{\Theta} = \vect{\theta}}^{\left( P_{S}, \beta \right)} }{\mathrm{d}P_S}: \set{X} \times \set{Y} \to \left(0, +\infty \right)$ is the Radon-Nikodym derivative of the measure~$P_{\hat{Z} | \vect{\Theta} = \vect{\theta}}^{\left( P_{S}, \beta \right)}$ with respect to~$P_S$;  and
the functions~$\ell$ and~$\mathsf{J}_{P_S, \vect{\theta}}$ are defined in~\eqref{EqEll} and~\eqref{EqOctober17at21h00in2024}.
\end{definition}
Using this notation, the second step of the method of data-driven gaps is formalized as follows.

\begin{lemma}\label{LemmaGinD}  
Consider the generalization error~$\overline{\overline{\mathsf{G}}}(P_{\vect{\Theta} | \vect{Z}}, P_{\vect{Z}} )$ in~\eqref{EqJun4at9h02in2024} and assume that: 
\begin{itemize}
\item[$(a)$] The probability measure~$P_{\vect{Z}}$ satisfies~\eqref{EqSaturdayMai20at13h20in2024} for some~$P_{Z} \in \triangle\left( \set{X} \times \set{Y}\right)$; 
\item[$(b)$]  For all~$\vect{z} \in \left( \set{X} \times \set{Y} \right)^n$, the probability measures~$P_{\vect{\Theta} | \vect{Z} = \vect{z}}$ and~$P_{\vect{\Theta}}$ in~\eqref{EqJune26at16h54in2024} are mutually absolutely continuous; 
\item[$(c)$]  For all~$\vect{\theta} \in \set{M}$, the probability measures~$P_{\vect{Z} | \vect{\Theta} = \vect{\theta}}$ and~$P_{\vect{Z}}$ in~\eqref{EqSaturdayJan27in2024b} are mutually absolutely continuous; and 
\item[$(d)$] For all~$\vect{\theta} \in \set{M}$,  the probability measures~$P_{Z | \vect{\Theta} = \vect{\theta}}$ and~$P_{Z}$  satisfy~\eqref{EqOctober17at14h40in2024} and are both absolutely continuous with the probability measure~$P_S \in \triangle\left( \set{X} \times \set{Y} \right)$ in~\eqref{Eqtheworstgeneral}. 
\end{itemize}
Then, 
\begin{IEEEeqnarray}{rCl}
		\nonumber
& &\overline{\overline{\mathsf{G}}}(P_{\vect{\Theta} | \vect{Z}}, P_{\vect{Z}} )  
		= \beta \int \Bigg( \KL{P_{Z | \vect{\Theta} = \vect{\theta}}}{P_{\hat{Z} | \vect{\Theta} = \vect{\theta}}^{\left( P_{S}, \beta \right)} }-\KL{P_{Z}}{P_{\hat{Z} | \vect{\Theta} = \vect{\theta}}^{\left( P_{S}, \beta \right)} }\Tsupersqueezeequ \\
\label{EqOctober19at18h32in2024InTheAirFlyingToAgadir}
		& & - \KL{P_{Z | \vect{\Theta} = \vect{\theta}}}{P_S} + \KL{P_Z}{P_S}  \Bigg) \mathrm{d} P_{\vect{\Theta}}\left( \vect{\theta} \right), \squeezeequ \IEEEeqnarraynumspace
	\end{IEEEeqnarray}
	where the probability measure~$P_{\hat{Z} | \vect{\Theta} = \vect{\theta}}^{\left( P_{S}, \beta \right)}~$ is defined in~\eqref{Eqtheworstgeneral}; and~$\beta \in \set{J}_{P_S}$, with~$\set{J}_{P_S}$ in~\eqref{EqOctober17at21h21in2024}.

\end{lemma}
\begin{IEEEproof}
Under Assumptions~$(a)$ and~$(c)$, Lemma~\ref{EqNovember15at10h08in2024SophiaAntipolis} and Lemma~\ref{EqNovember13at17h04in2024InTheBusToNice} (in Appendix~\ref{AppendixMiscellanea}) imply that for all~$\vect{\theta} \in \set{M}$,  the probability measures~$P_{Z | \vect{\Theta} = \vect{\theta}}$ and~$P_{Z}$   are mutually absolutely continuous.
Hence, the proof follows immediately from Lemma~\ref{LemmaOctober17at14h42in2024} and Theorem~\ref{theoremgengapexpression}, which is introduced later in Section~\ref{SecDataDrivenGaps}.
\end{IEEEproof}

%%%%%
Note that while the right-hand side of the equality in~\eqref{EqOctober19at18h32in2024InTheAirFlyingToAgadir} depends on the probability measure~$P_S$ and~$\beta$, the left-hand side does not depend upon these parameters. 
This shows the paramount relevance of the equality in~\eqref{EqOctober19at18h32in2024InTheAirFlyingToAgadir} as particular choices of~$P_S$ and~$\beta$ would yield different equivalent expressions for the generalization error~$\overline{\overline{\mathsf{G}}} \left(P_{\vect{\Theta} | \vect{Z}}, P_{\vect{Z}} \right)$ in~\eqref{EqJun4at9h02in2024}. 

The proof of Lemma~\ref{LemmaGinD} uses Lemma~\ref{LemmaNovember5at17h06in2024SophiaAntipolis} and thus, it is written in terms of the marginal measure $P_Z$ and the marginal conditional measure $P_{Z|\vect{\Theta}}$ instead of the joint probability measure $P_{\vect{Z}}$ and the joint conditional probability measure $P_{\vect{Z}|\vect{\Theta}}$. This is the key step that allows using the WCDG probability measure $P_{\hat{Z} | \vect{\Theta}}^{\left( P_{S}, \beta \right)}$ defined in~\eqref{Eqtheworstgeneral} for obtaining~\eqref{EqOctober19at18h32in2024InTheAirFlyingToAgadir}.
Section~\ref{SecExpressionsViaDataDrivenGaps} explores the equality in \eqref{EqOctober19at18h32in2024InTheAirFlyingToAgadir} and presents new explicit expressions for~$\overline{\overline{\mathsf{G}}} \left(P_{\vect{\Theta} | \vect{Z}}, P_{\vect{Z}} \right)$ in~\eqref{EqJun4at9h02in2024}. 
 
\section{Explicit Expressions Obtained Via the Method of Algorithm-Driven Gaps}\label{SecExpressionsViaAlgorithmDrivenGaps}

This section focuses on constructing novel explicit expressions for the generalization error~$\overline{\overline{\mathsf{G}}} \left(P_{\vect{\Theta} | \vect{Z}}, P_{\vect{Z}} \right)$ in~\eqref{EqJun4at9h02in2024} by algebraically manipulating the expression in~\eqref{EqJune4at11h20in2024}.
In these expressions, the central mathematical object is the  Gibbs probability measure~$P^{\left(Q, \lambda\right)}_{\vect{\Theta}| \vect{Z}}$ (Definition~\ref{DefGibbsAlgorithm}). From this perspective, these expressions are obtained, at least in part,  by comparing the algorithm~$P_{\vect{\Theta} | \vect{Z}}$ with the Gibbs algorithm~$P^{\left(Q, \lambda\right)}_{\vect{\Theta}| \vect{Z}}$.  
Hence, these new expressions should not be interpreted as computational formulas. In general, they do not make the numerical evaluation of the generalization error~$\overline{\overline{\mathsf{G}}} \left(P_{\vect{\Theta} | \vect{Z}}, P_{\vect{Z}} \right)$ simpler than its original definition. Instead, these expressions shall be studied within the context of hypothesis testing; information measures; and Euclidean geometry.
For this purpose, this section is divided in four subsections. The first three subsections respectively  provide explicit expressions aiming at establishing connections to existing results in statistics (statistical hypothesis testing), information theory (information measures), and Euclidean geometry (Pythagorean Theorems).  
The last subsection reviews the generalization error for the special case of the Gibbs algorithm. 

\subsection{Connections to Statistical Hypothesis Testing}\label{SecDecember22at16h52in2025BusToNice}

The equality in~\eqref{EqJune4at11h20in2024} can be alternatively expressed as shown by the following theorem.

\begin{theorem}\label{TheoAugust9at11h15in2024}
Consider the generalization error~$\overline{\overline{\mathsf{G}}} \left(P_{\vect{\Theta} | \vect{Z}}, P_{\vect{Z}} \right)$ in~\eqref{EqJun4at9h02in2024} and assume that for all~$\vect{z} \in \left( \set{X} \times \set{Y} \right)^n$, the  measures~$P_{\vect{\Theta} | \vect{Z} = \vect{z}}$, $P_{\vect{\Theta}}$ in~\eqref{EqJune26at16h54in2024}, and the~$\sigma$-finite measure~$Q$ in~\eqref{EqGenpdf} satisfy $Q \ll P_{\vect{\Theta} | \vect{Z} = \vect{z}} \ll P_{\vect{\Theta}} \ll Q$. 
Then,
\begin{IEEEeqnarray}{rCl}
\nonumber
& & \overline{\overline{\mathsf{G}}} \left(P_{\vect{\Theta} | \vect{Z}}, P_{\vect{Z}} \right) \\
\nonumber
& = & \lambda \iint \left( \log \frac{\mathrm{d} P^{\left(Q, \lambda\right)}_{\vect{\Theta}| \vect{Z} = \vect{z}}}{\mathrm{d} Q}\left( \vect{\theta} \right)  \right) \mathrm{d}P_{\vect{\Theta} | \vect{Z} = \vect{z}} \left( \vect{\theta} \right) \mathrm{d}P_{\vect{Z}} \left( \vect{z} \right) \\
\label{EqAugust9at11h44in2024}
& & - \lambda \iint \left( \log \frac{\mathrm{d} P^{\left(Q, \lambda\right)}_{\vect{\Theta}| \vect{Z} = \vect{z}}}{\mathrm{d} Q}\left( \vect{\theta} \right)  \right) \mathrm{d}P_{\vect{\Theta}}\left( \vect{\theta} \right) 
\mathrm{d} P_{\vect{Z}} \left( \vect{z} \right),\IEEEeqnarraynumspace
\end{IEEEeqnarray}
where the Radon-Nikodym derivative~$\frac{\mathrm{d} P^{\left(Q, \lambda\right)}_{\vect{\Theta}| \vect{Z} = \vect{z}}}{\mathrm{d} Q}$ is defined in~\eqref{EqGenpdf}; and~$\lambda \in \set{K}_{Q}$, with~$\set{K}_{Q}$ in~\eqref{EqSeptember3at18h10in2024}.
\end{theorem}

\begin{IEEEproof}
Given the assumptions, Lemma~\ref{LemmaAugust3at11h10in2024} holds.
The proof follows by observing that the differences~$\KL{P_{\vect{\Theta}}}{P^{\left(Q, \lambda\right)}_{\vect{\Theta}| \vect{Z} = \vect{z}}} - \KL{P_{\vect{\Theta}}}{Q}$ and~$ \KL{P_{\vect{\Theta} | \vect{Z} = \vect{z}} }{Q} - \KL{P_{\vect{\Theta} | \vect{Z} = \vect{z}} }{P^{\left(Q, \lambda\right)}_{\vect{\Theta}| \vect{Z} = \vect{z}}}$ in~\eqref{EqJune4at11h20in2024} respectively satisfy the following equalities:
\begin{IEEEeqnarray}{rCl}
\nonumber
& &\KL{P_{\vect{\Theta}}}{P^{\left(Q, \lambda\right)}_{\vect{\Theta}| \vect{Z} = \vect{z}}} - \KL{P_{\vect{\Theta}}}{Q} \\
& = & \int \left( \log \frac{\mathrm{d} P_{\vect{\Theta}}}{\mathrm{d} P^{\left(Q, \lambda\right)}_{\vect{\Theta}| \vect{Z} = \vect{z}}}\left( \vect{\theta} \right) - \log\frac{\mathrm{d} P_{\vect{\Theta}}}{\mathrm{d} Q}\left( \vect{\theta} \right)  \right) \mathrm{d}P_{\vect{\Theta}}\left( \vect{\theta} \right)\\
\label{EqAugust9at11h30in2024A}
& = & - \int \left( \log \frac{\mathrm{d} P^{\left(Q, \lambda\right)}_{\vect{\Theta}| \vect{Z} = \vect{z}}}{\mathrm{d} Q}\left( \vect{\theta} \right)  \right) \mathrm{d}P_{\vect{\Theta}}\left( \vect{\theta} \right),
\end{IEEEeqnarray}
where the equality in~\eqref{EqAugust9at11h30in2024A} follows from the fact that the measures~$P^{\left(Q, \lambda\right)}_{\vect{\Theta}| \vect{Z} = \vect{z}}$ and~$Q$ are mutually absolutely continuous \cite[Lemma~$3$]{PerlazaTIT2024} and \cite[Theorem 4]{InriaRR9591}.
Using the same arguments, 
\begin{IEEEeqnarray}{rCl}
\nonumber
& & \KL{P_{\vect{\Theta} | \vect{Z} = \vect{z}} }{Q} - \KL{P_{\vect{\Theta} | \vect{Z} = \vect{z}} }{P^{\left(Q, \lambda\right)}_{\vect{\Theta}| \vect{Z} = \vect{z}}}\\
\label{EqAugust9at11h30in2024B}
& = & \int \left( \log \frac{\mathrm{d} P^{\left(Q, \lambda\right)}_{\vect{\Theta}| \vect{Z} = \vect{z}}}{\mathrm{d} Q}\left( \vect{\theta} \right)  \right) \mathrm{d}P_{\vect{\Theta} | \vect{Z} = \vect{z}} \left( \vect{\theta} \right).
\end{IEEEeqnarray}
Plugging both~\eqref{EqAugust9at11h30in2024A} and~\eqref{EqAugust9at11h30in2024B} into~\eqref{EqJune4at11h20in2024} yields~\eqref{EqAugust9at11h44in2024}, which completes the proof. 
\end{IEEEproof}
When $Q$ is a probability measure, i.e., $Q \in \simplex{\set{M}}$, Theorem~\ref{TheoAugust9at11h15in2024} can be framed   within the context of a mismatched hypothesis test. See~\cite{boroumand2022mismatched} and~\cite{pmlr-v119-dutta20a}.
Consider a hypothesis test in which the objective is to determine whether an observation~$\left( \vect{\theta}, \vect{z} \right) \in \set{M} \times \left( \set{X} \times \set{Y} \right)^n$ is generated  from either a probability distribution~$P^{\left(Q, \lambda\right)}_{\vect{\Theta}| \vect{Z}} P_{\vect{Z}} \in \triangle\left( \set{M} \times \left( \set{X} \times \set{Y}\right)^n\right)$ ($H_0$, null hypothesis) or from the alternative distribution~$Q  P_{\vect{Z}} \in \triangle\left( \set{M} \times \left( \set{X} \times \set{Y}\right)^n\right)$ ($H_1$, alternative hypothesis): 
\begin{IEEEeqnarray}{c}
\label{EqNovember28at14h44in2025Nice}
\left\lbrace
\begin{array}{rcl}
H_0: & & \left( \vect{\theta}, \vect{z} \right) \sim P^{\left(Q, \lambda\right)}_{\vect{\Theta}| \vect{Z}}  P_{\vect{Z}}\\
H_1: & & \left( \vect{\theta}, \vect{z} \right) \sim Q  P_{\vect{Z}}.
\end{array}
\right.
\end{IEEEeqnarray}
Such a hypothesis test is said to be mismatched because none of the hypotheses matches the assumptions on the probability distribution of the observations.  
More specifically, the hypothesis test in~\eqref{EqNovember28at14h44in2025Nice} is studied twice under different assumptions. First, the observation~$\left( \vect{\theta}, \vect{z} \right)$ is assumed to be sampled from the joint probability measure $P_{\vect{\Theta} | \vect{Z}} P_{\vect{Z}}$. Second,   the observation~$\left( \vect{\theta}, \vect{z} \right)$ is assumed to be sampled from the product of the marginals $P_{\vect{\Theta}} P_{\vect{Z}}$.
The first case corresponds to the scenario in which the pairs $\left( \vect{\theta}, \vect{z}\right)$ are such that $\vect{z}$ is sampled from $P_{\vect{Z}}$ and is the training dataset of the algorithm $P_{\vect{\Theta} | \vect{Z}}$ under study. At the same time, the model $\vect{\theta}$ is an output of such an instance of the algorithm, i.e., the model $\vect{\theta}$ is sampled from $P_{\vect{\Theta} | \vect{Z} = \vect{z}}$.
The second case corresponds to the scenario in which models are independent of the datasets.  
Regardless of such scenarios, the log-likelihood ratio to decide based upon the observation~$\left( \vect{\theta},\vect{z} \right)$ on the null or alternative hypothesis in \eqref{EqNovember28at14h44in2025Nice} is~$\log \frac{\mathrm{d} P^{\left(Q, \lambda\right)}_{\vect{\Theta}| \vect{Z}}P_{\vect{Z}}}{\mathrm{d} Q P_{\vect{Z}}}\left( \vect{\theta}, \vect{z} \right)$. Essentially, given a~$\gamma > 0$, if~$\log \frac{\mathrm{d} P^{\left(Q, \lambda\right)}_{\vect{\Theta}| \vect{Z}}P_{\vect{Z}}}{\mathrm{d} Q P_{\vect{Z}}}\left( \vect{\theta} , \vect{z}\right) \geqslant \gamma$, then the null hypothesis is accepted. 
Alternatively, the null hypothesis is rejected if~$\log \frac{\mathrm{d} P^{\left(Q, \lambda\right)}_{\vect{\Theta}| \vect{Z}}P_{\vect{Z}}}{\mathrm{d} Q P_{\vect{Z}}}\left( \vect{\theta} , \vect{z}\right) < \gamma$.
Interestingly, the term~$\frac{\mathrm{d} P^{\left(Q, \lambda\right)}_{\vect{\Theta}| \vect{Z}}P_{\vect{Z}}}{\mathrm{d} Q P_{\vect{Z}}}\left( \vect{\theta} , \vect{z}\right)$ satisfies
\begin{IEEEeqnarray}{rcl}
\label{EqNovember28in2025at21h48HomeNice}
 \frac{\mathrm{d} P^{\left(Q, \lambda\right)}_{\vect{\Theta}| \vect{Z}}P_{\vect{Z}}}{\mathrm{d} Q P_{\vect{Z}}}\left( \vect{\theta} , \vect{z} \right)& = & \frac{\mathrm{d} P^{\left(Q, \lambda\right)}_{\vect{\Theta}| \vect{Z}}P_{\vect{Z}}}{\mathrm{d} P^{\left(Q, \lambda\right)}_{\vect{\Theta}} P_{\vect{Z}}}\left( \vect{\theta} , \vect{z} \right)  \frac{\mathrm{d} P^{\left(Q, \lambda\right)}_{\vect{\Theta}}P_{\vect{Z}}}{\mathrm{d} Q P_{\vect{Z}}}\left( \vect{\theta} , \vect{z} \right) \spnum \\
\label{EqNovember28in2025at21h48HomeNiceA}
& = & \frac{\mathrm{d} P^{\left(Q, \lambda\right)}_{\vect{\Theta}| \vect{Z} = \vect{z}}}{\mathrm{d} P^{\left(Q, \lambda\right)}_{\vect{\Theta}}}\left( \vect{\theta} \right)   \frac{\mathrm{d} P^{\left(Q, \lambda\right)}_{\vect{\Theta}}P_{\vect{Z}}}{\mathrm{d} Q P_{\vect{Z}}}\left( \vect{\theta} , \vect{z} \right) \\
\label{EqNovember28in2025at21h48HomeNiceB}
& = & \frac{\mathrm{d} P^{\left(Q, \lambda\right)}_{\vect{\Theta}| \vect{Z} = \vect{z}}}{\mathrm{d} P^{\left(Q, \lambda\right)}_{\vect{\Theta}}}\left( \vect{\theta} \right)   \frac{\mathrm{d} P^{\left(Q, \lambda\right)}_{\vect{\Theta}}}{\mathrm{d} Q }\left( \vect{\theta}  \right)  \\
\label{EqNovember28in2025at21h48HomeNiceC}
& = & \frac{\mathrm{d} P^{\left(Q, \lambda\right)}_{\vect{\Theta}| \vect{Z} = \vect{z}}}{\mathrm{d} Q}\left( \vect{\theta} \right) ,
\end{IEEEeqnarray}
where the probability measure $P^{\left(Q, \lambda\right)}_{\vect{\Theta}} \in \simplex{\set{M}}$ in~\eqref{EqNovember28in2025at21h48HomeNice} is the marginal in $\simplex{\set{M}}$ of the joint probability measure $P^{\left(Q, \lambda\right)}_{\vect{\Theta}| \vect{Z}}P_{\vect{Z}}$. That is, for all measurable sets $\set{A}$ in $\set{M}$,
\begin{IEEEeqnarray}{rcl}
P^{\left(Q, \lambda\right)}_{\vect{\Theta} } \left( \set{A} \right) & = & \int P^{\left(Q, \lambda\right)}_{\vect{\Theta}| \vect{Z} = \vect{z}} \left( \set{A} \right) \mathrm{d} P_{\vect{Z}} \left( \vect{z} \right) .
\end{IEEEeqnarray}
The equality in~\eqref{EqNovember28in2025at21h48HomeNice} follows from \cite[Theorem~4]{InriaRR9591};
the equality in~\eqref{EqNovember28in2025at21h48HomeNiceA} follows from \cite[Theorem~11]{InriaRR9591};
the equality in~\eqref{EqNovember28in2025at21h48HomeNiceB} follows from \cite[Theorem~8]{InriaRR9591}; and
the equality in~\eqref{EqNovember28in2025at21h48HomeNiceC} follows from \cite[Theorem~4]{InriaRR9591}.

The equality in~\eqref{EqNovember28in2025at21h48HomeNiceC}, allows re-stating the hypotheses $H_0$ and $H_1$ in~\eqref{EqNovember28at14h44in2025Nice} as follows:
\begin{IEEEeqnarray}{c}
\label{EqDecember1at16h48in2025BusToNice}
\left\lbrace
\begin{array}{rcl}
H_0: & &  \vect{\theta}  \sim P^{\left(Q, \lambda\right)}_{\vect{\Theta}| \vect{Z} = \vect{z}}\\
H_1: & &  \vect{\theta}  \sim Q.
\end{array}
\right.
\end{IEEEeqnarray}
This is essentially because both~\eqref{EqNovember28at14h44in2025Nice} and~\eqref{EqDecember1at16h48in2025BusToNice} exhibit the same log-likelihood  ratio.
In~\eqref{EqDecember1at16h48in2025BusToNice}, the objective is to determine whether the model~$\vect{\theta}$ is sampled from a Gibbs algorithm whose reference measure is $Q$ and has been trained upon the dataset~$\vect{z}$ ($H_0$ null hypothesis); or from the reference measure $Q$ ($H_1$ alternative hypothesis), which is invariant with respect to the dataset $\vect{z}$.
Interestingly, both measures, $P^{\left(Q, \lambda\right)}_{\vect{\Theta}| \vect{Z}}$ and $Q$ are exogenous to the algorithm under study $P_{\vect{\Theta} | \vect{Z}}$. More importantly, both $Q$ and $\lambda$ can be arbitrarily chosen subject to the conditions of the theorem. 
From this perspective, the connection between the hypothesis test in~\eqref{EqDecember1at16h48in2025BusToNice} and the generalization error in~\eqref{EqJun4at9h02in2024} appears via the assumptions on the observations $\left( \vect{\theta}, \vect{z} \right)$. 
In this context, the role of the parameter~$\gamma$ mentioned above is central.  
More specifically, consider the set
\begin{IEEEeqnarray}{rcl}
\label{EqNovember20at13h44in2025HomeNice}
\set{A}_{\gamma} & = &\left\lbrace  \left( \vect{\theta}, \vect{z} \right) \in \set{M} \times \left( \set{X} \times \set{Y} \right)^n : \log \frac{\mathrm{d} P^{\left(Q, \lambda\right)}_{\vect{\Theta}| \vect{Z} = \vect{z}}}{\mathrm{d} Q}\left( \vect{\theta} \right) \geqslant \gamma \right\rbrace, \squeezeequ \IEEEeqnarraynumspace
\end{IEEEeqnarray}
which is the acceptance region of the null hypothesis. That is, if a given observation $\left( \vect{\theta}, \vect{z} \right)$ satisfies $\left( \vect{\theta}, \vect{z} \right) \in \set{A}_{\gamma}$, then the null hypothesis $H_0$ in \eqref{EqNovember28at14h44in2025Nice} is adopted. 
Consider also the following probabilities
\begin{IEEEeqnarray}{rcl}
\label{EqNovember30at13h38in2025HomeNiceA}
P_{\vect{\Theta} | \vect{Z} } P_{\vect{Z}} \left( \set{A}_\gamma \right) & = & 1 - P_{\vect{\Theta} | \vect{Z} } P_{\vect{Z}} \left( \set{A}_\gamma^{\sfc} \right), \mbox{ and }\\
\label{EqNovember30at13h38in2025HomeNiceB}
P_{\vect{\Theta}} P_{\vect{Z}} \left( \set{A}_\gamma \right) & = & 1 - P_{\vect{\Theta} } P_{\vect{Z}} \left( \set{A}_\gamma^{\sfc} \right).
\end{IEEEeqnarray}
The term $P_{\vect{\Theta} | \vect{Z} } P_{\vect{Z}} \left( \set{A}_\gamma \right)$ represents the probability of choosing $H_0$ under the assumption that observations $\left( \vect{\theta}, \vect{z} \right)$ are obtained by sampling $P_{\vect{\Theta} | \vect{Z}} P_{\vect{Z}}$.
Alternatively, the term $P_{\vect{\Theta}} P_{\vect{Z}} \left( \set{A}_\gamma \right)$ represents the probability of choosing $H_0$ under the assumption that observations  $\left( \vect{\theta}, \vect{z} \right)$ are sampled from $P_{\vect{\Theta}}P_{\vect{Z}}$. 

The connection between the generalization error~$\overline{\overline{\mathsf{G}}} \left(P_{\vect{\Theta} | \vect{Z}}, P_{\vect{Z}} \right)$ in~\eqref{EqJun4at9h02in2024} and the hypothesis test in \eqref{EqNovember28at14h44in2025Nice} becomes more evident thanks to the following theorem, which uses the following parameters:
\begin{IEEEeqnarray}{rcl}
\label{EqNovember30at13h35in2025HomeNiceA}
\bar{\gamma} & \triangleq & \sup \left\lbrace \log \frac{\mathrm{d} P^{\left(Q, \lambda\right)}_{\vect{\Theta}| \vect{Z} = \vect{z}}}{\mathrm{d} Q}\left( \vect{\theta} \right) \in \reals  : \left( \vect{\theta}, \vect{z} \right)  \in \set{M} \times \left( \set{X} \times \set{Y} \right)^n \right\rbrace \squeezeequ\spnum
\end{IEEEeqnarray}
and 
\begin{IEEEeqnarray}{rcl}
\label{EqNovember30at13h35in2025HomeNiceB}
\underline{\gamma} & \triangleq & \inf \left\lbrace \log \frac{\mathrm{d} P^{\left(Q, \lambda\right)}_{\vect{\Theta}| \vect{Z} = \vect{z}}}{\mathrm{d} Q}\left( \vect{\theta} \right) \in \reals  : \left( \vect{\theta}, \vect{z} \right)  \in \set{M} \times \left( \set{X} \times \set{Y} \right)^n \right\rbrace. \squeezeequ\spnum
\end{IEEEeqnarray}
\begin{theorem}\label{TheoNovember29at13h20in2025HomeNice}
Consider the generalization error~$\overline{\overline{\mathsf{G}}} \left(P_{\vect{\Theta} | \vect{Z}}, P_{\vect{Z}} \right)$ in~\eqref{EqJun4at9h02in2024} and assume that for all~$\vect{z} \in \left( \set{X} \times \set{Y} \right)^n$, the  measures~$P_{\vect{\Theta} | \vect{Z} = \vect{z}}$, $P_{\vect{\Theta}}$ in~\eqref{EqJune26at16h54in2024}, and the~$\sigma$-finite measure~$Q$ in~\eqref{EqGenpdf} satisfy $Q \ll P_{\vect{\Theta} | \vect{Z} = \vect{z}} \ll P_{\vect{\Theta}} \ll Q$. Then,
\begin{IEEEeqnarray}{C}
\nonumber
\lambda \left(  \underline{\gamma}  - \gamma \right) P_{\vect{\Theta} | \vect{Z} }P_{\vect{Z}} \left( \set{A}_{\gamma}^{\sfc} \right)  + \lambda \left( \gamma - \bar{\gamma} \right) P_{\vect{\Theta}  }P_{\vect{Z}} \left( \set{A}_{\gamma} \right)\\ 
\nonumber
 \leqslant \overline{\overline{\mathsf{G}}} \left(P_{\vect{\Theta} | \vect{Z}}, P_{\vect{Z}} \right) \leqslant\\
\label{EqDecember1at06h31in2025BusToSophia}
  \lambda \left(  \bar{\gamma} - \gamma  \right) P_{\vect{\Theta} | \vect{Z} }P_{\vect{Z}} \left( \set{A}_{\gamma} \right)  + \lambda \left( \gamma - \underline{\gamma} \right) P_{\vect{\Theta}  }P_{\vect{Z}} \left( \set{A}_{\gamma}^{\sfc} \right),   \spnum 
\end{IEEEeqnarray}
where $\gamma$ satisfies
\begin{IEEEeqnarray}{rCl}
\underline{\gamma} & \leqslant \gamma \leqslant & \bar{\gamma},
\end{IEEEeqnarray}
and $\bar{\gamma}$ and $\underline{\gamma}$ are respectively defined in~\eqref{EqNovember30at13h35in2025HomeNiceA} and~\eqref{EqNovember30at13h35in2025HomeNiceB}; the set $\set{A}_{\gamma}$ is defined in~\eqref{EqNovember20at13h44in2025HomeNice}; and the measures $P_{\vect{\Theta} | \vect{Z} }P_{\vect{Z}}$ and $P_{\vect{\Theta}}P_{\vect{Z}}$ are both in $\simplex{\set{M} \times \left( \set{X} \times \set{Y}\right)^n}$. 
\end{theorem}
\begin{IEEEproof}
The proof is presented in Appendix~\ref{AppProofOfTheoNovember29at13h20in2025HomeNice}.
\end{IEEEproof}

The interest of Theorem~\ref{TheoNovember29at13h20in2025HomeNice} is beyond the tightness of the bounds in~\eqref{EqDecember1at06h31in2025BusToSophia}, which can be optimized with respect to $\gamma$, $\lambda$, and $Q$. The key observation is the central role played by the hypothesis test in~\eqref{EqNovember28at14h44in2025Nice} in the analysis of  the generalization error~$\overline{\overline{\mathsf{G}}}(P_{\vect{\Theta} | \vect{Z}}, P_{\vect{Z}} )$ in~\eqref{EqJun4at9h02in2024}. 
In particular, note that the constants $\bar{\gamma}$ in~\eqref{EqNovember30at13h35in2025HomeNiceA}  and $\underline{\gamma}$ in~\eqref{EqNovember30at13h35in2025HomeNiceB} are independent of the algorithm $P_{\vect{\Theta} | \vect{Z}}$ under study. 
The dependence on the algorithm $P_{\vect{\Theta} | \vect{Z}}$ and the probability measure on the datasets $P_{\vect{Z}}$ is via the probabilities $P_{\vect{\Theta} | \vect{Z} } P_{\vect{Z}} \left( \set{A}_\gamma \right)$ and $P_{\vect{\Theta}} P_{\vect{Z}} \left( \set{A}_\gamma \right)$, which depend on $\gamma$, $\lambda$, and $Q$; and inform on the probability of choosing the null hypothesis in~\eqref{EqNovember28at14h44in2025Nice} under two different assumptions. The former is under the assumption that models are drawn from the specific instance of the algorithm $P_{\vect{\Theta} | \vect{Z} = \vect{z}}$ and the training dataset $\vect{z}$ is drawn from $P_{\vect{Z}}$. The latter is under the assumptions that models are drawn from $P_{\vect{\Theta}}$ (barycenter of the algorithm~$P_{\vect{\Theta} | \vect{Z}}$) and the training dataset $\vect{z}$ is drawn from~$P_{\vect{Z}}$.

\subsection{Connections to Information Measures}\label{SecDecember26at10h51in2025HomeNice}
The following theorem shows an expression for the generalization error~$\overline{\overline{\mathsf{G}}} \left(P_{\vect{\Theta} | \vect{Z}}, P_{\vect{Z}} \right)$ in~\eqref{EqJun4at9h02in2024} in terms of the sum of the mutual information~$I\left( P_{\vect{\Theta}| \vect{Z}}; P_{\vect{Z}} \right)$ and the lautum information~$L\left( P_{\vect{\Theta}| \vect{Z}}; P_{\vect{Z}} \right)$. 
\begin{theorem}\label{TheoremAugust11at11h02in2024}%
Consider the generalization error~$\overline{\overline{\mathsf{G}}} \left(P_{\vect{\Theta} | \vect{Z}}, P_{\vect{Z}} \right)$ in~\eqref{EqJun4at9h02in2024} and assume that  for all~$\vect{z} \in \left( \set{X} \times \set{Y} \right)^n$, the probability measures~$P_{\vect{\Theta}|\vect{Z} = \vect{z}}$ and~$P_{\vect{\Theta}}$ in~\eqref{EqJune26at16h54in2024} satisfy
%\begin{IEEEeqnarray}{rcl}
$Q \ll P_{\vect{\Theta}|\vect{Z} = \vect{z}}  \ll P_{\vect{\Theta}} \ll Q$, 
%\end{IEEEeqnarray}
with $Q$ the~$\sigma$-finite measure~\eqref{EqGenpdf}.
Then, 
\begin{IEEEeqnarray}{rcl}
	\nonumber
& &  \overline{\overline{\mathsf{G}}}(P_{\vect{\Theta} | \vect{Z}},P_{\vect{Z}} ) = \lambda \left( I\left( P_{\vect{\Theta}| \vect{Z}}; P_{\vect{Z}} \right) + L\left( P_{\vect{\Theta}| \vect{Z}}; P_{\vect{Z}} \right) \right) \\
	\nonumber
	&& + \lambda \iint \log \frac{\mathrm{d} P^{\left(Q, \lambda\right)}_{\vect{\Theta}| \vect{Z} = \vect{z}}}{\mathrm{d} P_{\vect{\Theta}| \vect{Z} = \vect{z}}}\left( \vect{\theta} \right) \mathrm{d} P_{\vect{\Theta} | \vect{Z} = \vect{z}} \left( \vect{\theta} \right)
	  \mathrm{d} P_{\vect{Z}} \left( \vect{z} \right)\\	\IEEEeqnarraynumspace
\label{EqAugust11at11h42in2024}	  
	  	&& - \lambda \iint \log \frac{\mathrm{d} P^{\left(Q, \lambda\right)}_{\vect{\Theta}| \vect{Z} = \vect{z}}}{\mathrm{d} P_{\vect{\Theta}| \vect{Z} = \vect{z}}}\left( \vect{\theta} \right) \mathrm{d} P_{\vect{\Theta}} \left( \vect{\theta} \right)
	  \mathrm{d} P_{\vect{Z}} \left( \vect{z} \right),
\end{IEEEeqnarray}
where the probability measure~$P^{\left(Q, \lambda\right)}_{\vect{\Theta}| \vect{Z} = \vect{z}}$ is defined in~\eqref{EqGenpdf}; and~$\lambda \in \set{K}_{Q}$, with~$\set{K}_{Q}$ in~\eqref{EqSeptember3at18h10in2024}.
\end{theorem}
\begin{IEEEproof}
The proof follows from Lemma~\ref{LemmaAugust3at11h10in2024}, which holds under the current assumptions.  More specifically, the proof is divided into two parts.
The first part focuses on the difference~$\displaystyle\int \bigg(\KL{P_{\vect{\Theta} | \vect{Z} = \vect{z}} }{Q} - \KL{P_{\vect{\Theta}}}{Q} \bigg) 
\mathrm{d} P_{\vect{Z}} \left( \vect{z} \right)$ in~\eqref{EqJune4at11h20in2024}, which satisfies the following equalities: 
\begin{IEEEeqnarray}{rCl}
\nonumber
& &\int \Big( \KL{P_{\vect{\Theta}| \vect{Z} = \vect{z}}}{Q}  - \KL{P_{\vect{\Theta}}}{Q} \Big) \mathrm{d} P_{\vect{Z}}(\vect{z}) \Dsupersqueezeequ \IEEEeqnarraynumspace\\
& = &\int  \KL{P_{\vect{\Theta}| \vect{Z} = \vect{z}}}{Q} \mathrm{d} P_{\vect{Z}}(\vect{z}) -\KL{P_{\vect{\Theta}}}{Q} 
\Dsupersqueezeequ \IEEEeqnarraynumspace \\
\nonumber
& = &\displaystyle\int \left( \int \log\left(\frac{\mathrm{d} P_{\vect{\Theta}| \vect{Z} = \vect{z}}}{\mathrm{d} Q} (\vect{\theta}) \right) \mathrm{d} P_{\vect{\Theta}| \vect{Z} = \vect{z}}(\vect{\theta}) \right) \mathrm{d} P_{\vect{Z}}(\vect{z})   \\
& & -\KL{P_{\vect{\Theta}}}{Q} 
\Dsupersqueezeequ \IEEEeqnarraynumspace \\
\nonumber
& = &\displaystyle\int \left( \int \log\left(\frac{\mathrm{d} P_{\vect{\Theta}| \vect{Z} = \vect{z}}}{\mathrm{d} Q} (\vect{\theta}) \right) \mathrm{d} P_{\vect{\Theta}| \vect{Z} = \vect{z}}(\vect{\theta}) \right) \mathrm{d} P_{\vect{Z}}(\vect{z}) \\
& &-\int\log\left( \frac{\mathrm{d}P_{\vect{\Theta}}}{\mathrm{d}Q} (\vect{\theta}) \right)\mathrm{d}P_{\vect{\Theta}}(\vect{\theta}) 
\Dsupersqueezeequ \IEEEeqnarraynumspace \\
\nonumber
& = &\displaystyle\int \left( \int \log\left(\frac{\mathrm{d} P_{\vect{\Theta}| \vect{Z} = \vect{z}}}{\mathrm{d} Q} (\vect{\theta}) \right) \mathrm{d} P_{\vect{\Theta}| \vect{Z} = \vect{z}}(\vect{\theta}) \right) \mathrm{d}P_{\vect{Z}}(\vect{z}) 
\Dsupersqueezeequ \\
\label{EqMay11at19h28in2024DaliIsInBusan}
&  & - \int \left( \int\log\left( \frac{\mathrm{d}P_{\vect{\Theta}}}{\mathrm{d}Q} (\vect{\theta}) \right)\mathrm{d} P_{\vect{\Theta}| \vect{Z} = \vect{z}}(\vect{\theta})  \right) \mathrm{d}P_{\vect{Z}}(\vect{z}) 
\Dsupersqueezeequ \IEEEeqnarraynumspace \\
\label{EqMay11at19h28in2024DaliIsInBusanAA}
& = &\displaystyle\int \Bigg( \int  \log\left(\frac{\mathrm{d} P_{\vect{\Theta}| \vect{Z} = \vect{z}}}{\mathrm{d} P_{\vect{\Theta}}} (\vect{\theta}) \right) 
\mathrm{d} P_{\vect{\Theta}| \vect{Z} = \vect{z}}(\vect{\theta}) \Bigg) \mathrm{d}P_{\vect{Z}}(\vect{z}) \Dsupersqueezeequ \IEEEeqnarraynumspace\\
\label{EqDecember19in16h53in2025}
& = & I\left( P_{\vect{\Theta}| \vect{Z}}; P_{\vect{Z}} \right), 
\end{IEEEeqnarray} 
where the equality in~\eqref{EqMay11at19h28in2024DaliIsInBusanAA} follows from \cite[Theorem 4]{InriaRR9591}.
Plugging~\eqref{EqDecember19in16h53in2025} into~\eqref{EqJune4at11h20in2024} leads to
\begin{IEEEeqnarray}{rcl}
	\nonumber
& &  \overline{\overline{\mathsf{G}}}(P_{\vect{\Theta} | \vect{Z}},P_{\vect{Z}} ) = \lambda I\left( P_{\vect{\Theta}| \vect{Z}}; P_{\vect{Z}} \right) \\
\label{EqDecember31at23h50in2023}
	&& + \lambda \int \left( \KL{P_{\vect{\Theta}}}{P^{\left(Q, \lambda\right)}_{\vect{\Theta}| \vect{Z} = \vect{z}}} - \KL{P_{\vect{\Theta}|\vect{Z} = \vect{z}}}{P^{\left(Q, \lambda\right)}_{\vect{\Theta}| \vect{Z} = \vect{z}}}\right) \mathrm{d} P_{\vect{Z}} \left( \vect{z} \right). 
	\Dsupersqueezeequ\IEEEeqnarraynumspace
\end{IEEEeqnarray}
The second part of the proof focuses on the term~$\int \KL{P_{\vect{\Theta}}}{P^{\left(Q, \lambda\right)}_{\vect{\Theta}| \vect{Z} = \vect{z}}} \mathrm{d} P_{\vect{Z}} \left( \vect{z} \right)$ in~\eqref{EqDecember31at23h50in2023}. Hence, the following holds
\begin{IEEEeqnarray}{rcl}
\nonumber
& &\int \KL{P_{\vect{\Theta}}}{P^{\left(Q, \lambda\right)}_{\vect{\Theta}| \vect{Z} = \vect{z}}} \mathrm{d} P_{\vect{Z}} \left( \vect{z} \right) \\
& = & \iint \log \frac{\mathrm{d}P_{\vect{\Theta}}}{\mathrm{d}P^{\left(Q, \lambda\right)}_{\vect{\Theta}| \vect{Z} = \vect{z}}}\left( \vect{\theta} \right) \mathrm{d}P_{\vect{\Theta}}\left( \vect{\theta} \right)\mathrm{d} P_{\vect{Z}} \left( \vect{z} \right) \\
\label{EqDecember19at16h01in2025}
& = &\iint \log \left( \frac{\mathrm{d}P_{\vect{\Theta}}}{\mathrm{d}P^{\left(Q, \lambda\right)}_{\vect{\Theta}| \vect{Z} = \vect{z}}}\left( \vect{\theta} \right)  \frac{\mathrm{d}P_{\vect{\Theta}|\vect{Z} = \vect{z}}}{\mathrm{d}P_{\vect{\Theta}|\vect{Z} = \vect{z}}}\left( \vect{\theta} \right) \right)\mathrm{d}P_{\vect{\Theta}}\left( \vect{\theta} \right) \mathrm{d} P_{\vect{Z}} \left( \vect{z} \right) \middlesqueezeequ \spnum\\
%& = & \int \log \frac{\mathrm{d}P_{\vect{\Theta}|\vect{Z} = \vect{z}}}{\mathrm{d}P^{\left(Q, \lambda\right)}_{\vect{\Theta}| \vect{Z} = \vect{z}}}\left( \vect{\theta} \right)  \frac{\mathrm{d}P_{\vect{\Theta}}}{\mathrm{d}P_{\vect{\Theta}|\vect{Z} = \vect{z}}}\left( \vect{\theta} \right) \mathrm{d}P_{\vect{\Theta}}\left( \vect{\theta} \right)\\
\nonumber 
& = & \iint \log \frac{\mathrm{d}P_{\vect{\Theta}|\vect{Z} = \vect{z}}}{\mathrm{d}P^{\left(Q, \lambda\right)}_{\vect{\Theta}| \vect{Z} = \vect{z}}}\left( \vect{\theta} \right)   \mathrm{d}P_{\vect{\Theta}}\left( \vect{\theta} \right)\mathrm{d} P_{\vect{Z}} \left( \vect{z} \right) \\ 
\label{EqAugust11at13h55in2024}
&& + \iint \log \frac{\mathrm{d}P_{\vect{\Theta}}}{\mathrm{d}P_{\vect{\Theta}|\vect{Z} = \vect{z}}}\left( \vect{\theta} \right) \mathrm{d}P_{\vect{\Theta}}\left( \vect{\theta} \right)\mathrm{d} P_{\vect{Z}} \left( \vect{z} \right) \\
\label{EqNovember7at11h37in2024Nice}
& = &\iint \log \frac{\mathrm{d}P_{\vect{\Theta}|\vect{Z} = \vect{z}}}{\mathrm{d}P^{\left(Q, \lambda\right)}_{\vect{\Theta}| \vect{Z} = \vect{z}}}\left( \vect{\theta} \right)   \mathrm{d}P_{\vect{\Theta}}\left( \vect{\theta} \right)\mathrm{d} P_{\vect{Z}} \left( \vect{z} \right)  + L\left( P_{\vect{\Theta}| \vect{Z}}; P_{\vect{Z}} \right), \squeezeequ \spnum
\end{IEEEeqnarray}
where  the equality in~\eqref{EqDecember19at16h01in2025} follows from~\cite[Theorem 3]{InriaRR9591}; 
and the equality in~\eqref{EqAugust11at13h55in2024} follows from~\cite[Theorem 4]{InriaRR9591} and \cite[Lemma~$3$]{PerlazaTIT2024}.
Finally, plugging~\eqref{EqNovember7at11h37in2024Nice} into~\eqref{EqDecember31at23h50in2023} completes the proof.
\end{IEEEproof}
In Theorem~\ref{TheoremAugust11at11h02in2024}, the term
\begin{IEEEeqnarray}{rCl}
\nonumber
	&&\lambda \iint \log \frac{\mathrm{d} P^{\left(Q, \lambda\right)}_{\vect{\Theta}| \vect{Z} = \vect{z}}}{\mathrm{d} P_{\vect{\Theta}| \vect{Z} = \vect{z}}}\left( \vect{\theta} \right) \mathrm{d} P_{\vect{\Theta} | \vect{Z} = \vect{z}} \left( \vect{\theta} \right)
	  \mathrm{d} P_{\vect{Z}} \left( \vect{z} \right)\\	\IEEEeqnarraynumspace
\label{EqAugust25at15h31in2024}
	  	&& - \lambda \iint \log \frac{\mathrm{d} P^{\left(Q, \lambda\right)}_{\vect{\Theta}| \vect{Z} = \vect{z}}}{\mathrm{d} P_{\vect{\Theta}| \vect{Z} = \vect{z}}}\left( \vect{\theta} \right) \mathrm{d} P_{\vect{\Theta}} \left( \vect{\theta} \right)
	  \mathrm{d} P_{\vect{Z}} \left( \vect{z} \right),
\end{IEEEeqnarray}
is of the same form as the one in the right-hand side of~\eqref{EqAugust9at11h44in2024}.
Thus, it can also be interpreted in the context of a hypothesis test. 
Consider the training dataset~$\vect{z}$ in~\eqref{EqTheDataSet}, which is obtained by sampling from the measure~$P_{\vect{Z}}$. Assume that both the algorithm~$P_{\vect{\Theta} | \vect{Z}}$ and the Gibbs algorithm~$P^{\left(Q, \lambda\right)}_{\vect{\Theta}| \vect{Z}}$ are trained upon the same dataset~$\vect{z}$.
Hence, the hypothesis test consists in determining whether a model~$\vect{\theta}$ has been obtained from the probability measure~$P^{\left(Q, \lambda\right)}_{\vect{\Theta}| \vect{Z} = \vect{z}}$, which represents a Gibbs algorithm trained upon $\vect{z}$ ($H_0$ null hypothesis); or  from the probability measure~$P_{\vect{\Theta} | \vect{Z} = \vect{z}}$, which represents the algorithm under study trained upon $\vect{z}$ ($H_1$ alternative hypothesis). That is, 
\begin{IEEEeqnarray}{c}
\label{EqDecember20at21h11in2025Nice}
\left\lbrace
\begin{array}{rcl}
H_0: & &  \vect{\theta}  \sim P^{\left(Q, \lambda\right)}_{\vect{\Theta}| \vect{Z} = \vect{z}}\\
H_1: & &  \vect{\theta}  \sim  P_{\vect{\Theta}| \vect{Z} = \vect{z}}.
\end{array}
\right.
\end{IEEEeqnarray}
The main objective of such a hypothesis test is to determine whether models are being sampled from a Gibbs algorithm or the algorithm under study, assuming that both algorithms have been trained upon the same dataset.
The resulting log-likelihood ratio to decide on either hypothesis in \eqref{EqDecember20at21h11in2025Nice}, based on the observation of the pair~$\left(\vect{\theta},
\vect{z} \right)$, is~$\log \frac{\mathrm{d} P^{\left(Q, \lambda\right)}_{\vect{\Theta}| \vect{Z} = \vect{z}}}{\mathrm{d} P_{\vect{\Theta}| \vect{Z} = \vect{z}}}\left( \vect{\theta} \right)$.
Hence, given a nonnegative real $\gamma$, the null hypothesis is accepted if $\log \frac{\mathrm{d} P^{\left(Q, \lambda\right)}_{\vect{\Theta}| \vect{Z} = \vect{z}}}{\mathrm{d} P_{\vect{\Theta}| \vect{Z} = \vect{z}}}\left( \vect{\theta} \right) \geqslant \gamma$. Conversely, the alternative hypothesis is accepted if $\log \frac{\mathrm{d} P^{\left(Q, \lambda\right)}_{\vect{\Theta}| \vect{Z} = \vect{z}}}{\mathrm{d} P_{\vect{\Theta}| \vect{Z} = \vect{z}}}\left( \vect{\theta} \right) < \gamma$.
The acceptance region of the null hypothesis is the set
\begin{IEEEeqnarray}{rcl}
\label{EqDecember20at21h49in2025HomeNice}
\set{B}_{\gamma} & \triangleq &\left\lbrace  \left( \vect{\theta}, \vect{z} \right) \in \set{M} \times \left( \set{X} \times \set{Y} \right)^n : \log \frac{\mathrm{d} P^{\left(Q, \lambda\right)}_{\vect{\Theta}| \vect{Z} = \vect{z}}}{\mathrm{d} P_{\vect{\Theta}| \vect{Z} = \vect{z}}}\left( \vect{\theta} \right) \geqslant \gamma \right\rbrace. \squeezeequ \IEEEeqnarraynumspace
\end{IEEEeqnarray}
Consider the following probabilities
\begin{IEEEeqnarray}{rcl}
\label{EqDecember20at21h55in2025HomeNiceA}
P_{\vect{\Theta} | \vect{Z} } P_{\vect{Z}} \left( \set{B}_\gamma \right) & = & 1 - P_{\vect{\Theta} | \vect{Z} } P_{\vect{Z}} \left( \set{B}_\gamma^{\sfc} \right), \mbox{ and }\\
\label{EqDecember20at21h55in2025HomeNiceB}
P_{\vect{\Theta}} P_{\vect{Z}} \left( \set{B}_\gamma \right) & = & 1 - P_{\vect{\Theta} } P_{\vect{Z}} \left( \set{B}_\gamma^{\sfc} \right).
\end{IEEEeqnarray}
The former represents the probability of choosing the null hypothesis under the assumption that the observation $\left(\vect{\theta}, \vect{z} \right)$ is sampled from the joint probability measure $P_{\vect{\Theta} | \vect{Z} } P_{\vect{Z}}$.
The latter represents the probability of choosing the null hypothesis under the assumption that the observation $\left(\vect{\theta}, \vect{z} \right)$ is sampled from the product probability measure $P_{\vect{\Theta}} P_{\vect{Z}}$.
Consider also the following values
\begin{IEEEeqnarray}{rcl}
\label{EqDecember20at22h15in2025HomeNiceA}
\bar{\gamma} & \triangleq & \sup \left\lbrace \log \frac{\mathrm{d} P^{\left(Q, \lambda\right)}_{\vect{\Theta}| \vect{Z} = \vect{z}}}{\mathrm{d} P_{\vect{\Theta}| \vect{Z} = \vect{z}}}\left( \vect{\theta} \right) \in \reals  : \left( \vect{\theta}, \vect{z} \right)  \in \set{M} \times \left( \set{X} \times \set{Y} \right)^n \right\rbrace \squeezeequ\spnum
\end{IEEEeqnarray}
and 
\begin{IEEEeqnarray}{rcl}
\label{EqDecember20at22h15in2025HomeNiceB}
\underline{\gamma} & \triangleq & \inf  \left\lbrace \log \frac{\mathrm{d} P^{\left(Q, \lambda\right)}_{\vect{\Theta}| \vect{Z} = \vect{z}}}{\mathrm{d} P_{\vect{\Theta}| \vect{Z} = \vect{z}}}\left( \vect{\theta} \right) \in \reals  : \left( \vect{\theta}, \vect{z} \right)  \in \set{M} \times \left( \set{X} \times \set{Y} \right)^n \right\rbrace. \squeezeequ\spnum
\end{IEEEeqnarray}
Using this notation, the following theorem strengthens the connexion between the generalization error~$\overline{\overline{\mathsf{G}}} \left(P_{\vect{\Theta} | \vect{Z}}, P_{\vect{Z}} \right)$ in~\eqref{EqJun4at9h02in2024} and the hypothesis test in~\eqref{EqDecember20at21h11in2025Nice}.
\begin{theorem}\label{TheoDecember21at7h07in2025HomeNice}
Consider the generalization error~$\overline{\overline{\mathsf{G}}} \left(P_{\vect{\Theta} | \vect{Z}}, P_{\vect{Z}} \right)$ in~\eqref{EqJun4at9h02in2024} and assume that for all~$\vect{z} \in \left( \set{X} \times \set{Y} \right)^n$, the  measures~$P_{\vect{\Theta} | \vect{Z} = \vect{z}}$, $P_{\vect{\Theta}}$ in~\eqref{EqJune26at16h54in2024}, and the~$\sigma$-finite measure~$Q$ in~\eqref{EqGenpdf}, satisfy $Q \ll P_{\vect{\Theta} | \vect{Z} = \vect{z}} \ll P_{\vect{\Theta}} \ll Q$. 
 Then,
\begin{IEEEeqnarray}{C}
\nonumber
\lambda \left(  \underline{\gamma}  - \gamma \right) P_{\vect{\Theta} | \vect{Z} }P_{\vect{Z}} \left( \set{B}_{\gamma}^{\sfc} \right)  + \lambda \left( \gamma - \bar{\gamma} \right) P_{\vect{\Theta}  }P_{\vect{Z}} \left( \set{B}_{\gamma} \right)\\ 
\nonumber
 \leqslant \overline{\overline{\mathsf{G}}} \left(P_{\vect{\Theta} | \vect{Z}}, P_{\vect{Z}} \right) -  \lambda \left( I\left( P_{\vect{\Theta}| \vect{Z}}; P_{\vect{Z}} \right) + L\left( P_{\vect{\Theta}| \vect{Z}}; P_{\vect{Z}} \right) \right) \leqslant\\
\label{EqDecember21at06h31in2025BusToSophia}
  \lambda \left(  \bar{\gamma} - \gamma  \right) P_{\vect{\Theta} | \vect{Z} }P_{\vect{Z}} \left( \set{B}_{\gamma} \right)  + \lambda \left( \gamma - \underline{\gamma} \right) P_{\vect{\Theta}  }P_{\vect{Z}} \left( \set{B}_{\gamma}^{\sfc} \right),   \spnum 
\end{IEEEeqnarray}
where $\gamma$ satisfies
\begin{IEEEeqnarray}{rCl}
\underline{\gamma} & \leqslant \gamma \leqslant & \bar{\gamma},
\end{IEEEeqnarray}
and $\bar{\gamma}$ and $\underline{\gamma}$ are respectively defined in~\eqref{EqDecember20at22h15in2025HomeNiceA} and~\eqref{EqDecember20at22h15in2025HomeNiceB}; the set $\set{B}_{\gamma}$ is defined in~\eqref{EqDecember20at21h49in2025HomeNice}; and the measures $P_{\vect{\Theta} | \vect{Z} }P_{\vect{Z}}$ and $P_{\vect{\Theta}}P_{\vect{Z}}$ are both in $\simplex{\set{M} \times \left( \set{X} \times \set{Y}\right)^n}$. 
\end{theorem}
\begin{IEEEproof}
The proof is presented in Appendix~\ref{AppProofOfTheoDecember21at7h07in2025HomeNice}.
\end{IEEEproof}
The lower and upper bounds in Theorem~\ref{TheoDecember21at7h07in2025HomeNice} are tight in at least two cases. First, consider that the algorithm under study is the Gibbs algorithm, i.e.,  $P^{\left(Q, \lambda\right)}_{\vect{\Theta}| \vect{Z}}$ in~\eqref{EqGenpdf}. In such a case, $\bar{\gamma} = \underline{\gamma} = 0$ and thus, $\overline{\overline{\mathsf{G}}} \left(P^{\left(Q, \lambda\right)}_{\vect{\Theta}| \vect{Z}}, P_{\vect{Z}} \right) =  \lambda \left( I\left( P^{\left(Q, \lambda\right)}_{\vect{\Theta}| \vect{Z}}; P_{\vect{Z}} \right) +  L\left( P^{\left(Q, \lambda\right)}_{\vect{\Theta}| \vect{Z}}; P_{\vect{Z}} \right) \right)$. This case is studied in more detail in Section~\ref{SecDecember21at9h33in2025HomeNice}.  
The second case is for an algorithm whose output is independent of the training dataset, i.e., an algorithm such that for all $\vect{z} \in \left( \set{X} \times \set{Y} \right)^n$, $\KL{P_{\vect{\Theta}| \vect{Z} = \vect{z}}}{P_{\vect{\Theta}}} = 0$.  In this case, the generalization error~$\overline{\overline{\mathsf{G}}} \left(P_{\vect{\Theta} | \vect{Z}}, P_{\vect{Z}} \right)$ in~\eqref{EqJun4at9h02in2024} is zero. Moreover,
\begin{IEEEeqnarray}{rCl}
0 & =& I\left( P_{\vect{\Theta}| \vect{Z}}; P_{\vect{Z}} \right) \\
& = & L\left( P_{\vect{\Theta}| \vect{Z}}; P_{\vect{Z}} \right)  \\
\nonumber
& = & \iint \log \frac{\mathrm{d} P^{\left(Q, \lambda\right)}_{\vect{\Theta}| \vect{Z} = \vect{z}}}{\mathrm{d} P_{\vect{\Theta}| \vect{Z} = \vect{z}}}\left( \vect{\theta} \right) \mathrm{d} P_{\vect{\Theta} | \vect{Z} = \vect{z}} \left( \vect{\theta} \right)
	  \mathrm{d} P_{\vect{Z}} \left( \vect{z} \right)\\	\IEEEeqnarraynumspace
\label{EqDecember21at15h33in2025HomeNice}
	  	&& - \iint \log \frac{\mathrm{d} P^{\left(Q, \lambda\right)}_{\vect{\Theta}| \vect{Z} = \vect{z}}}{\mathrm{d} P_{\vect{\Theta}| \vect{Z} = \vect{z}}}\left( \vect{\theta} \right) \mathrm{d} P_{\vect{\Theta}} \left( \vect{\theta} \right)
	  \mathrm{d} P_{\vect{Z}} \left( \vect{z} \right).
\end{IEEEeqnarray}
Beyond the tightness of these bounds, which might be optimized by strategically choosing the value of $\gamma$ in the case of any arbitrary algorithm $P_{\vect{\Theta} | \vect{Z}}$, the interest of Theorem~\ref{TheoDecember21at7h07in2025HomeNice} lies on the following observation. The generalization error~$\overline{\overline{\mathsf{G}}} \left(P_{\vect{\Theta} | \vect{Z}}, P_{\vect{Z}} \right)$ in~\eqref{EqAugust11at11h42in2024} is the sum of two quantities: $(a)$ A first quantity that characterizes the dependence of the models on the training data via the mutual and lautum information, i.e., $\lambda\left( I\left( P_{\vect{\Theta}| \vect{Z}}; P_{\vect{Z}} \right) + L\left( P_{\vect{\Theta}| \vect{Z}}; P_{\vect{Z}} \right)\right)$; and $(b)$ a second quantity that characterizes a statistical distance between the algorithm under study and a Gibbs algorithm via the expression in~\eqref{EqAugust25at15h31in2024}. 
While the first quantity $(a)$ is always nonnegative, the second quantity $(b)$ might be negative, zero, or positive. As shown in~\eqref{EqDecember21at15h33in2025HomeNice}, it is zero when the algorithm under study is the  Gibbs algorithm $P^{\left(Q, \lambda\right)}_{\vect{\Theta}| \vect{Z}}$.

Clearly, a variation in any of these quantities, $(a)$ or $(b)$, implies a variation in the other, and thus, cannot be separately optimized.  
Nonetheless, these quantities place Gibbs algorithms as natural references to evaluate the generalization capabilities of supervised machine learning algorithms. 
For instance, the deviation of the generalization error~$\overline{\overline{\mathsf{G}}} \left(P_{\vect{\Theta} | \vect{Z}}, P_{\vect{Z}} \right)$ in~\eqref{EqJun4at9h02in2024}  from the first quantity $ \lambda \left( I\left( P_{\vect{\Theta}| \vect{Z}}; P_{\vect{Z}} \right) +  L\left( P_{\vect{\Theta}| \vect{Z}}; P_{\vect{Z}} \right) \right)$ is upper and lower bounded by a function of the smallest and the largest log-likelihoods in the hypothesis test problem in \eqref{EqDecember20at21h11in2025Nice}. 
More specifically, Theorem~\ref{TheoDecember21at7h07in2025HomeNice}, using $\gamma^{\star} \triangleq \frac{\bar{\gamma} + \underline{\gamma}}{2}$, with $\bar{\gamma}$ in~\eqref{EqDecember20at22h15in2025HomeNiceA} and $\underline{\gamma}$ in~\eqref{EqDecember20at22h15in2025HomeNiceB}, leads to 
\begin{IEEEeqnarray}{C}
\nonumber
- \lambda \left(  \frac{\bar{\gamma} - \underline{\gamma}}{2} \right) P_{\vect{\Theta} | \vect{Z} }P_{\vect{Z}} \left( \set{B}_{\gamma^{\star}}^{\sfc} \right)  - \lambda \left( \frac{\bar{\gamma} - \underline{\gamma}}{2}  \right) P_{\vect{\Theta}  }P_{\vect{Z}} \left( \set{B}_{\gamma^{\star}} \right)\\ 
\nonumber
 \leqslant \overline{\overline{\mathsf{G}}} \left(P_{\vect{\Theta} | \vect{Z}}, P_{\vect{Z}} \right) -  \lambda \left( I\left( P_{\vect{\Theta}| \vect{Z}}; P_{\vect{Z}} \right) + L\left( P_{\vect{\Theta}| \vect{Z}}; P_{\vect{Z}} \right) \right) \leqslant\\
\label{EqDecember22at08h24in2025Sophia}
  \lambda \left( \frac{\bar{\gamma} - \underline{\gamma}}{2}   \right) P_{\vect{\Theta} | \vect{Z} }P_{\vect{Z}} \left( \set{B}_{\gamma^{\star}} \right)  + \lambda \left( \frac{\bar{\gamma} - \underline{\gamma}}{2}  \right) P_{\vect{\Theta}  }P_{\vect{Z}} \left( \set{B}_{\gamma^{\star}}^{\sfc} \right).   \spnum 
\end{IEEEeqnarray}
This implies that the largest deviation of the generalization error~$\overline{\overline{\mathsf{G}}} \left(P_{\vect{\Theta} | \vect{Z}}, P_{\vect{Z}} \right)$ in~\eqref{EqJun4at9h02in2024}  from the quantity $ \lambda \left( I\left( P_{\vect{\Theta}| \vect{Z}}; P_{\vect{Z}} \right) +  L\left( P_{\vect{\Theta}| \vect{Z}}; P_{\vect{Z}} \right) \right)$ is at most $ \lambda \left( \frac{\bar{\gamma} - \underline{\gamma}}{2}   \right)$, which is the difference between the lower and upper bounds in \eqref{EqDecember22at08h24in2025Sophia}. 
This leads to the conclusion that the generalization error~$\overline{\overline{\mathsf{G}}} \left(P_{\vect{\Theta} | \vect{Z}}, P_{\vect{Z}} \right)$ in~\eqref{EqJun4at9h02in2024} can be \emph{well} approximated by $ \lambda \left( I\left( P_{\vect{\Theta}| \vect{Z}}; P_{\vect{Z}} \right) +  L\left( P_{\vect{\Theta}| \vect{Z}}; P_{\vect{Z}} \right) \right)$, when the  algorithm $P_{\vect{\Theta} | \vect{Z}}$ is \emph{close enough}  to the Gibbs algorithm $P^{\left(Q, \lambda\right)}_{\vect{\Theta}| \vect{Z}}$.
More specifically, if the absolute value of the statistical distance in~\eqref{EqAugust25at15h31in2024} is smaller than or equal to some $\delta \geqslant 0$, that is,
\begin{IEEEeqnarray}{rCl}
\nonumber
& &\abs{\int \left( 1 - \frac{\mathrm{d} P_{\vect{\Theta}}}{\mathrm{d} P_{\vect{\Theta}| \vect{Z} = \vect{z}}}\left( \vect{\theta} \right)  \right) \log \frac{\mathrm{d} P^{\left(Q, \lambda\right)}_{\vect{\Theta}| \vect{Z} = \vect{z}}}{\mathrm{d} P_{\vect{\Theta}| \vect{Z} = \vect{z}}}\left( \vect{\theta} \right) \mathrm{d} P_{\vect{\Theta} | \vect{Z}}P_{\vect{Z}} \left(\vect{\theta}, \vect{z} \right)} \middlesqueezeequ\\
\label{EqDecember22at11h56in2025Sophia}
& \leqslant & \frac{\delta}{\lambda}, 
\end{IEEEeqnarray}
 then, 
 \begin{IEEEeqnarray}{rCl}
\abs{\overline{\overline{\mathsf{G}}} \left(P_{\vect{\Theta} | \vect{Z}}, P_{\vect{Z}} \right) -  \lambda \left( I\left( P_{\vect{\Theta}| \vect{Z}}; P_{\vect{Z}} \right) + L\left( P_{\vect{\Theta}| \vect{Z}}; P_{\vect{Z}} \right) \right)} \leqslant \delta,  \squeezeequ \spnum
 \end{IEEEeqnarray}
which justifies such an approximation.

\subsection{Connections to Euclidean Geometry}

Another expression for the generalization error~$\overline{\overline{\mathsf{G}}} \left(P_{\vect{\Theta} | \vect{Z}}, P_{\vect{Z}} \right)$ in~\eqref{EqJun4at9h02in2024}, which leads to interesting connections to Euclidean geometry is presented hereunder. 

\begin{theorem}\label{TheoremJuin26at11h03in2024}
Consider the generalization error~$\overline{\overline{\mathsf{G}}} \left(P_{\vect{\Theta} | \vect{Z}}, P_{\vect{Z}} \right)$ in~\eqref{EqJun4at9h02in2024} and assume that  for all~$\vect{z} \in \left( \set{X} \times \set{Y} \right)^n$,~$P_{\vect{\Theta}|\vect{Z} = \vect{z}} \ll Q$, with $Q$ the~$\sigma$-finite measure~\eqref{EqGenpdf}. Then,
\begin{IEEEeqnarray}{rcl}
	\nonumber
 \overline{\overline{\mathsf{G}}}(P_{\vect{\Theta} | \vect{Z}},P_{\vect{Z}} )  
& = & \lambda \iint\bigg( \KL{P_{\vect{\Theta} | \vect{Z} = \vect{z}}}{P^{\left(Q, \lambda\right)}_{\vect{\Theta}| \vect{Z} = \vect{u}}} \Dsupersqueezeequ\IEEEeqnarraynumspace\\
 & & - \KL{P_{\vect{\Theta} | \vect{Z} = \vect{z}}}{P^{\left(Q, \lambda\right)}_{\vect{\Theta}| \vect{Z} = \vect{z}}} \bigg) \mathrm{d}P_{\vect{Z}} \left( \vect{u} \right) \mathrm{d}P_{\vect{Z}} \left( \vect{z} \right),
 \Dsupersqueezeequ\IEEEeqnarraynumspace
\end{IEEEeqnarray}
where the probability measure~$P^{\left(Q, \lambda\right)}_{\vect{\Theta}| \vect{Z} = \vect{z}}$ satisfies~\eqref{EqGenpdf}; and~$\lambda \in \set{K}_{Q}$, with~$\set{K}_{Q}$ in~\eqref{EqSeptember3at18h10in2024}.
\end{theorem}
\begin{IEEEproof}
The proof consists in re-writing the difference~$\mathsf{R}_{\vect{u}} \left( P_{\vect{\Theta} | \vect{Z} = \vect{z}} \right) - 
\mathsf{R}_{\vect{z}}\left( P_{\vect{\Theta} | \vect{Z} = \vect{z}} \right)$ in~\eqref{EqJun4At8h36in2024} in terms of an algorithm-driven gap, i.e., using the functional~$\mathsf{G}$ in~\eqref{EqGZeta}. More specifically, consider the equalities
\begin{IEEEeqnarray}{Lcl}
\mathsf{R}_{\vect{u}} \left( P_{\vect{\Theta} | \vect{Z} = \vect{z}} \right)   & = & \mathsf{G}\left( \vect{u}, P_{\vect{\Theta} | \vect{Z} = \vect{z}} , P^{\left(Q, \lambda\right)}_{\vect{\Theta}| \vect{Z} = \vect{u}}\right)    + \mathsf{R}_{\vect{u}}\left( P^{\left(Q, \lambda\right)}_{\vect{\Theta}| \vect{Z} = \vect{u}} \right)\Dsupersqueezeequ \IEEEeqnarraynumspace
\end{IEEEeqnarray}
and 
\begin{IEEEeqnarray}{Lcl}
\mathsf{R}_{\vect{z}} \left( P_{\vect{\Theta} | \vect{Z} = \vect{z}} \right)   & = & \mathsf{G}\left( \vect{z}, P_{\vect{\Theta} | \vect{Z} = \vect{z}} , P^{\left(Q, \lambda\right)}_{\vect{\Theta}| \vect{Z} = \vect{z}}\right)    + \mathsf{R}_{\vect{z}}\left( P^{\left(Q, \lambda\right)}_{\vect{\Theta}| \vect{Z} = \vect{z}} \right),\Dsupersqueezeequ \IEEEeqnarraynumspace
\end{IEEEeqnarray}
that implies, 
\begin{IEEEeqnarray}{rCl}
\nonumber
& & \mathsf{R}_{\vect{u}} \left( P_{\vect{\Theta} | \vect{Z} = \vect{z}} \right) - 
\mathsf{R}_{\vect{z}}\left( P_{\vect{\Theta} | \vect{Z} = \vect{z}} \right) \\
\nonumber
& = &  \mathsf{G}\left( \vect{u}, P_{\vect{\Theta} | \vect{Z} = \vect{z}} , P^{\left(Q, \lambda\right)}_{\vect{\Theta}| \vect{Z} = \vect{u}}\right)    + \mathsf{R}_{\vect{u}}\left( P^{\left(Q, \lambda\right)}_{\vect{\Theta}| \vect{Z} = \vect{u}} \right)\\
\label{EqJune26at14h16in2024a}
& & -  \mathsf{G}\left( \vect{z}, P_{\vect{\Theta} | \vect{Z} = \vect{z}} , P^{\left(Q, \lambda\right)}_{\vect{\Theta}| \vect{Z} = \vect{z}}\right)    -  \mathsf{R}_{\vect{z}}\left( P^{\left(Q, \lambda\right)}_{\vect{\Theta}| \vect{Z} = \vect{z}} \right) \\
%\nonumber
%& = &  \lambda\bigg( \KL{P^{\left(Q, \lambda\right)}_{\vect{\Theta}| \vect{Z} = \vect{u}}}{Q} + \KL{P_{\vect{\Theta} | \vect{Z} = \vect{z}}}{P^{\left(Q, \lambda\right)}_{\vect{\Theta}| \vect{Z} = \vect{u}}}\Dsupersqueezeequ\\
%\nonumber
%& & - \KL{P^{\left(Q, \lambda\right)}_{\vect{\Theta}| \vect{Z} = \vect{z}}}{Q} - \KL{P_{\vect{\Theta} | \vect{Z} = \vect{z}}}{P^{\left(Q, \lambda\right)}_{\vect{\Theta}| \vect{Z} = \vect{z}}} \bigg) \\
%& &   + \mathsf{R}_{\vect{u}}\left( P^{\left(Q, \lambda\right)}_{\vect{\Theta}| \vect{Z} = \vect{u}} \right) -  \mathsf{R}_{\vect{z}}\left( P^{\left(Q, \lambda\right)}_{\vect{\Theta}| \vect{Z} = \vect{z}} \right) \\
\nonumber
& = &   \mathsf{R}_{\vect{u}}\left( P^{\left(Q, \lambda\right)}_{\vect{\Theta}| \vect{Z} = \vect{u}} \right) +\lambda \KL{P^{\left(Q, \lambda\right)}_{\vect{\Theta}| \vect{Z} = \vect{u}}}{Q} \\
\nonumber
& & -  \mathsf{R}_{\vect{z}}\left( P^{\left(Q, \lambda\right)}_{\vect{\Theta}| \vect{Z} = \vect{z}} \right)  - \lambda\KL{P^{\left(Q, \lambda\right)}_{\vect{\Theta}| \vect{Z} = \vect{z}}}{Q}\\
\label{EqJune26at14h16in2024b}
& & + \lambda\bigg( \KL{P_{\vect{\Theta} | \vect{Z} = \vect{z}}}{P^{\left(Q, \lambda\right)}_{\vect{\Theta}| \vect{Z} = \vect{u}}} - \KL{P_{\vect{\Theta} | \vect{Z} = \vect{z}}}{P^{\left(Q, \lambda\right)}_{\vect{\Theta}| \vect{Z} = \vect{z}}} \bigg)\Dsupersqueezeequ\IEEEeqnarraynumspace\\
\nonumber
& = &   \mathsf{R}_{\vect{u}}\left( Q \right) -\lambda \KL{Q}{P^{\left(Q, \lambda\right)}_{\vect{\Theta}| \vect{Z} = \vect{u}}} \\
\nonumber
& & -  \mathsf{R}_{\vect{z}}\left( Q \right)  + \lambda\KL{Q}{P^{\left(Q, \lambda\right)}_{\vect{\Theta}| \vect{Z} = \vect{z}}}\\
\label{EqJune26at14h16in2024c}
& & + \lambda\bigg( \KL{P_{\vect{\Theta} | \vect{Z} = \vect{z}}}{P^{\left(Q, \lambda\right)}_{\vect{\Theta}| \vect{Z} = \vect{u}}} - \KL{P_{\vect{\Theta} | \vect{Z} = \vect{z}}}{P^{\left(Q, \lambda\right)}_{\vect{\Theta}| \vect{Z} = \vect{z}}} \bigg),\Dsupersqueezeequ\IEEEeqnarraynumspace
\end{IEEEeqnarray}
where the equality in~\eqref{EqJune26at14h16in2024b} follows from \cite[Theorem~$1$]{Perlaza-ISIT2023b}, which holds under the assumption of the theorem and \cite[Lemma~$3$]{PerlazaTIT2024}; 
and the equality in~\eqref{EqJune26at14h16in2024c} follows from \cite[Lemma~$20$]{PerlazaTIT2024}.
Finally, from~\eqref{EqJune26at14h16in2024c} and Definition \ref{DefDEGG}, it follows that
\begin{IEEEeqnarray}{rcl}
\nonumber
& & \overline{\overline{\mathsf{G}}} \left(P_{\vect{\Theta} | \vect{Z}}, P_{\vect{Z}} \right) \\
& = & \iint  \left(  
\mathsf{R}_{\vect{u}} \left( P_{\vect{\Theta} | \vect{Z} = \vect{z}} \right) - 
\mathsf{R}_{\vect{z}}\left( P_{\vect{\Theta} | \vect{Z} = \vect{z}} \right)
 \right) \mathrm{d}P_{\vect{Z}} \left( \vect{u} \right) \mathrm{d}P_{\vect{Z}} \left( \vect{z} \right)\Dsupersqueezeequ\IEEEeqnarraynumspace\\
\nonumber
\nonumber
& = &    \iint  \left( \mathsf{R}_{\vect{u}}\left( Q \right) -\lambda \KL{Q}{P^{\left(Q, \lambda\right)}_{\vect{\Theta}| \vect{Z} = \vect{u}}}\right) \mathrm{d}P_{\vect{Z}} \left( \vect{u} \right) \mathrm{d}P_{\vect{Z}} \left( \vect{z} \right) \\
\nonumber
&  & -  \iint  \left( \mathsf{R}_{\vect{z}}\left( Q \right) -\lambda \KL{Q}{P^{\left(Q, \lambda\right)}_{\vect{\Theta}| \vect{Z} = \vect{z}}} \right) \mathrm{d}P_{\vect{Z}} \left( \vect{u} \right) \mathrm{d}P_{\vect{Z}} \left( \vect{z} \right)\\
\nonumber
 &  & + \lambda \iint  \bigg( \KL{P_{\vect{\Theta} | \vect{Z} = \vect{z}}}{P^{\left(Q, \lambda\right)}_{\vect{\Theta}| \vect{Z} = \vect{u}}} \\
 \label{EqJune26at14h26in2024a}
 & & - \KL{P_{\vect{\Theta} | \vect{Z} = \vect{z}}}{P^{\left(Q, \lambda\right)}_{\vect{\Theta}| \vect{Z} = \vect{z}}} \bigg) \mathrm{d}P_{\vect{Z}} \left( \vect{u} \right) \mathrm{d}P_{\vect{Z}} \left( \vect{z} \right)\Dsupersqueezeequ\IEEEeqnarraynumspace\\
  &=  &  \lambda \iint  \bigg( \KL{P_{\vect{\Theta} | \vect{Z} = \vect{z}}}{P^{\left(Q, \lambda\right)}_{\vect{\Theta}| \vect{Z} = \vect{u}}} \\
 \label{EqJune26at14h26in2024b}
 & & - \KL{P_{\vect{\Theta} | \vect{Z} = \vect{z}}}{P^{\left(Q, \lambda\right)}_{\vect{\Theta}| \vect{Z} = \vect{z}}} \bigg) \mathrm{d}P_{\vect{Z}} \left( \vect{u} \right) \mathrm{d}P_{\vect{Z}} \left( \vect{z} \right),\Dsupersqueezeequ\IEEEeqnarraynumspace
\end{IEEEeqnarray}
which completes the proof.
\end{IEEEproof}

Theorem~\ref{TheoremJuin26at11h03in2024} expresses the generalization error~$\overline{\overline{\mathsf{G}}} \left(P_{\vect{\Theta} | \vect{Z}}, P_{\vect{Z}} \right)$ in~\eqref{EqJun4at9h02in2024} as the difference between the expectations of two relative entropies. The first relative entropy compares the algorithm~$P_{\vect{\Theta} | \vect{Z}}$ to the Gibbs algorithm~$P^{\left(Q, \lambda\right)}_{\vect{\Theta}| \vect{Z}}$, under the assumption that such algorithms are trained on independent datasets drawn from the same probability distribution~$P_{\vect{Z}}$. The second relative entropy compares these algorithms under the assumption that they are trained on the same dataset drawn from~$P_{\vect{Z}}$.
In particular, the first relative entropy satisfies the following property.
\begin{lemma}\label{LemmaSeptember4at17h56in2024}
Consider a probability measure~$P_{\vect{Z}}\in \triangle\left( \left( \set{X} \times \set{Y}\right)^n\right)$ and  a conditional probability measure~$P_{\vect{\Theta}|\vect{Z}} \in \triangle\left( \set{M} | \left( \set{X} \times \set{Y}\right)^n\right)$ and assume that for all~$\vect{z} \in \left( \set{X} \times \set{Y} \right)^n$,  
\begin{itemize}
\item The measures~$P_{\vect{\Theta}|\vect{Z} = \vect{z}}$ and~$P_{\vect{\Theta}}$ in~\eqref{EqJune26at16h54in2024} are absolutely continuous with respect to the $\sigma$-finite measure~$Q$ in~\eqref{EqGenpdf};
\item  The measure~$P_{\vect{\Theta}|\vect{Z} = \vect{z}}$ is absolutely continuous with respect to~$P_{\vect{\Theta}}$.
\end{itemize}
Then,
\begin{IEEEeqnarray}{rcl}
	\nonumber
& &	\iint \KL{P_{\vect{\Theta} | \vect{Z} = \vect{u}}}{P^{\left(Q, \lambda\right)}_{\vect{\Theta}| \vect{Z} = \vect{z}}} \mathrm{d}P_{\vect{Z}} \left( \vect{u} \right) \mathrm{d}P_{\vect{Z}} \left( \vect{z} \right)\\
	& = & \int \left( \KL{P_{\vect{\Theta}| \vect{Z} = \vect{z}}}{P_{\vect{\Theta}}} + \KL{P_{\vect{\Theta}}}{P^{\left(Q, \lambda\right)}_{\vect{\Theta}| \vect{Z} = \vect{z}}}  \right) \mathrm{d}P_{\vect{Z}} \left( \vect{z} \right),
 \Dsupersqueezeequ\IEEEeqnarraynumspace
\end{IEEEeqnarray}
where the probability measure~$P^{\left(Q, \lambda\right)}_{\vect{\Theta}| \vect{Z} = \vect{z}}$ satisfies~\eqref{EqGenpdf};  and~$\lambda \in \set{K}_{Q}$, with~$\set{K}_{Q}$ in~\eqref{EqSeptember3at18h10in2024}.
\end{lemma}
\begin{IEEEproof}
The proof follows by noticing that
\begin{IEEEeqnarray}{rcl}
	\nonumber
& &	\iint \KL{P_{\vect{\Theta} | \vect{Z} = \vect{u}}}{P^{\left(Q, \lambda\right)}_{\vect{\Theta}| \vect{Z} = \vect{z}}} \mathrm{d}P_{\vect{Z}} \left( \vect{u} \right) \mathrm{d}P_{\vect{Z}} \left( \vect{z} \right)\\
\nonumber
& = & 	\iint \int \log\left( \frac{\mathrm{d}P_{\vect{\Theta} | \vect{Z} = \vect{u}}}{\mathrm{d}P^{\left(Q, \lambda\right)}_{\vect{\Theta}| \vect{Z} = \vect{z}}} \left( \vect{\theta} \right) \right) \mathrm{d}P_{\vect{\Theta} | \vect{Z} = \vect{u}}   \left( \vect{\theta} \right) 
\mathrm{d}P_{\vect{Z}} \left( \vect{u} \right) \mathrm{d}P_{\vect{Z}} \left( \vect{z} \right)
 \Dsupersqueezeequ \\%\IEEEeqnarraynumspace\\
\nonumber
& = & 	\iint \int \log\left( \frac{\mathrm{d}P_{\vect{\Theta} }}{\mathrm{d}P^{\left(Q, \lambda\right)}_{\vect{\Theta}| \vect{Z} = \vect{z}}} \left( \vect{\theta} \right) \right) \mathrm{d}P_{\vect{\Theta} | \vect{Z} = \vect{u}}   \left( \vect{\theta} \right) 
\mathrm{d}P_{\vect{Z}} \left( \vect{u} \right) \mathrm{d}P_{\vect{Z}} \left( \vect{z} \right)
 \Dsupersqueezeequ\\
 \nonumber
 && + \iint \int \log\left( \frac{\mathrm{d}P_{\vect{\Theta} | \vect{Z} = \vect{u}}}{\mathrm{d}P_{\vect{\Theta}}} \left( \vect{\theta} \right) \right) \mathrm{d}P_{\vect{\Theta} | \vect{Z} = \vect{u}}   \left( \vect{\theta} \right) 
\mathrm{d}P_{\vect{Z}} \left( \vect{u} \right) \mathrm{d}P_{\vect{Z}} \left( \vect{z} \right)
 \Dsupersqueezeequ \\%\IEEEeqnarraynumspace
 \nonumber
& = & 	\iint \log\left( \frac{\mathrm{d}P_{\vect{\Theta} }}{\mathrm{d}P^{\left(Q, \lambda\right)}_{\vect{\Theta}| \vect{Z} = \vect{z}}} \left( \vect{\theta} \right) \right) \mathrm{d}P_{\vect{\Theta}}   \left( \vect{\theta} \right) 
 \mathrm{d}P_{\vect{Z}} \left( \vect{z} \right)
 \Dsupersqueezeequ\\
 && + \iint \log\left( \frac{\mathrm{d}P_{\vect{\Theta} | \vect{Z} = \vect{u}}}{\mathrm{d}P_{\vect{\Theta}}} \left( \vect{\theta} \right) \right) \mathrm{d}P_{\vect{\Theta} | \vect{Z} = \vect{u}}   \left( \vect{\theta} \right) 
\mathrm{d}P_{\vect{Z}} \left( \vect{u} \right) \\
 & = & \int \left( \KL{P_{\vect{\Theta}}}{P^{\left(Q, \lambda\right)}_{\vect{\Theta}| \vect{Z} = \vect{z}}} + \KL{P_{\vect{\Theta}| \vect{Z} = \vect{z}}}{P_{\vect{\Theta}}} \right) \mathrm{d}P_{\vect{Z}} \left( \vect{z} \right),
 \Dsupersqueezeequ
 \end{IEEEeqnarray}
 which completes the proof.
\end{IEEEproof}

\begin{figure}[t]
\begin{center}
\begin{tikzpicture}
\draw[thick] (0,0) -- node[below] {$\sqrt{\displaystyle\int  \KL{P_{\vect{\Theta}}}{P^{\left(Q, \lambda\right)}_{\vect{\Theta}| \vect{Z} = \vect{z}}} \mathrm{d}P_{\vect{Z}} \left( \vect{z} \right)}~$} ++(6,0);
\draw[thick] (0,0) -- node[left,rotate=90, above] {$\supersqueezeequ\sqrt{\displaystyle  \int  \KL{P_{\vect{\Theta}| \vect{Z} = \vect{z}}}{P_{\vect{\Theta}}} \mathrm{d}P_{\vect{Z}} \left( \vect{z} \right) }$} ++(0,4);
\draw[thick] (0,4) -- (6,0);
\node[left, rotate=-34, above] at (3,2) {$ \sqrt{ \frac{1}{\lambda} \overline{\overline{\mathsf{G}}}(P_{\vect{\Theta} | \vect{Z}},P_{\vect{Z}} ) + \int\KL{P_{\vect{\Theta} | \vect{Z} = \vect{z}}}{P^{\left(Q, \lambda\right)}_{\vect{\Theta}| \vect{Z} = \vect{z}}}  \mathrm{d}P_{\vect{Z}} \left( \vect{z} \right)}$};
\end{tikzpicture}
\end{center}
\caption{Geometric interpretation of  Lemma~\ref{LemmaSeptember4at17h56in2024} involving the generalization error~$\overline{\overline{\mathsf{G}}} \left(P_{\vect{\Theta} | \vect{Z}}, P_{\vect{Z}} \right)$ in~\eqref{EqJun4at9h02in2024}.}
\label{FigOctober20at17h43in2024atAgadir}
\end{figure}
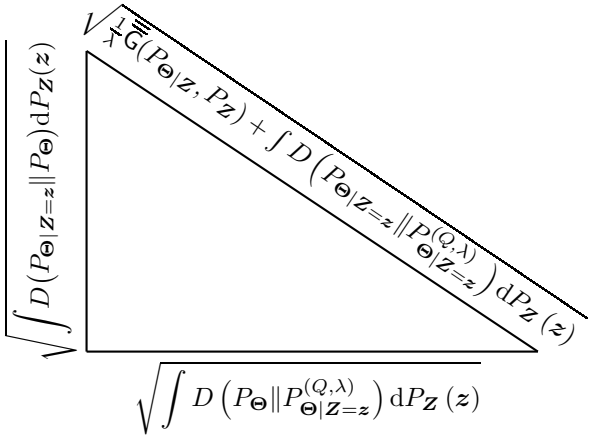

The converse of  the Pythagorean theorem~\cite[Book~I, Proposition~$48$]{heath1956thirteen} together with  Lemma~\ref{LemmaSeptember4at17h56in2024}, lead to the geometric construction shown in Figure~\ref{FigSeptember4at18h58in2024}. 
Such interpretation becomes more interesting by noticing that, from Theorem~\ref{TheoremJuin26at11h03in2024}, the square of the hypotenuse of the triangle in Figure~\ref{FigSeptember4at18h58in2024} satisfies
\begin{IEEEeqnarray}{rCl}
\nonumber
& & \iint \KL{P_{\vect{\Theta} | \vect{Z} = \vect{u}}}{P^{\left(Q, \lambda\right)}_{\vect{\Theta}| \vect{Z} = \vect{z}}} \mathrm{d}P_{\vect{Z}} \left( \vect{u} \right) \mathrm{d}P_{\vect{Z}} \left( \vect{z} \right)  = \frac{1}{\lambda} \overline{\overline{\mathsf{G}}}(P_{\vect{\Theta} | \vect{Z}},P_{\vect{Z}} ) \middlesqueezeequ\\
 & & + \int\KL{P_{\vect{\Theta} | \vect{Z} = \vect{z}}}{P^{\left(Q, \lambda\right)}_{\vect{\Theta}| \vect{Z} = \vect{z}}} \mathrm{d}P_{\vect{Z}} \left( \vect{z} \right), \squeezeequ
\end{IEEEeqnarray}
which involves the generalization error~$\overline{\overline{\mathsf{G}}} \left(P_{\vect{\Theta} | \vect{Z}}, P_{\vect{Z}} \right)$ in~\eqref{EqJun4at9h02in2024}; 
and 
the length~$\sqrt{\displaystyle \int  \KL{P_{\vect{\Theta}| \vect{Z} = \vect{z}}}{P_{\vect{\Theta}}} \mathrm{d}P_{\vect{Z}} \left( \vect{z} \right) }$ is the square root of a mutual information, i.e.,~$\sqrt{I\left(P_{\vect{\Theta}| \vect{Z}}; P_{\vect{Z}} \right)}$. This leads to the alternative interpretation shown in Figure~\ref{FigOctober20at17h43in2024atAgadir}.
%

%This interpretation unveils a number of geometric tools for the study of the generalization error~$\overline{\overline{\mathsf{G}}} \left(P_{\vect{\Theta} | \vect{Z}}, P_{\vect{Z}} \right)$ in~\eqref{EqJun4at9h02in2024}, offering new perspectives, which are discussed later in Section~\ref{SecGeometry}.
%
\begin{figure}[t]
\begin{center}
\begin{tikzpicture}
\draw[thick] (0,0) -- node[below] {$\sqrt{\displaystyle\int  \KL{P_{\vect{\Theta}}}{P^{\left(Q, \lambda\right)}_{\vect{\Theta}| \vect{Z} = \vect{z}}} \mathrm{d}P_{\vect{Z}} \left( \vect{z} \right)}~$} ++(6,0);
\draw[thick] (0,0) -- node[left,rotate=90, above] {$\supersqueezeequ\sqrt{\displaystyle \int  \KL{P_{\vect{\Theta}| \vect{Z} = \vect{z}}}{P_{\vect{\Theta}}} \mathrm{d}P_{\vect{Z}} \left( \vect{z} \right) }$} ++(0,4);
\draw[thick] (0,4) -- (6,0);
\node[left, rotate=-34, above] at (3,2) {$\supersqueezeequ \sqrt{\displaystyle \iint \KL{P_{\vect{\Theta} | \vect{Z} = \vect{u}}}{P^{\left(Q, \lambda\right)}_{\vect{\Theta}| \vect{Z} = \vect{z}}} \mathrm{d}P_{\vect{Z}} \left( \vect{u} \right) \mathrm{d}P_{\vect{Z}} \left( \vect{z} \right)}$};
\end{tikzpicture}
\end{center}
\caption{Geometric interpretation of  Lemma~\ref{LemmaSeptember4at17h56in2024} involving only  relative entropies. Note that~$ \int  \KL{P_{\vect{\Theta}| \vect{Z} = \vect{z}}}{P_{\vect{\Theta}}} \mathrm{d}P_{\vect{Z}} \left( \vect{z} \right) = I\left(P_{\vect{\Theta}| \vect{Z}}; P_{\vect{Z}} \right)$.}
\label{FigSeptember4at18h58in2024}
\end{figure}
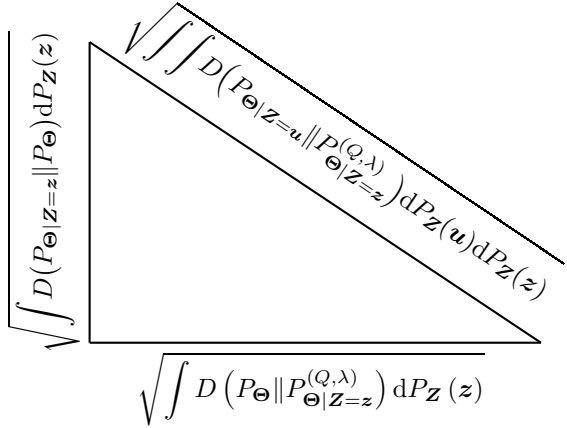  

\subsection{Generalization Error of the Gibbs Algorithm}\label{SecDecember21at9h33in2025HomeNice}

The generalization error of the Gibbs algorithm~$P^{\left(Q, \lambda\right)}_{\vect{\Theta}|\vect{Z}}$ in~\eqref{EqGenpdf},  can be immediately obtained from Theorem~\ref{TheoAugust9at11h15in2024} or Theorem~\ref{TheoremAugust11at11h02in2024}. The following corollary formalizes this observation.
\begin{corollary}\label{CorollaryOneMoreProofGenError}
The generalization error~$\overline{\overline{\mathsf{G}}} \left(P^{\left(Q, \lambda\right)}_{\vect{\Theta}|\vect{Z}}, P_{\vect{Z}} \right)$ in~\eqref{EqJun4at9h02in2024}, with the measure~$P^{\left(Q, \lambda\right)}_{\vect{\Theta}|\vect{Z}}$ in~\eqref{EqGenpdf}, is	
\begin{IEEEeqnarray}{rcl}
		\label{EqJanuary31at13h28}
		\overline{\overline{\mathsf{G}}}(P^{\left(Q, \lambda\right)}_{\vect{\Theta}| \vect{Z}},P_{\vect{Z}} ) & = &  \lambda \left( I\left( P^{\left(Q, \lambda\right)}_{\vect{\Theta}| \vect{Z}}; P_{\vect{Z}} \right) +  L\left( P^{\left(Q, \lambda\right)}_{\vect{\Theta}| \vect{Z}}; P_{\vect{Z}} \right) \right). \Dsupersqueezeequ \IEEEeqnarraynumspace
	\end{IEEEeqnarray}
\end{corollary}

\begin{figure}[t]
\begin{center}
\begin{tikzpicture}
\draw[thick] (0,0) -- node[below] {$\sqrt{ L\left(P^{\left(Q, \lambda\right)}_{\vect{\Theta}| \vect{Z} }; P_{\vect{Z}} \right) }~$} ++(6,0);
\draw[thick] (0,0) -- node[left,rotate=90, above] {$\sqrt{\displaystyle I\left(P^{\left(Q, \lambda\right)}_{\vect{\Theta}| \vect{Z} }; P_{\vect{Z}} \right) }$} ++(0,2);
\draw[thick] (0,2) -- (6,0);
\node[left, rotate=-20, above] at (3,1) {$ \sqrt{ \frac{1}{\lambda} \overline{\overline{\mathsf{G}}}(P^{\left(Q, \lambda\right)}_{\vect{\Theta}| \vect{Z} },P_{\vect{Z}} )}$};
\end{tikzpicture}
\end{center}
\caption{Geometric interpretation of Lemma~\ref{LemmaSeptember4at17h56in2024} for the Gibbs algorithm~$P^{\left(Q, \lambda\right)}_{\vect{\Theta}| \vect{Z}}$ (Definition~\ref{DefGibbsAlgorithm}).}
\label{FigOctober20at17h44in2024atAgadir}
\end{figure}
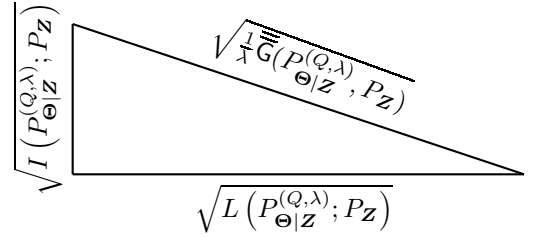

Corollary~\ref{CorollaryOneMoreProofGenError} has been proved numerous times before using several techniques. For instance, a proof for the case in which~$Q$ is a probability measure appears in \cite{aminian2021exact} and~\cite{aminian2024information}. A proof for the  case in which~$Q$ is a probability measure and datasets are formed by independent and identically distributed data points is presented in \cite{zou2024Generalization}. The more general case in which~$Q$ is a~$\sigma$-finite measure is proved in \cite{PerlazaTIT2024}.
Interestingly, independently of the choice of $\lambda$ and $Q$ in~\eqref{EqJanuary31at13h28}, such an expression does not simplify beyond the product of $\lambda$ and the sum of a mutual information and a lautum information. 
Figure~\ref{FigOctober20at17h44in2024atAgadir} shows a geometric interpretation of Corollary~\ref{CorollaryOneMoreProofGenError}, which is reminiscent of the Pythagorean theorem mentioned in the previous subsection.

Theorem~\ref{TheoremJuin26at11h03in2024} provides another expression for the generalization error of the Gibbs algorithm~$P^{\left(Q, \lambda\right)}_{\vect{\Theta}|\vect{Z}}$ in~\eqref{EqGenpdf}, as shown by the following corollary. 

\begin{corollary}\label{CorollaryOneMoreExpressionForGibs}
The generalization error~$\overline{\overline{\mathsf{G}}} \left(P^{\left(Q, \lambda\right)}_{\vect{\Theta}|\vect{Z}}, P_{\vect{Z}} \right)$ in~\eqref{EqJun4at9h02in2024}, with the measure~$P^{\left(Q, \lambda\right)}_{\vect{\Theta}|\vect{Z}}$ in~\eqref{EqGenpdf}, is
	\begin{IEEEeqnarray}{rcl}
\nonumber
\overline{\overline{\mathsf{G}}}(P^{\left(Q, \lambda\right)}_{\vect{\Theta}| \vect{Z}},P_{\vect{Z}} )
& = &  \lambda  \displaystyle\int \displaystyle\int  \KL{P^{\left(Q, \lambda\right)}_{\vect{\Theta}| \vect{Z}  = \vect{u}}}{P^{\left(Q, \lambda\right)}_{\vect{\Theta}| \vect{Z} = \vect{z}}} \mathrm{d} P_{\vect{Z}} \left( \vect{u} \right) \mathrm{d} P_{\vect{Z}} \left( \vect{z} \right). \Dsupersqueezeequ \\
\label{EqJun26at11h58in2024}
	\end{IEEEeqnarray}	
\end{corollary}
Corollary~\ref{CorollaryOneMoreExpressionForGibs} shows that the generalization error of the Gibbs algorithm~$P^{\left(Q, \lambda\right)}_{\vect{\Theta}|\vect{Z}}$ in~\eqref{EqGenpdf} is equal to the average (expectation with respect to~$P_{\vect{Z}}  P_{\vect{Z}}~$) of a pair-wise comparison via relative entropy of all the instances of such an algorithm. 
The expression in~\eqref{EqJun26at11h58in2024} appears to be the simplest form in which the generalization error $\overline{\overline{\mathsf{G}}}(P^{\left(Q, \lambda\right)}_{\vect{\Theta}| \vect{Z}},P_{\vect{Z}} )$ can be written. As in the previous case, Corollary~\ref{CorollaryOneMoreProofGenError}, the choice of $\lambda$ and $Q$ do not appear to lead to further simplification.

\section{Explicit Expressions Obtained Via the Method of Data-Driven Gaps}\label{SecExpressionsViaDataDrivenGaps}

This section focuses on constructing novel explicit expressions for the generalization error~$\overline{\overline{\mathsf{G}}} \left(P_{\vect{\Theta} | \vect{Z}}, P_{\vect{Z}} \right)$ in~\eqref{EqJun4at9h02in2024} by exploring different choices of the parameters~$P_S$ and~$\beta$ in~\eqref{EqOctober19at18h32in2024InTheAirFlyingToAgadir}.
It is important to highlight that in these expressions, the central building-block is the conditional WCDG probability measure~$P_{\hat{Z} | \vect{\Theta}}^{\left( P_{S}, \beta \right)} \in \triangle\left( \set{X} \times \set{Y} | \set{M} \right)$ in~\eqref{Eqtheworstgeneral} (Definition~\ref{DefWCDG}). 
As in Section V, the identities derived in this section are not proposed as direct computational tools for evaluating the generalization error. Instead, they provide an exact data-driven representation of the generalization error by using the WCDG probability measure as a reference object.  Thus, the contribution of the data-driven method of gaps is conceptual: it reveals the information-theoretic structure of the generalization error through comparisons with the WCDG probability measure.

The first three subsections provide explicit expressions for the generalization error~$\overline{\overline{\mathsf{G}}} \left(P_{\vect{\Theta} | \vect{Z}}, P_{\vect{Z}} \right)$ aiming at establishing connections to existing results in statistics (statistical hypothesis testing), information theory (information measures), and Euclidean geometry (Pythagorean Theorems), respectively.
%
%The last section focuses on the algorithm resulting from the WCDG probability measure.

\subsection{Connections to Hypothesis Testing}\label{SecDecember26at10h48in2025HomeNice}

A first connection to statistical hypothesis testing is made by the following theorem.

\begin{theorem}\label{TheoremMismatchP_S}
Consider the generalization error~$\overline{\overline{\mathsf{G}}}(P_{\vect{\Theta} | \vect{Z}}, P_{\vect{Z}} )$ in~\eqref{EqJun4at9h02in2024} and assume that: 
\begin{itemize}
\item[$(a)$] The probability measure~$P_{\vect{Z}}$ satisfies~\eqref{EqSaturdayMai20at13h20in2024} for some~$P_{Z} \in \triangle\left( \set{X} \times \set{Y}\right)$; 
\item[$(b)$]  For all~$\vect{z} \in \left( \set{X} \times \set{Y} \right)^n$, the probability measures~$P_{\vect{\Theta} | \vect{Z} = \vect{z}}$ and~$P_{\vect{\Theta}}$ in~\eqref{EqJune26at16h54in2024} are mutually absolutely continuous; 
\item[$(c)$]  For all~$\vect{\theta} \in \set{M}$, the probability measures~$P_{\vect{Z} | \vect{\Theta} = \vect{\theta}}$ and~$P_{\vect{Z}}$ in~\eqref{EqSaturdayJan27in2024b} are mutually absolutely continuous; and 
\item[$(d)$] For all~$\vect{\theta} \in \set{M}$,  the probability measures~$P_{Z | \vect{\Theta} = \vect{\theta}}$ and~$P_{Z}$  satisfy~\eqref{EqOctober17at14h40in2024} and are both mutually absolutely continuous with the probability measure~$P_S \in \triangle\left( \set{X} \times \set{Y} \right)$ in~\eqref{Eqtheworstgeneral}. 
\end{itemize}
Then, 
	\begin{IEEEeqnarray}{rCl}
		\nonumber
		&&\overline{\overline{\mathsf{G}}}(P_{\vect{\Theta} | \vect{Z}}, P_{\vect{Z}} )  \\
		\nonumber
		&=& - \beta \Bigg(\iint \log{\left(\frac{\mathrm{d}P_{\hat{Z} | \vect{\Theta} = \vect{\theta}}^{\left( P_{S}, \beta \right)} }{\mathrm{d}P_S}(z)\right)} \mathrm{d}P_{Z|\vect{\Theta} = \vect{\theta}}(z)\mathrm{d}P_{\vect{\Theta}}\left(\vect{\theta}\right) \\
		\label{EqMismatchP_S}
		&&+  \iint \log{\left(\frac{\mathrm{d}P_{\hat{Z} | \vect{\Theta} = \vect{\theta}}^{\left( P_{S}, \beta \right)} }{\mathrm{d}P_S}(z)\right)} \mathrm{d}P_{Z}(z)\mathrm{d}P_{\vect{\Theta}}\left(\vect{\theta}\right)\Bigg),
	\end{IEEEeqnarray}
	where the probability measure~$P_{\hat{Z} | \vect{\Theta} = \vect{\theta}}^{\left( P_{S}, \beta \right)}$ is  the WCDG probability measure in~\eqref{Eqtheworstgeneral}; and~$\beta \in \set{J}_{P_S}$, with~$\set{J}_{P_S}$ in~\eqref{EqOctober17at21h21in2024}.
\end{theorem}
\begin{IEEEproof}
Under Assumptions~$(a)$,~$(b)$,~$(c)$, and~$(d)$, Lemma~\ref{LemmaGinD} holds. Then, the  proof follows by observing that the differences~$\KL{P_{Z|\vect{\Theta}=\vect{\theta}}}{P_{\hat{Z} | \vect{\Theta} = \vect{\theta}}^{\left( P_{S}, \beta \right)} } - \KL{P_{Z|\vect{\Theta}=\vect{\theta}}}{P_S}$ and~$\KL{P_Z}{P_S} -\KL{P_Z}{P_{\hat{Z} | \vect{\Theta} = \vect{\theta}}^{\left( P_{S}, \beta \right)} }$ in~\eqref{EqOctober19at18h32in2024InTheAirFlyingToAgadir} respectively satisfy the following equalities:
	\begin{IEEEeqnarray}{rCl} 
		\nonumber
	&&	\KL{P_{Z|\vect{\Theta}=\vect{\theta}}}{P_{\hat{Z} | \vect{\Theta} = \vect{\theta}}^{\left( P_{S}, \beta \right)} } - \KL{P_{Z|\vect{\Theta}=\vect{\theta}}}{P_S}\\
		&=& \int \Bigl(\log{\frac{\mathrm{d}P_{Z|\vect{\Theta}=\vect{\theta}}}{\mathrm{d}P_{\hat{Z} | \vect{\Theta} = \vect{\theta}}^{\left( P_{S}, \beta \right)} }\left(z\right)} -\log{\frac{\mathrm{d}P_{Z|\vect{\Theta}=\vect{\theta}}}{\mathrm{d}P_S}\left(z\right)}  \Bigr) \mathrm{d} P_{Z|\vect{\Theta}=\vect{\theta}}(z) \squeezeequ\spnum\\
		\label{equalityusingcontinuity}
		&=& \int \log{\frac{\mathrm{d}P_S}{\mathrm{d}P_{\hat{Z} | \vect{\Theta} = \vect{\theta}}^{\left( P_{S}, \beta \right)} }\left(z\right)} \mathrm{d}P_{Z|\vect{\Theta}=\vect{\theta}}\left(z\right)\\
\label{EqDecember23at14h55in2025Sophia}
		&=&- \int \log{\frac{\mathrm{d}P_{\hat{Z} | \vect{\Theta} = \vect{\theta}}^{\left( P_{S}, \beta \right)} }{\mathrm{d}P_S}\left(z\right)} \mathrm{d}P_{Z|\vect{\Theta}=\vect{\theta}}\left(z\right),
	\end{IEEEeqnarray}
	where the equality in~\eqref{equalityusingcontinuity} follows from the Assumption~$(d)$ and~\cite[Theorem 4]{InriaRR9591}; and the equality in~\eqref{EqDecember23at14h55in2025Sophia} follows from~\cite[Theorem 5]{InriaRR9591}.
	Similarly,
	\begin{IEEEeqnarray}{rCl}
		\nonumber
&&\KL{P_Z}{P_S} -\KL{P_Z}{P_{\hat{Z} | \vect{\Theta} = \vect{\theta}}^{\left( P_{S}, \beta \right)} }\\
&=& \int \Bigl(\log{\frac{\mathrm{d}P_Z}{\mathrm{d}P_S}(z) - \log\frac{\mathrm{d}P_Z}{\mathrm{d}P_{\hat{Z} | \vect{\Theta} = \vect{\theta}}^{\left( P_{S}, \beta \right)} }(z) }\Bigr)\mathrm{d}P_Z(z)\\
\label{equalityusingcontinuity2}
&=&-\int \log{\frac{\mathrm{d}P_S}{\mathrm{d}P_{\hat{Z} | \vect{\Theta} = \vect{\theta}}^{\left( P_{S}, \beta \right)} }\left(z\right)} \mathrm{d}P_{Z}\left(z\right)\\
\label{EqDecember23at14h58in2025Sophia}
&=&\int \log{\frac{\mathrm{d}P_{\hat{Z} | \vect{\Theta} = \vect{\theta}}^{\left( P_{S}, \beta \right)} }{\mathrm{d}P_S}\left(z\right)} \mathrm{d}P_{Z}\left(z\right),
	\end{IEEEeqnarray}
where the equality in~\eqref{equalityusingcontinuity2} follows from Assumption~$(d)$ and~\cite[Theorem 4]{InriaRR9591}; and the equality in~\eqref{EqDecember23at14h58in2025Sophia} follows from~\cite[Theorem 5]{InriaRR9591}.
	
Plugging both~\eqref{equalityusingcontinuity} and~\eqref{equalityusingcontinuity2} into~\eqref{EqOctober19at18h32in2024InTheAirFlyingToAgadir} yields~\eqref{EqMismatchP_S}, which completes the proof.
\end{IEEEproof}

Theorem~\ref{TheoremMismatchP_S} is reminiscent of Theorem~\ref{TheoAugust9at11h15in2024}, and thus, it admits a similar analysis in terms of mismatched hypothesis tests~\cite{boroumand2022mismatched,pmlr-v119-dutta20a}. 
Consider for instance a hypothesis test in which the objective is to determine whether an observed data point $z \in \set{X} \times \set{Y}$ is generated  from either the WCDG probability measure given a model $\vect{\theta}$, i.e., the measure~$P_{\hat{Z} | \vect{\Theta} = \vect{\theta}}^{\left( P_{S}, \beta \right)}  \in \triangle\left( \set{X} \times \set{Y} \right)$ in~\eqref{Eqtheworstgeneral} ($H_0$, null hypothesis); or from the reference probability measure~$P_S \in \simplex{\set{X} \times \set{Y}}$, regardless of the model $\vect{\theta}$, ($H_1$, alternative hypothesis): 
\begin{IEEEeqnarray}{c}
\label{EqDecember22at16h54in2025BusToNice}
\left\lbrace
\begin{array}{rcl}
H_0: & &z  \sim P_{\hat{Z} | \vect{\Theta} = \vect{\theta}}^{\left( P_{S}, \beta \right)}\\
H_1: & &z  \sim P_S.
\end{array}
\right.
\end{IEEEeqnarray}
As in Section~\ref{SecDecember22at16h52in2025BusToNice}, the hypothesis test in~\eqref{EqDecember22at16h54in2025BusToNice} is said to be mismatched because none of the hypotheses matches the assumptions on the probability distribution of the observations.  
More specifically, the hypothesis test in~\eqref{EqDecember22at16h54in2025BusToNice} is studied in two different settings. 
First, the pair~$\left( z , \vect{\theta}\right)$ is assumed to be sampled from the joint probability measure $P_{Z| \vect{\Theta}} P_{\vect{\Theta}}$,  with~$P_{\vect{\Theta}}$ in~\eqref{EqJune26at16h54in2024} and~$P_{Z | \vect{\Theta}}$ in~\eqref{EqOctober17at14h40in2024}. 
Second,   the observation~$\left( z, \vect{\theta} \right)$ is assumed to be sampled from the product measure $P_{Z} P_{\vect{\Theta}}$.
Regardless of such scenarios, the log-likelihood ratio to decide based upon the pair~$\left( z, \vect{\theta}\right)$ on the null or alternative hypothesis in \eqref{EqDecember22at16h54in2025BusToNice} is~$\log \frac{\mathrm{d}P_{\hat{Z} | \vect{\Theta} = \vect{\theta}}^{\left( P_{S}, \beta \right)}}{\mathrm{d} P_S} \left( z \right)$. 
Essentially, given an~$\eta \geqslant 0$, if~$\log \frac{\mathrm{d}P_{\hat{Z} | \vect{\Theta} = \vect{\theta}}^{\left( P_{S}, \beta \right)}}{\mathrm{d} P_S} \left( z \right) \geqslant \eta$, then the null hypothesis is accepted. 
Alternatively, the null hypothesis is rejected if~$\log \frac{\mathrm{d}P_{\hat{Z} | \vect{\Theta} = \vect{\theta}}^{\left( P_{S}, \beta \right)}}{\mathrm{d} P_S} \left( z \right) < \eta$.
From these assumptions, the acceptance region of the null hypothesis is the set
\begin{IEEEeqnarray}{rcl}
\label{EqDecember22at19h53in2025HomeNice}
\set{C}_{\eta} & = &\left\lbrace  \left(z,  \vect{\theta} \right) \in \left( \set{X} \times \set{Y} \right) \times \set{M}  : \log \frac{\mathrm{d}P_{\hat{Z} | \vect{\Theta} = \vect{\theta}}^{\left( P_{S}, \beta \right)}}{\mathrm{d} P_S} \left( z \right) \geqslant \eta \right\rbrace. \squeezeequ \IEEEeqnarraynumspace
\end{IEEEeqnarray}
That is, given an observation $\left( z, \vect{\theta} \right)$, the null hypothesis is accepted if $\left( z, \vect{\theta} \right) \in \set{C}_{\eta}$; or rejected otherwise, i.e., $\left( z, \vect{\theta} \right) \in \set{C}_{\eta}^{\sfc}$.
Consider the following probabilities
\begin{IEEEeqnarray}{rcl}
\label{EqDecember23at06h46in2025BusToSophiaA}
P_{Z | \vect{\Theta}} P_{\vect{\Theta}} \left( \set{C}_\eta \right) & = & 1 - P_{Z | \vect{\Theta} } P_{\vect{\Theta}} \left( \set{C}_{\eta}^{\sfc} \right), \mbox{ and }\\
\label{EqDecember23at06h46in2025BusToSophiaB}
P_{Z}P_{\vect{\Theta}}  \left( \set{C}_{\eta} \right) & = & 1 - P_{Z}  P_{\vect{\Theta} }  \left( \set{C}_{\eta}^{\sfc} \right).
\end{IEEEeqnarray}
The term $P_{Z | \vect{\Theta} } P_{\vect{\Theta}} \left( \set{C}_{\eta} \right)$ represents the probability of choosing $H_0$ under the assumption that observations $\left(z,  \vect{\theta} \right)$ are obtained by sampling $P_{Z | \vect{\Theta} } P_{\vect{\Theta}}$.
Alternatively, the term $P_{Z} P_{\vect{\Theta}} \left( \set{C}_{\eta} \right)$ represents the probability of choosing $H_0$ under the assumption that observations  $\left( z, \vect{\theta} \right)$ are sampled from $P_{Z}P_{\vect{\Theta}}$. 
Consider also the following constants
\begin{IEEEeqnarray}{rcl}
\label{EqDecember23at07h02in2025BusToSophiaA}
\bar{\eta} & \triangleq & \sup \left\lbrace \log \frac{\mathrm{d} P_{\hat{Z} | \vect{\Theta} = \vect{\theta}}^{\left( P_{S}, \beta \right)}}{\mathrm{d} P_S} \left( z \right) \in \reals  : \left( z, \vect{\theta} \right)  \in  \left( \set{X} \times \set{Y} \right) \times \set{M} \right\rbrace. \squeezeequ\spnum
\end{IEEEeqnarray}
and 
\begin{IEEEeqnarray}{rcl}
\label{EqDecember23at07h02in2025BusToSophiaB}
\underline{\eta} & \triangleq & \inf \left\lbrace \log \frac{\mathrm{d}P_{\hat{Z} | \vect{\Theta} = \vect{\theta}}^{\left( P_{S}, \beta \right)}}{\mathrm{d} P_S} \left( z \right) \in \reals  : \left( z, \vect{\theta} \right)  \in  \left( \set{X} \times \set{Y} \right) \times \set{M} \right\rbrace. \squeezeequ\spnum
\end{IEEEeqnarray}
The reference measure $P_{S}$ and the WCDG probability measure $P_{\hat{Z} | \vect{\Theta} = \vect{\theta}}^{\left( P_{S}, \beta \right)}$ in the hypothesis test in~\eqref{EqDecember22at16h54in2025BusToNice}, are both exogenous to the study of  the generalization error~$\overline{\overline{\mathsf{G}}}(P_{\vect{\Theta} | \vect{Z}}, P_{\vect{Z}})$ in~\eqref{EqJun4at9h02in2024}. Nonetheless, a strong connection exists as shown by the following theorem.
\begin{theorem}\label{TheoDecember23at7h51in2025atSophia}
Consider the generalization error~$\overline{\overline{\mathsf{G}}}(P_{\vect{\Theta} | \vect{Z}}, P_{\vect{Z}} )$ in~\eqref{EqJun4at9h02in2024} and assume that: 
\begin{itemize}
\item[$(a)$] The probability measure~$P_{\vect{Z}}$ satisfies~\eqref{EqSaturdayMai20at13h20in2024} for some~$P_{Z} \in \triangle\left( \set{X} \times \set{Y}\right)$; 
\item[$(b)$]  For all~$\vect{z} \in \left( \set{X} \times \set{Y} \right)^n$, the probability measures~$P_{\vect{\Theta} | \vect{Z} = \vect{z}}$ and~$P_{\vect{\Theta}}$ in~\eqref{EqJune26at16h54in2024} are mutually absolutely continuous; 
\item[$(c)$]  For all~$\vect{\theta} \in \set{M}$, the probability measures~$P_{\vect{Z} | \vect{\Theta} = \vect{\theta}}$ and~$P_{\vect{Z}}$ in~\eqref{EqSaturdayJan27in2024b} are mutually absolutely continuous; and 
\item[$(d)$] For all~$\vect{\theta} \in \set{M}$,  the probability measures~$P_{Z | \vect{\Theta} = \vect{\theta}}$ and~$P_{Z}$  satisfy~\eqref{EqOctober17at14h40in2024} and are both mutually absolutely continuous with the probability measure~$P_S \in \triangle\left( \set{X} \times \set{Y} \right)$ in~\eqref{Eqtheworstgeneral}. 
\end{itemize}
Then,  
\begin{IEEEeqnarray}{C}
\nonumber
- \beta \left(  \bar{\eta}  - \eta \right) P_{Z | \vect{\Theta} } P_{\vect{\Theta}} \left( \set{C}_{\eta} \right)  - \beta \left( \eta - \underline{\eta} \right) P_{Z}P_{\vect{\Theta}} \left( \set{C}_{\eta}^{\sfc} \right)\\ 
\nonumber
 \leqslant \overline{\overline{\mathsf{G}}} \left(P_{\vect{\Theta} | \vect{Z}}, P_{\vect{Z}} \right) \leqslant\\
\label{EqDecember23at7h53in2025atSophia}
\beta \left(  \eta -  \underline{\eta} \right) P_{Z | \vect{\Theta} } P_{\vect{\Theta}} \left( \set{C}_{\eta}^{\sfc}\right)  + \beta \left( \bar{\eta}  -  \eta \right) P_{Z}P_{\vect{\Theta}} \left( \set{C}_{\eta} \right), \spnum 
\end{IEEEeqnarray}
where $\eta$ satisfies
\begin{IEEEeqnarray}{rCl}
\underline{\eta} & \leqslant \eta \leqslant & \bar{\eta},
\end{IEEEeqnarray}
and $\bar{\eta}$ and $\underline{\eta}$ are respectively defined in~\eqref{EqDecember23at07h02in2025BusToSophiaA} and~\eqref{EqDecember23at07h02in2025BusToSophiaB}; the set $\set{C}_{\eta}$ is defined in~\eqref{EqDecember22at19h53in2025HomeNice}; and the measures $P_{Z | \vect{\Theta}} P_{\vect{\Theta}}$ and $P_{Z}P_{\vect{\Theta}}$ are both in $\simplex{\left( \set{X} \times \set{Y}\right) \times \set{M}}$. 
\end{theorem}
\begin{IEEEproof}
The proof follows the same steps as those in the proof of Theorem~\ref{TheoNovember29at13h20in2025HomeNice}.
\end{IEEEproof} 
The upper and lower bounds in Theorem~\ref{TheoDecember23at7h51in2025atSophia} can be optimized by strategically choosing the parameter $\beta$  and $P_S$ of the conditional Gibbs probability measure $P_{\hat{Z} | \vect{\Theta}}^{\left( P_{S}, \beta \right)} \in \simplex{\set{X} \times \set{Y} | \set{M}}$; and the nonnegative real $\eta$. 
Nonetheless, the interest of Theorem~\ref{TheoDecember23at7h51in2025atSophia} lies in the fact that it unveils a connection between the generalization error~$\overline{\overline{\mathsf{G}}}(P_{\vect{\Theta} | \vect{Z}}, P_{\vect{Z}} )$ in~\eqref{EqJun4at9h02in2024} and the hypothesis test in \eqref{EqDecember22at16h54in2025BusToNice}, in which both hypotheses are determined by the WCDG probability measure $P_{\hat{Z} | \vect{\Theta} = \vect{\theta}}^{\left( P_{S}, \beta \right)}$ and its corresponding reference measure $P_S$.

\subsection{Connections to Information Measures} 
  
The following theorem is reminiscent of Theorem~\ref{TheoremAugust11at11h02in2024}, which provides an expression for the generalization error~$\overline{\overline{\mathsf{G}}}(P_{\vect{\Theta} | \vect{Z}},P_{\vect{Z}} )$ in~\eqref{EqJun4at9h02in2024}  involving the sum of a mutual and a lautum information between models and data points.

\begin{theorem}\label{Corollarycomparisonnegative}
Consider the generalization error~$\overline{\overline{\mathsf{G}}}(P_{\vect{\Theta} | \vect{Z}}, P_{\vect{Z}} )$ in~\eqref{EqJun4at9h02in2024} and assume that: 
\begin{itemize}
\item[$(a)$] The probability measure~$P_{\vect{Z}}$ satisfies~\eqref{EqSaturdayMai20at13h20in2024} for some~$P_{Z} \in \triangle\left( \set{X} \times \set{Y}\right)$; 
\item[$(b)$]  For all~$\vect{z} \in \left( \set{X} \times \set{Y} \right)^n$, the probability measures~$P_{\vect{\Theta} | \vect{Z} = \vect{z}}$ and~$P_{\vect{\Theta}}$ in~\eqref{EqJune26at16h54in2024} are mutually absolutely continuous; 
\item[$(c)$]  For all~$\vect{\theta} \in \set{M}$, the probability measures~$P_{\vect{Z} | \vect{\Theta} = \vect{\theta}}$ and~$P_{\vect{Z}}$ in~\eqref{EqSaturdayJan27in2024b} are mutually absolutely continuous; and 
\item[$(d)$] For all~$\vect{\theta} \in \set{M}$,  the probability measures~$P_{Z | \vect{\Theta} = \vect{\theta}}$ and~$P_{Z}$  satisfy~\eqref{EqOctober17at14h40in2024} and are both mutually absolutely continuous with the probability measure~$P_S \in \triangle\left( \set{X} \times \set{Y} \right)$ in~\eqref{Eqtheworstgeneral}. 
\end{itemize}
Then,
	\begin{IEEEeqnarray}{rCl}
		\nonumber
		&&\overline{\overline{\mathsf{G}}}(P_{\vect{\Theta} | \vect{Z}},P_{\vect{Z}} )= - \beta \left(  I\left(P_{Z | \vect{\Theta}}; P_{\vect{\Theta}} \right)  +  L\left(P_{Z | \vect{\Theta}}; P_{\vect{\Theta}} \right) \right)\squeezeequ\\
		\nonumber
		&&- \beta \iint \log{\left(\frac{\mathrm{d}P_{\hat{Z} | \vect{\Theta} = \vect{\theta}}^{\left( P_{S}, \beta \right)} }{\mathrm{d}P_{Z|\vect{\Theta} = \vect{\theta}}}(z)\right)} \mathrm{d}P_{Z|\vect{\Theta} = \vect{\theta}}\left(z\right)\mathrm{d}P_{\vect{\Theta}}\left(\vect{\theta}\right)\\
		\label{EqNovember17at11h16in2024Nice}
		&&+\beta \iint \log{\left(\frac{\mathrm{d}P_{\hat{Z} | \vect{\Theta} = \vect{\theta}}^{\left( P_{S}, \beta \right)} }{\mathrm{d}P_{Z|\vect{\Theta} = \vect{\theta}}}(z)\right)} \mathrm{d}P_{Z}\left(z\right)\mathrm{d}P_{\vect{\Theta}}\left(\vect{\theta}\right),\spnum
	\end{IEEEeqnarray}
	where the probability measure~$P_{\hat{Z} | \vect{\Theta} = \vect{\theta}}^{\left( P_{S}, \beta \right)}~$ is defined in~\eqref{Eqtheworstgeneral};~$\beta \in \set{J}_{P_S}$, with~$\set{J}_{P_S}$ in~\eqref{EqOctober17at21h21in2024}. 
\end{theorem}
	\begin{IEEEproof}
The proof follows from Lemma~\ref{LemmaGinD}, which holds under Assumptions~$(a)$,~$(b)$,~$(c)$, and~$(d)$.  More specifically, the proof is divided into two parts. The first part focuses on the difference $\int\left(\KL{P_{Z | \vect{\Theta} = \vect{\theta}}}{P_S} - \KL{P_Z}{P_S}  \right) \mathrm{d} P_{\vect{\Theta}}\left( \vect{\theta} \right)$ in~\eqref{EqOctober19at18h32in2024InTheAirFlyingToAgadir}. Following similar steps as in the proof of Theorem~\ref{TheoremAugust11at11h02in2024}, it can be verified that
\begin{IEEEeqnarray}{rCl}
\nonumber
& &\int\left(\KL{P_{Z | \vect{\Theta} = \vect{\theta}}}{P_S} - \KL{P_Z}{P_S}  \right) \mathrm{d} P_{\vect{\Theta}}\left( \vect{\theta} \right) \\
\label{EqNovember13at07h06in2024InTheBusToSophia}
& = & I\left(P_{Z | \vect{\Theta}}; P_{\vect{\Theta}} \right).    \spnum
\end{IEEEeqnarray}
Using \eqref{EqNovember13at07h06in2024InTheBusToSophia} in~\eqref{EqOctober19at18h32in2024InTheAirFlyingToAgadir} yields
\begin{IEEEeqnarray}{rcl}
		\nonumber
& &		\overline{\overline{\mathsf{G}}}(P_{\vect{\Theta} | \vect{Z}},P_{\vect{Z}} ) \\
		\nonumber
&=& - \beta   I\left( P_{ Z | \vect{\Theta}}; P_{\vect{\Theta}} \right) + \beta \int \KL{P_{Z|\vect{\Theta} = \vect{\theta}}}{P_{\hat{Z} | \vect{\Theta} = \vect{\theta}}^{\left( P_{S}, \beta \right)} } \mathrm{d} P_{\vect{\Theta}} \left( \vect{\theta} \right) \\
\label{EqOctober21at12h21in2024Agadir}		
		&& - \beta \int \KL{P_{Z}}{P_{\hat{Z} | \vect{\Theta} = \vect{\theta}}^{\left( P_{S}, \beta \right)} } \mathrm{d} P_{\vect{\Theta}} \left( \vect{\theta} \right).
	\end{IEEEeqnarray}
The second part of the proof focuses on the term $\int \KL{P_{Z}}{P_{\hat{Z} | \vect{\Theta} = \vect{\theta}}^{\left( P_{S}, \beta \right)} } \mathrm{d} P_{\vect{\Theta}} \left( \vect{\theta} \right)$ in~\eqref{EqOctober21at12h21in2024Agadir}, which satisfies
	\begin{IEEEeqnarray}{rCl}
		\nonumber
		&& \int \KL{P_{Z}}{P_{\hat{Z} | \vect{\Theta} = \vect{\theta}}^{\left( P_{S}, \beta \right)} } \mathrm{d} P_{\vect{\Theta}} \left( \vect{\theta} \right) \\
\label{EqDecember23at10h47in2025Sophia}		
		&=&\iint \log{\left(\frac{\mathrm{d}P_{Z}}{\mathrm{d}P_{\hat{Z} | \vect{\Theta} = \vect{\theta}}^{\left( P_{S}, \beta \right)} }(z)\frac{\mathrm{d}P_{Z|\vect{\Theta} = \vect{\theta}}}{\mathrm{d}P_{Z|\vect{\Theta} =\vect{\theta}}}\left(z\right)\right)}\mathrm{d}P_{Z}(z)\mathrm{d}P_{\vect{\Theta}}\left(\vect{\theta}\right)\supersqueezeequ \spnum\\
		\nonumber
		&=& \int \Bigg(\int \log{\left(\frac{\mathrm{d}P_Z}{\mathrm{d}P_{Z|\vect{\Theta}=\vect{\theta}}}(z)\right)}\mathrm{d}P_{Z}(z) \\
		\label{equalityusingcintinuity3}
		&&+  \int \log{\left(\frac{\mathrm{d}P_{Z|\vect{\Theta} = \vect{\theta}}}{\mathrm{d}P_{\hat{Z} | \vect{\Theta} = \vect{\theta}}^{\left( P_{S}, \beta \right)} }(z)\right)} \mathrm{d}P_{Z}\left(z\right)\Bigg)\mathrm{d}P_{\vect{\Theta}}\left(\vect{\theta}\right)\\
		\nonumber
		&=& L\left( P_{Z| \vect{\Theta}}; P_{\vect{\Theta}} \right)\\
		\label{EqOctober21at12h40in2024Agadir}
		&&+ \iint\log{\left(\frac{\mathrm{d}P_{Z|\vect{\Theta} = \vect{\theta}}}{\mathrm{d}P_{\hat{Z} | \vect{\Theta} = \vect{\theta}}^{\left( P_{S}, \beta \right)} }(z)\right)} \mathrm{d}P_{Z}\left(z\right)\mathrm{d}P_{\vect{\Theta}}\left(\vect{\theta}\right),
	\end{IEEEeqnarray}
where the equality in~\eqref{EqDecember23at10h47in2025Sophia} follows from~\cite[Theorem 3]{InriaRR9591};  the equality in~\eqref{equalityusingcintinuity3} follows from~\cite[Theorem 4]{InriaRR9591} and Lemma~\ref{EqNovember13at17h04in2024InTheBusToNice} (in Appendix~\ref{AppendixMiscellanea}). 
The proof ends by using the equality~\eqref{EqOctober21at12h40in2024Agadir} in~\eqref{EqOctober21at12h21in2024Agadir}, which yields~\eqref{EqNovember17at11h16in2024Nice}.
\end{IEEEproof}

In Theorem~\ref{Corollarycomparisonnegative}, the term 
	\begin{IEEEeqnarray}{rCl}
		\nonumber
		&&- \beta \iint \log{\left(\frac{\mathrm{d}P_{\hat{Z} | \vect{\Theta} = \vect{\theta}}^{\left( P_{S}, \beta \right)} }{\mathrm{d}P_{Z|\vect{\Theta} = \vect{\theta}}}(z)\right)} \mathrm{d}P_{Z|\vect{\Theta} = \vect{\theta}}\left(z\right)\mathrm{d}P_{\vect{\Theta}}\left(\vect{\theta}\right)\\
		&&+\beta \iint \log{\left(\frac{\mathrm{d}P_{\hat{Z} | \vect{\Theta} = \vect{\theta}}^{\left( P_{S}, \beta \right)} }{\mathrm{d}P_{Z|\vect{\Theta} = \vect{\theta}}}(z)\right)} \mathrm{d}P_{Z}\left(z\right)\mathrm{d}P_{\vect{\Theta}}\left(\vect{\theta}\right),\spnum
	\end{IEEEeqnarray}
admits a similar analysis as the term in \eqref{EqAugust25at15h31in2024} in the context of a hypothesis test.  In this case, the hypothesis test consists in determining whether a data point $z$ has been drawn from the WCDG probability distribution $P_{\hat{Z} | \vect{\Theta} = \vect{\theta}}^{\left( P_{S}, \beta \right)}$ or from the likelihood of such a data point $z$ given the observation of the model $\vect{\theta}$. This can be formulated as follows
 \begin{IEEEeqnarray}{c}
\label{EqDecember23at14h42in2025Sophia}
\left\lbrace
\begin{array}{rcl}
H_0: & &z  \sim P_{\hat{Z} | \vect{\Theta} = \vect{\theta}}^{\left( P_{S}, \beta \right)}\\
H_1: & &z  \sim P_{ Z| \vect{\Theta} = \vect{\theta}},
\end{array}
\right.
\end{IEEEeqnarray}
 and leads to conclusions similar to those in Section~\ref{SecDecember26at10h51in2025HomeNice}.

\subsection{Connections to Euclidean Geometry} 

The following lemma presents an alternative expression for the generalization error~$\overline{\overline{\mathsf{G}}}(P_{\vect{\Theta} | \vect{Z}}, P_{\vect{Z}} )$ in~\eqref{EqJun4at9h02in2024}, in terms of the variation of the  functional~$\mathsf{R}_{\vect{\theta}}~$ in~\eqref{EqRModel} due to changes in the parameter~$\vect{\theta} \in \set{M}$.

\begin{lemma}\label{LemmaGEDataDrivenDef}
Consider the generalization error~$\overline{\overline{\mathsf{G}}}(P_{\vect{\Theta} | \vect{Z}}, P_{\vect{Z}} )$ in~\eqref{EqJun4at9h02in2024} and assume that for all~$\vect{\theta} \in \set{M}$
\begin{itemize}
\item[$(a)$] The probability measures~$P_{\vect{\Theta}}$ in~\eqref{EqJune26at16h54in2024} and the probability measure~$P_{\vect{Z}|\vect{\Theta}=\vect{\theta}}$ satisfy~\eqref{EqSaturdayJan27in2024b}; 
\item[$(b)$] for all~$\vect{\theta} \in \set{M}$, the probability measures~$P_{\vect{Z} | \vect{\Theta} = \vect{\theta}}$ and~$P_{\vect{Z}}$ are mutually absolutely continuous; and
\item[$(c)$]  For all~$\vect{z} \in \left( \set{X} \times \set{Y} \right)^n$, the probability measures~$P_{\vect{\Theta} | \vect{Z} = \vect{z}}$ and~$P_{\vect{\Theta}}$ in~\eqref{EqJune26at16h54in2024} are mutually absolutely continuous. 
\end{itemize}
Then, 
\begin{IEEEeqnarray}{rCl}
   \nonumber
	&&\overline{\overline{\mathsf{G}}}(P_{\vect{\Theta} | \vect{Z}}, P_{\vect{Z}} ) \\
	\label{EqGEDefinition2}
	&=& \iint \bigg( \mathsf{R}_{\vect{\nu}}\left(P_{\vect{Z}|\vect{\Theta}=\vect{\theta}}\right) - \mathsf{R}_{\vect{\theta}}  \left(P_{\vect{Z}|\vect{\Theta}=\vect{\theta}}\right) \bigg)\mathrm{d}P_{\vect{\Theta}}\left(\vect{\nu}\right)\mathrm{d}P_{\vect{\Theta}}\left(\vect{\theta}\right),\supersqueezeequ  \spnum
\end{IEEEeqnarray}
where the functional~$\mathsf{R}_{\vect{\theta}}~$ is defined in~\eqref{EqRModel}.
\end{lemma}

\begin{IEEEproof}
The proof follows from observing two series of equalities. The first  equalities are:
  \begin{IEEEeqnarray}{rCl}
\nonumber
&&\iint  \mathsf{R}_{\vect{\nu}}\left(P_{\vect{Z}|\vect{\Theta}=\vect{\theta}}\right)\mathrm{d}P_{\vect{\Theta}}\left(\vect{\nu}\right)\mathrm{d}P_{\vect{\Theta}}\left(\vect{\theta}\right)\\
  \label{EqNovember3at11h38in2024Niceb}
  &=&\iint \left( \int \mathsf{L} (\vect{z},\vect{\nu}) \mathrm{d} P_{\vect{Z}|\vect{\Theta}=\vect{\theta}}(\vect{z})\right) \mathrm{d}P_{\vect{\Theta}}\left(\vect{\nu}\right)\mathrm{d}P_{\vect{\Theta}}\left(\vect{\theta}\right) \spnum\\
    \label{EqNovember3at11h38in2024Nicec}
&=& \iint \left( \int \mathsf{L} (\vect{z},\vect{\nu}) \mathrm{d}P_{\vect{\Theta}}\left(\vect{\nu}\right) \right) \mathrm{d}P_{\vect{Z}|\vect{\Theta}=\vect{\theta}}(\vect{z})\mathrm{d}P_{\vect{\Theta}}\left(\vect{\theta}\right)\\
  \label{EqNovember3at11h38in2024Niced}
&=& \iint \mathsf{R}_{\vect{z}}\left(P_{\vect{\Theta}}\right) \mathrm{d}P_{\vect{Z}|\vect{\Theta}=\vect{\theta}}(\vect{z})\mathrm{d}P_{\vect{\Theta}}\left(\vect{\theta}\right)\\
  \label{EqNovember3at11h38in2024Nicee}
&=&  \iint \mathsf{R}_{\vect{z}}\left(P_{\vect{\Theta}}\right) \frac{\mathrm{d}P_{\vect{Z}|\vect{\Theta}=\vect{\theta}}}{\mathrm{d}P_{\vect{Z}}}(\vect{z})\mathrm{d}P_{\vect{Z}}(\vect{z})\mathrm{d}P_{\vect{\Theta}}\left(\vect{\theta}\right)\\
  \label{EqNovember3at11h38in2024Niceg}
&=& \int  \mathsf{R}_{\vect{z}}\left(P_{\vect{\Theta}}\right)\left( \int \frac{\mathrm{d}P_{\vect{Z}|\vect{\Theta}=\vect{\theta}}}{\mathrm{d}P_{\vect{Z}}}(\vect{z})\mathrm{d}P_{\vect{\Theta}}\left(\vect{\theta}\right)\right)\mathrm{d}P_{\vect{Z}}(\vect{z})\\
  \label{EqNovember3at11h38in2024Niceh}
&=& \int  \mathsf{R}_{\vect{z}}\left(P_{\vect{\Theta}}\right) \mathrm{d}P_{\vect{Z}}(\vect{z}),
  \end{IEEEeqnarray}
where the equality in~\eqref{EqNovember3at11h38in2024Niceb} follows from~\eqref{EqRModel}; 
the equality in~\eqref{EqNovember3at11h38in2024Nicec} follows by exchanging the order of the integrals \cite[Theorem~$2.6.6$]{ash2000probability}; 
the equality in~\eqref{EqNovember3at11h38in2024Niced} follows from~\eqref{EqRxy}; 
the equality in~\eqref{EqNovember3at11h38in2024Nicee} follows from~Assumption~$(b)$ and~\cite[Theorem~2]{InriaRR9591};
the equality in~\eqref{EqNovember3at11h38in2024Niceg} follows from \cite[Theorem~$2.6.3$]{ash2000probability}; and
the equality in~\eqref{EqNovember3at11h38in2024Niceh} follows from~\cite[Theorem~10]{InriaRR9591}.
 The second equalities are:
\begin{IEEEeqnarray}{ll}
\nonumber
&	\iint  \mathsf{R}_{\vect{\theta}}  \left(P_{\vect{Z}|\vect{\Theta}=\vect{\theta}}\right) \mathrm{d}P_{\vect{\Theta}}\left(\vect{\nu}\right)\mathrm{d}P_{\vect{\Theta}}\left(\vect{\theta}\right)\\
\label{EqNovember3at19h03in2024NiceA}
=& \int \mathsf{R}_{\vect{\theta}}  \left(P_{\vect{Z}|\vect{\Theta}=\vect{\theta}}\right)\mathrm{d}P_{\vect{\Theta}}\left(\vect{\theta}\right)\\
\label{EqNovember3at19h03in2024NiceB}
=& \iint \mathsf{L}\left(\vect{z},\vect{\theta}\right) \mathrm{d}P_{\vect{Z}|\vect{\Theta}=\vect{\theta}}\left(\vect{z}\right)\mathrm{d}P_{\vect{\Theta}}\left(\vect{\theta}\right)\\
\label{EqNovember3at19h03in2024NiceBa}
=& \iint \mathsf{L}\left(\vect{z},\vect{\theta}\right) \frac{\mathrm{d}P_{\vect{Z}|\vect{\Theta}=\vect{\theta}}}{\mathrm{d}P_{\vect{Z}}}\left(\vect{z}\right) \mathrm{d}P_{\vect{Z}}\left(\vect{z}\right) \mathrm{d}P_{\vect{\Theta}}\left(\vect{\theta}\right)\\
\label{EqNovember3at19h03in2024NiceBb}
=& \iint \mathsf{L}\left(\vect{z},\vect{\theta}\right) \frac{\mathrm{d}P_{\vect{Z}|\vect{\Theta}=\vect{\theta}}}{\mathrm{d}P_{\vect{Z}}}\left(\vect{z}\right)\mathrm{d}P_{\vect{\Theta}}\left(\vect{\theta}\right) \mathrm{d}P_{\vect{Z}}\left(\vect{z}\right) \\
\nonumber
=& \iint \mathsf{L}\left(\vect{z},\vect{\theta}\right) \frac{\mathrm{d}P_{\vect{Z}|\vect{\Theta}=\vect{\theta}}}{\mathrm{d}P_{\vect{Z}}}\left(\vect{z}\right) \frac{\mathrm{d}P_{\vect{\Theta}}}{\mathrm{d}P_{\vect{\Theta}|\vect{Z} = \vect{z}}}\left(\vect{\theta}\right) \mathrm{d}P_{\vect{\Theta}|\vect{Z} = \vect{z}}\left(\vect{\theta}\right) \mathrm{d}P_{\vect{Z}}\left(\vect{z}\right) \middlesqueezeequ  \\
\label{EqNovember3at19h03in2024NiceBc}
\\
	\label{EqNovember3at19h03in2024NiceC}	
=& \iint \mathsf{L}\left(\vect{z},\vect{\theta}\right) \mathrm{d}P_{\vect{\Theta}|\vect{Z}=\vect{z}}\left(\vect{\theta}\right)\mathrm{d}P_{\vect{Z}}\left(\vect{z}\right)\\
\label{EqNovember3at19h03in2024NiceD}
=& \int \mathsf{R}_{\vect{z}}\left(P_{\vect{\Theta}|\vect{Z}=\vect{z}}\right) \mathrm{d}P_{\vect{Z}}(\vect{z}),
\end{IEEEeqnarray}
where the equality in~\eqref{EqNovember3at19h03in2024NiceB} follows from~\eqref{EqRModel};
the equality in~\eqref{EqNovember3at19h03in2024NiceBa} follows from Assumption~$(b)$ and~\cite[Theorem~2]{InriaRR9591};
the equality in~\eqref{EqNovember3at19h03in2024NiceBb} follows from~\cite[Theorem~$2.6.6$]{ash2000probability};
the equality in~\eqref{EqNovember3at19h03in2024NiceBc} follows from Assumption~$(c)$ and~\cite[Theorem~2]{InriaRR9591};
the equality in~\eqref{EqNovember3at19h03in2024NiceC} follows from  Assumptions~$(b)$ and~$(c)$, \cite[Theorem~5]{InriaRR9591}; and~\cite[Theorem~11]{InriaRR9591};
and the equality in~\eqref{EqNovember3at19h03in2024NiceD} follows from~\eqref{EqRxy}.

From~\eqref{EqNovember3at11h38in2024Niceh} and~\eqref{EqNovember3at19h03in2024NiceD}, it follows that
\begin{IEEEeqnarray}{rCl}
&&\iint\bigg( \mathsf{R}_{\vect{\nu}}\left(P_{\vect{Z}|\vect{\Theta}=\vect{\theta}}\right) - \mathsf{R}_{\vect{\theta}}  \left(P_{\vect{Z}|\vect{\Theta}=\vect{\theta}}\right) \bigg) \mathrm{d}P_{\vect{\Theta}}\left(\vect{\nu}\right)\mathrm{d}P_{\vect{\Theta}}\left(\vect{\theta}\right)\supersqueezeequ  \spnum\\
&=&\int  \mathsf{R}_{\vect{z}}\left(P_{\vect{\Theta}}\right) \mathrm{d}P_{\vect{Z}}(\vect{z}) -  \int \mathsf{R}_{\vect{z}}\left(P_{\vect{\Theta}|\vect{Z}=\vect{z}}\right) \mathrm{d}P_{\vect{Z}}(\vect{z})\\
\label{EqNovember3at19h49in2024Nice}
&=& \int \mathsf{G}\left(\vect{z},P_{\vect{\Theta}},P_{\vect{\Theta}|\vect{Z}=\vect{z}} \right)\mathrm{d}P_{\vect{Z}}\left(\vect{z}\right)\\
\label{EqNovember3at19h49in2024Niceb}
& = & \overline{\overline{\mathsf{G}}} \left(P_{\vect{\Theta} | \vect{Z}}, P_{\vect{Z}} \right), 
\end{IEEEeqnarray}
where the equality in~\eqref{EqNovember3at19h49in2024Nice} follows from~\eqref{EqGZeta}; and the equality in~\eqref{EqNovember3at19h49in2024Niceb} follows from Lemma~\ref{LemmaJun4at9h32in2024}. 
This completes the proof.
\end{IEEEproof}

Lemma~\ref{LemmaGEDataDrivenDef} introduces an alternative expression for  the generalization error~$\overline{\overline{\mathsf{G}}}(P_{\vect{\Theta} | \vect{Z}}, P_{\vect{Z}} )$ that is reminiscent of the expression in~\eqref{EqJun4at9h02in2024}.
More specifically, Lemma~\ref{LemmaGEDataDrivenDef} presents~$\overline{\overline{\mathsf{G}}}(P_{\vect{\Theta} | \vect{Z}}, P_{\vect{Z}})$ as the expectation of the variation~$$ \mathsf{R}_{\vect{\nu}}\left(P_{\vect{Z}|\vect{\Theta}=\vect{\theta}}\right) - \mathsf{R}_{\vect{\theta}}  \left(P_{\vect{Z}|\vect{\Theta}=\vect{\theta}}\right),$$ under the assumption that models~$\vect{\nu}$ and~$\vect{\theta}$ are independently sampled from the same probability measure~$P_{\vect{\Theta}} \in \triangle\left( \set{M} \right)$ in~\eqref{EqJune26at16h54in2024}.
The following theorem leverages this observation to obtain an equivalent expression for the generalization error~$\overline{\overline{\mathsf{G}}}(P_{\vect{\Theta} | \vect{Z}}, P_{\vect{Z}} )$ that is of a similar form to the one introduced by  Theorem~\ref{TheoremJuin26at11h03in2024}. 

\begin{theorem}\label{TheoremTrainEntropyGE}
Consider the generalization error~$\overline{\overline{\mathsf{G}}}(P_{\vect{\Theta} | \vect{Z}}, P_{\vect{Z}} )$ in~\eqref{EqJun4at9h02in2024} and assume that:
\begin{itemize}
\item[$(a)$] The probability measure~$P_{\vect{Z}}$ satisfies~\eqref{EqSaturdayMai20at13h20in2024} for some~$P_{Z} \in \triangle\left( \set{X} \times \set{Y}\right)$; 
\item[$(b)$] The probability measures~$P_{\vect{\Theta}}$ in~\eqref{EqJune26at16h54in2024} and the probability measure~$P_{\vect{Z}|\vect{\Theta}=\vect{\theta}}$ satisfy~\eqref{EqSaturdayJan27in2024b}; 
\item[$(c)$] For all~$\vect{\theta} \in \set{M}$, the probability measures~$P_{\vect{Z} | \vect{\Theta} = \vect{\theta}}$ and~$P_{\vect{Z}}$ are mutually absolutely continuous; 
\item[$(d)$]  For all~$\vect{z} \in \left( \set{X} \times \set{Y} \right)^n$, the probability measures~$P_{\vect{\Theta} | \vect{Z} = \vect{z}}$ and~$P_{\vect{\Theta}}$ in~\eqref{EqJune26at16h54in2024} are mutually absolutely continuous; and
\item[$(e)$] For all~$\vect{\theta} \in \set{M}$,  the probability measures~$P_{Z | \vect{\Theta} = \vect{\theta}}$ and~$P_{Z}$  satisfy~\eqref{EqOctober17at14h40in2024} and are both mutually absolutely continuous with the probability measure~$P_S \in \triangle\left( \set{X} \times \set{Y} \right)$ in~\eqref{Eqtheworstgeneral}. 
\end{itemize}
	Then,
\begin{IEEEeqnarray}{rcl}
	\nonumber
\overline{\overline{\mathsf{G}}} \left(P_{\vect{\Theta} | \vect{Z}}, P_{\vect{Z}} \right) \Dsupersqueezeequ &=& \beta \iint \bigg(\KL{P_{Z|\vect{\Theta}=\vect{\mu}}}{P^{(P_{S},\beta)}_{\hat{Z}|\vect{\Theta}=\vect{\mu}}} \Dsupersqueezeequ\IEEEeqnarraynumspace\\
&&- \KL{P_{Z|\vect{\Theta}=\vect{\mu}}}{P^{(P_{S},\beta)}_{\hat{Z}|\vect{\Theta}=\vect{\nu}} }\bigg)\mathrm{d}P_{\vect{\Theta}}(\vect{\nu})\mathrm{d}P_{\vect{\Theta}}(\vect{\mu}), \Dsupersqueezeequ\IEEEeqnarraynumspace
\end{IEEEeqnarray}
where the measure~$P_{\hat{Z} | \vect{\Theta} = \vect{\mu}}^{\left( P_{S}, \beta \right)}$ is  the WCDG probability measure in~\eqref{Eqtheworstgeneral}; and~$\beta \in \set{J}_{P_S}$, with~$\set{J}_{P_S}$ in~\eqref{EqOctober17at21h21in2024}.
\end{theorem}
\begin{IEEEproof}
The proof follows from Lemma~\ref{LemmaGEDataDrivenDef}, which holds under Assumptions $(b)$, $(c)$, and $(d)$.
More specifically, the proof consists of re-writing the difference~$\mathsf{R}_{\vect{\nu}}\left(P_{\vect{Z}|\vect{\Theta}=\vect{\mu}}\right) - \mathsf{R}_{\vect{\mu}}\left(P_{\vect{Z}|\vect{\Theta}=\vect{\mu}}\right)$ in~\eqref{EqGEDefinition2} using the functional~$\mathsf{G}$ in~\eqref{EqGTheta}.
Under Assumption~$(a)$ and Lemma~\ref{LemmaNovember5at17h06in2024SophiaAntipolis}, note that for all $\left( \vect{\nu}, \vect{\mu} \right) \in \set{M} \times \set{M}$, 
	\begin{IEEEeqnarray}{rCl}
\mathsf{R}_{\vect{\nu}}\left(P_{\vect{Z}|\vect{\Theta}=\vect{\mu}}\right)
	&=& \int \ell\left(x,y,\vect{\nu}\right)\mathrm{d}P_{Z|\vect{\Theta}=\vect{\mu}}(x,y),
	\end{IEEEeqnarray}
where the conditional probability measure $P_{Z | \vect{\Theta}}$ satisfies~\eqref{EqOctober17at14h40in2024}.
Hence, the following holds for all~$\left( \vect{\mu} , \vect{\nu} \right) \in \set{M}\times \set{M}$,
	\begin{IEEEeqnarray}{rCl}
		\nonumber
	&&\mathsf{R}_{\vect{\nu}}\left(P_{\vect{Z}|\vect{\Theta}=\vect{\mu}}\right)\\
\label{EqNovember3at20h51in2024Nice}
	&=& 	\mathsf{G}\left(\vect{\nu},P_{Z|\vect{\Theta}=\vect{\mu}},P^{(P_{S},\beta)}_{\hat{Z}|\vect{\Theta}=\vect{\nu}} \right)  + \int \ell\left(x,y,\vect{\nu}\right)\mathrm{d} P_{\hat{Z} | \vect{\Theta} = \vect{\nu}}^{\left( P_{S}, \beta \right)} (x,y),  \supersqueezeequ \spnum
	\end{IEEEeqnarray}	
	and 
	\begin{IEEEeqnarray}{rCl}
		\nonumber
	&&\mathsf{R}_{\vect{\mu}} \left(P_{\vect{Z}|\vect{\Theta}=\vect{\mu}}\right)\\
	\label{EqNovember3at21h00in2024Nice}
	&=& \mathsf{G}\left(\vect{\mu},P_{Z|\vect{\Theta}=\vect{\mu}},P^{(P_{S},\beta)}_{\hat{Z}|\vect{\Theta}=\vect{\mu}} \right)  + \int \ell\left(x,y,\vect{\mu}\right)\mathrm{d} P_{\hat{Z} | \vect{\Theta} = \vect{\mu}}^{\left( P_{S}, \beta \right)} (x,y),  \supersqueezeequ \spnum
	\end{IEEEeqnarray}
where the functional~$\mathsf{G}$ is defined in~\eqref{EqGTheta}.	
From Assumption~$(e)$ and Lemma~\ref{LemmaDeviatefromWorst} (in Section~\ref{SecDataDrivenGaps}), it holds that
\begin{IEEEeqnarray}{rCl}
	\nonumber
&&	\mathsf{G}\left(\vect{\nu},P_{Z|\vect{\Theta}=\vect{\mu}},P^{(P_{S},\beta)}_{\hat{Z}|\vect{\Theta}=\vect{\nu}} \right)  \\
\nonumber
	&=& \beta \bigg( \KL{P_{Z|\vect{\Theta}=\vect{\mu}}}{P_S} - \KL{P_{Z|\vect{\Theta}=\vect{\mu}}}{P^{(P_{S},\beta)}_{\hat{Z}|\vect{\Theta}=\vect{\nu}} } \\
	\label{EqNovember3at21h06in2024Nice}
	&&- \KL{P^{(P_{S},\beta)}_{\hat{Z}|\vect{\Theta}=\vect{\nu}} }{P_S}\bigg),
\end{IEEEeqnarray}
and
\begin{IEEEeqnarray}{rCl}
	\nonumber
		&&\mathsf{G}\left(\vect{\mu},P_{Z|\vect{\Theta}=\vect{\mu}},P^{(P_{S},\beta)}_{\hat{Z}|\vect{\Theta}=\vect{\mu}} \right) \\
\nonumber
		&=& \beta \bigg( \KL{P_{Z|\vect{\Theta}=\vect{\mu}}}{P_S} - \KL{P_{Z|\vect{\Theta}=\vect{\mu}}}{P^{(P_{S},\beta)}_{\hat{Z}|\vect{\Theta}=\vect{\mu}}} \\
	\label{EqNovember3at21h07in2024Nice}
	&&- \KL{P^{(P_{S},\beta)}_{\hat{Z}|\vect{\Theta}=\vect{\mu}}}{P_S}\bigg).
	\end{IEEEeqnarray}

Using these equalities, it follows from~\eqref{EqNovember3at20h51in2024Nice} and~\eqref{EqNovember3at21h00in2024Nice}	that
\begin{IEEEeqnarray}{rCl}
	\nonumber
&&	\mathsf{R}_{\vect{\nu}}\left(P_{\vect{Z}|\vect{\Theta}=\vect{\mu}}\right) - \mathsf{R}_{\vect{\mu}} \left(P_{\vect{Z}|\vect{\Theta}=\vect{\mu}}\right)\\
\nonumber
&=& 	\mathsf{G}\left(\vect{\nu},P_{Z|\vect{\Theta}=\vect{\mu}},P^{(P_{S},\beta)}_{\hat{Z}|\vect{\Theta}=\vect{\nu}} \right)  + \int \ell\left(x,y,\vect{\nu}\right)\mathrm{d} P_{\hat{Z} | \vect{\Theta} = \vect{\nu}}^{\left( P_{S}, \beta \right)} (x,y) \supersqueezeequ \spnum\\
\label{EqNovember3at21h05in2024Nice}		
& & - \mathsf{G}\left(\vect{\mu},P_{Z|\vect{\Theta}=\vect{\mu}},P^{(P_{S},\beta)}_{\hat{Z}|\vect{\Theta}=\vect{\mu}} \right) - \int \ell\left(x,y,\vect{\mu}\right)\mathrm{d} P_{\hat{Z} | \vect{\Theta} = \vect{\mu}}^{\left( P_{S}, \beta \right)} (x,y) \supersqueezeequ \spnum\\
\nonumber
	&=& \beta \bigg( - \KL{P_{Z|\vect{\Theta}=\vect{\mu}}}{P^{(P_{S},\beta)}_{\hat{Z}|\vect{\Theta}=\vect{\nu}} } - \KL{P^{(P_{S},\beta)}_{\hat{Z}|\vect{\Theta}=\vect{\nu}} }{P_S}\bigg)\\
\nonumber
	&& -\beta \bigg(-\KL{P_{Z|\vect{\Theta}=\vect{\mu}}}{P^{(P_{S},\beta)}_{\hat{Z}|\vect{\Theta}=\vect{\mu}}} -\KL{P^{(P_{S},\beta)}_{\hat{Z}|\vect{\Theta}=\vect{\mu}}}{P_S}\bigg)\supersqueezeequ  \spnum \\
\nonumber
	&&+ \int \ell\left(x,y,\vect{\nu}\right)\mathrm{d} P_{\hat{Z} | \vect{\Theta} = \vect{\nu}}^{\left( P_{S}, \beta \right)} (x,y) \\
\label{EqNovember17at13h09in2024Nice}
	&&- \int \ell\left(x,y,\vect{\mu}\right)\mathrm{d} P_{\hat{Z} | \vect{\Theta} = \vect{\mu}}^{\left( P_{S}, \beta \right)} (x,y) \Dsupersqueezeequ  \spnum\\
	\nonumber
	&=& \beta \KL{P_{Z|\vect{\Theta}=\vect{\mu}}}{P^{(P_{S},\beta)}_{\hat{Z}|\vect{\Theta}=\vect{\mu}}} - \beta  \KL{P_{Z|\vect{\Theta}=\vect{\mu}}}{P^{(P_{S},\beta)}_{\hat{Z}|\vect{\Theta}=\vect{\nu}} } \Dsupersqueezeequ  \spnum\\
\nonumber
	&&+ \int \ell\left(x,y,\vect{\nu}\right)\mathrm{d} P_{\hat{Z} | \vect{\Theta} = \vect{\nu}}^{\left( P_{S}, \beta \right)} (x,y) - \beta \KL{P^{(P_{S},\beta)}_{\hat{Z}|\vect{\Theta}=\vect{\nu}} }{P_S}\\
\label{EqNovember3at21h05in2024NiceB}		
	&&- \int \ell\left(x,y,\vect{\mu}\right)\mathrm{d} P_{\hat{Z} | \vect{\Theta} = \vect{\mu}}^{\left( P_{S}, \beta \right)} (x,y) + \beta \KL{P^{(P_{S},\beta)}_{\hat{Z}|\vect{\Theta}=\vect{\mu}}}{P_S}
\end{IEEEeqnarray}
where the equality in~\eqref{EqNovember17at13h09in2024Nice} follows from both~\eqref{EqNovember3at21h06in2024Nice} and~\eqref{EqNovember3at21h07in2024Nice}.

Therefore, from Lemma~\ref{LemmaGEDataDrivenDef}, it follows that
\begin{IEEEeqnarray}{rCl}
	\nonumber
	&&\overline{\overline{\mathsf{G}}} \left(P_{\vect{\Theta} | \vect{Z}}, P_{\vect{Z}} \right)\\
	&=&\iint 	\mathsf{R}_{\vect{\nu}}\left(P_{\vect{Z}|\vect{\Theta}=\vect{\mu}}\right) - \mathsf{R}_{\vect{\mu}} \left(P_{\vect{Z}|\vect{\Theta}=\vect{\mu}}\right)\mathrm{d}P_{\vect{\Theta}}(\vect{\nu})\mathrm{d}P_{\vect{\Theta}}(\vect{\mu})\supersqueezeequ  \spnum \\
	\nonumber
	&=& \beta \iint \bigg(\KL{P_{Z|\vect{\Theta}=\vect{\mu}}}{P^{(P_{S},\beta)}_{\hat{Z}|\vect{\Theta}=\vect{\mu}}} \\
	&&- \KL{P_{Z|\vect{\Theta}=\vect{\mu}}}{P^{(P_{S},\beta)}_{\hat{Z}|\vect{\Theta}=\vect{\nu}} }\bigg)\mathrm{d}P_{\vect{\Theta}}(\vect{\nu})\mathrm{d}P_{\vect{\Theta}}(\vect{\mu}),
\end{IEEEeqnarray}
which completes the proof.
\end{IEEEproof}

Theorem~\ref{TheoremTrainEntropyGE} expresses the generalization error as the difference between the expectations of two relative entropies. The first relative entropy compares the likelihood~$P_{Z|\vect{\Theta}=\vect{\theta}}$ to the WCDG probability measure~$P^{(P_{S},\beta)}_{\hat{Z}|\vect{\Theta}=\vect{\theta}}$, under the observation of the same model~$\vect{\theta}$, which is drawn from~$P_{\vect{\Theta}}$ in~\eqref{EqJune26at16h54in2024}. The second relative entropy compares these two conditional probability measures under the case that the observed models are drawn  independently from the same probability measure~$P_{\vect{\Theta}}$. More interestingly, the second relative entropy satisfies the following property.

\begin{lemma}\label{LemmaGeometricEntropy}
Consider a probability measure~$P_{\vect{Z}}\in \triangle\left( \left( \set{X} \times \set{Y}\right)^n\right)$ and  a conditional probability measure~$P_{\vect{\Theta}|\vect{Z}} \in \triangle\left( \set{M} | \left( \set{X} \times \set{Y}\right)^n\right)$ and assume that:
\begin{itemize}
\item[$(a)$] The probability measure~$P_{\vect{Z}}$ satisfies~\eqref{EqSaturdayMai20at13h20in2024} for some~$P_{Z} \in \triangle\left( \set{X} \times \set{Y}\right)$; 
\item[$(b)$] The probability measures~$P_{\vect{\Theta}}$ in~\eqref{EqJune26at16h54in2024} and the probability measure~$P_{\vect{Z}|\vect{\Theta}=\vect{\theta}}$ satisfy~\eqref{EqSaturdayJan27in2024b}; 
\item[$(c)$] For all~$\vect{\theta} \in \set{M}$, the probability measure~$P_{\vect{Z} | \vect{\Theta} = \vect{\theta}}$ is absolutely continuous with respect to~$P_{\vect{Z}}$; and
\item[$(d)$] For all~$\vect{\theta} \in \set{M}$,  the probability measures~$P_{Z | \vect{\Theta} = \vect{\theta}}$ and~$P_{Z}$  satisfy~\eqref{EqOctober17at14h40in2024} and are both absolutely continuous with respect to the probability measure~$P_S \in \triangle\left( \set{X} \times \set{Y} \right)$ in~\eqref{Eqtheworstgeneral}. 
\end{itemize}
Then,
	\begin{IEEEeqnarray}{rCl}
		\nonumber
&&	\iint \KL{P_{Z|\vect{\Theta}=\vect{\theta}}}{P^{(P_{S},\beta)}_{\hat{Z}|\vect{\Theta}=\vect{\nu}} }\bigg)\mathrm{d}P_{\vect{\Theta}}(\vect{\nu})\mathrm{d}P_{\vect{\Theta}}(\vect{\theta})\\
\nonumber
	&=&\int \bigg(\KL{P_{Z|\vect{\Theta}=\vect{\theta}}}{P_Z}  + \KL{P_Z}{P^{(P_{S},\beta)}_{\hat{Z}|\vect{\Theta}=\vect{\theta}}} \bigg)\mathrm{d}P_{\vect{\Theta}}\left(\vect{\theta}\right),
	\end{IEEEeqnarray}
where   the probability measure~$P_{\hat{Z} | \vect{\Theta} = \vect{\theta}}^{\left( P_{S}, \beta \right)}~$ is defined in~\eqref{Eqtheworstgeneral};~$\beta \in \set{J}_{P_S}$, with~$\set{J}_{P_S}$ in~\eqref{EqOctober17at21h21in2024}. 
\end{lemma}
\begin{IEEEproof}
The proof follows by noticing that
\begin{IEEEeqnarray}{rCl}
\nonumber
&&	\iint \KL{P_{Z|\vect{\Theta}=\vect{\theta}}}{P^{(P_{S},\beta)}_{\hat{Z}|\vect{\Theta}=\vect{\nu}} }\bigg)\mathrm{d}P_{\vect{\Theta}}(\vect{\nu})\mathrm{d}P_{\vect{\Theta}}(\vect{\theta}) \supersqueezeequ \\
\label{November4at15h33in2024Nice}
		&=& \int \iint  \log{\frac{\mathrm{d}P_{Z|\vect{\Theta}=\vect{\theta}}}{\mathrm{d}P^{(P_{S},\beta)}_{\hat{Z}|\vect{\Theta}=\vect{\nu}} } (z)}\mathrm{d}P_{Z|\vect{\Theta}=\vect{\theta}}(z) \mathrm{d}P_{\vect{\Theta}}(\vect{\nu})\mathrm{d}P_{\vect{\Theta}}(\vect{\theta})\supersqueezeequ\\
\nonumber
		&=& \iint \int \log{\frac{\mathrm{d}P_{Z|\vect{\Theta}=\vect{\theta}}}{\mathrm{d}P_Z}(z)} \mathrm{d}P_{Z|\vect{\Theta}=\vect{\theta}}(z) \mathrm{d}P_{\vect{\Theta}}(\vect{\theta}) \\
\label{November4at15h33in2024Nicea}
		&+& \iint \int \log{\frac{\mathrm{d}P_Z}{\mathrm{d}P^{(P_{S},\beta)}_{\hat{Z}|\vect{\Theta}=\vect{\nu}} }(z)}\mathrm{d}P_{Z|\vect{\Theta}=\vect{\theta}}(z)\mathrm{d}P_{\vect{\Theta}}(\vect{\nu})\mathrm{d}P_{\vect{\Theta}}(\vect{\theta}),\Tsupersqueezeequ \spnum\\
\nonumber
		&=& \int \KL{P_{Z|\vect{\Theta}=\vect{\theta}}}{P_Z}\mathrm{d}P_{\vect{\Theta}}(\vect{\theta}) \\
\label{November4at15h33in2024Niceb}
		&+& \iint \int \log{\frac{\mathrm{d}P_Z}{\mathrm{d}P^{(P_{S},\beta)}_{\hat{Z}|\vect{\Theta}=\vect{\nu}} }(z)}\mathrm{d}P_{Z|\vect{\Theta}=\vect{\theta}}(z)\mathrm{d}P_{\vect{\Theta}}(\vect{\nu})\mathrm{d}P_{\vect{\Theta}}(\vect{\theta}),\Tsupersqueezeequ \spnum
\end{IEEEeqnarray}		
where the equality in~\eqref{November4at15h33in2024Nicea} follows from Assumptions~$(a)$ and~$(c)$ together with  Lemma~\ref{EqNovember15at10h08in2024SophiaAntipolis} (in Appendix~\ref{AppendixMiscellanea}) and \cite[Theorem~$2$]{InriaRR9591}. 

The rest of the proof follows from~\eqref{November4at15h33in2024Niceb} by noticing that
\begin{IEEEeqnarray}{rcl}		
\nonumber
&& \iint \int \log{\frac{\mathrm{d}P_Z}{\mathrm{d}P^{(P_{S},\beta)}_{\hat{Z}|\vect{\Theta}=\vect{\nu}} }(z)}\mathrm{d}P_{Z|\vect{\Theta}=\vect{\theta}}(z)\mathrm{d}P_{\vect{\Theta}}(\vect{\nu})\mathrm{d}P_{\vect{\Theta}}(\vect{\theta})\\
\nonumber
		&=& \iint \int \log \left( \frac{\mathrm{d}P_Z}{\mathrm{d}P^{(P_{S},\beta)}_{\hat{Z}|\vect{\Theta}=\vect{\nu}} }(z)\right) \\
\label{EqNovember4at16h06in2024Nice}
		& & \frac{\mathrm{d}P_{Z|\vect{\Theta}=\vect{\theta}}}{\mathrm{d} P_{Z}}(z) \mathrm{d}P_{Z}(z)\mathrm{d}P_{\vect{\Theta}}(\vect{\nu})\mathrm{d}P_{\vect{\Theta}}(\vect{\theta}) \spnum\\
\nonumber
&=&  \iint \log \left( \frac{\mathrm{d}P_Z}{\mathrm{d}P^{(P_{S},\beta)}_{\hat{Z}|\vect{\Theta}=\vect{\nu}} }(z)\right) \\
\label{EqNovember4at16h06in2024Niceb}
& & \left( \int \frac{\mathrm{d}P_{Z|\vect{\Theta}=\vect{\theta}}}{\mathrm{d} P_{Z}}(z) \mathrm{d}P_{\vect{\Theta}}(\vect{\theta}) \right) \mathrm{d}P_{Z}(z)\mathrm{d}P_{\vect{\Theta}}(\vect{\nu})  \spnum\\
\label{EqNovember4at16h06in2024Nicec}
&=&  \iint \log \left( \frac{\mathrm{d}P_Z}{\mathrm{d}P^{(P_{S},\beta)}_{\hat{Z}|\vect{\Theta}=\vect{\nu}} }(z)\right) 
\mathrm{d}P_{Z}(z)\mathrm{d}P_{\vect{\Theta}}(\vect{\nu}) \\
\label{EqNovember4at16h06in2024Niced}
		&=& \int \KL{P_Z}{P^{(P_{S},\beta)}_{\hat{Z}|\vect{\Theta}=\vect{\theta}}} \bigg)\mathrm{d}P_{\vect{\Theta}}\left(\vect{\theta}\right),
\end{IEEEeqnarray}
where the equality in~\eqref{EqNovember4at16h06in2024Nice} follows from  Assumptions~$(a)$ and~$(c)$ together with  Lemma~\ref{EqNovember15at10h08in2024SophiaAntipolis} (in Appendix~\ref{AppendixMiscellanea}) and~\cite[Theorem~$2$]{InriaRR9591}; 
the equality in~\eqref{EqNovember4at16h06in2024Niceb} follows by exchanging the order of the integrals \cite[Theorem~$2.6.6$]{ash2000probability}; 
the equality in~\eqref{EqNovember4at16h06in2024Nicec} follows from~\cite[Theorem~$10$]{InriaRR9591}.
The proof is completed by plugging~\eqref{EqNovember4at16h06in2024Niced} into~\eqref{November4at15h33in2024Niceb}.
\end{IEEEproof}

The converse of  the Pythagorean theorem~\cite[Book~I, Proposition~$48$]{heath1956thirteen} together with  Lemma~\ref{LemmaGeometricEntropy}, lead to the geometric construction shown in Figure~\ref{FigNovember4at16h20in2024Nice}. 
Note the duality between Figure~\ref{FigSeptember4at18h58in2024} and Figure~\ref{FigNovember4at16h20in2024Nice}.
From Theorem~\ref{TheoremTrainEntropyGE}, it follows that 
\begin{IEEEeqnarray}{rcl}
	\nonumber
& & \iint  \KL{P_{Z|\vect{\Theta}=\vect{\mu}}}{P^{(P_{S},\beta)}_{\hat{Z}|\vect{\Theta}=\vect{\nu}} }  \mathrm{d}P_{\vect{\Theta}}(\vect{\nu})\mathrm{d}P_{\vect{\Theta}}(\vect{\mu}) \\
& =  & \int\KL{P_{Z|\vect{\Theta}=\vect{\mu}}}{P^{(P_{S},\beta)}_{\hat{Z}|\vect{\Theta}=\vect{\mu}}}  \mathrm{d}P_{\vect{\Theta}}(\vect{\mu}) - \frac{1}{\beta}\overline{\overline{\mathsf{G}}} \left(P_{\vect{\Theta} | \vect{Z}}, P_{\vect{Z}} \right)  , \squeezeequ \IEEEeqnarraynumspace
\end{IEEEeqnarray}
which involves the generalization error~$\overline{\overline{\mathsf{G}}} \left(P_{\vect{\Theta} | \vect{Z}}, P_{\vect{Z}} \right)$ in~\eqref{EqJun4at9h02in2024}.
This observation leads to the construction in Figure~\ref{FigNovember4at19h58in2024Nice}, which exhibits a duality with Figure~\ref{FigOctober20at17h43in2024atAgadir}.

\begin{figure}
	\begin{center}
		\begin{tikzpicture}
			\draw[thick] (0,0) -- node[below] {$\Tsupersqueezeequ\sqrt{ \displaystyle	\int \KL{P_Z}{P^{(P_{S},\beta)}_{\hat{Z}|\vect{\Theta}=\vect{\nu}} } \mathrm{d}P_{\vect{\Theta}}\left(\vect{\theta}\right)}$} ++(6,0);
			\draw[thick] (0,0) -- node[left,rotate=90, above] {$\supersqueezeequ\sqrt{\displaystyle \int \KL{P_{Z|\vect{\Theta}=\vect{\theta}}}{P_Z}\mathrm{d}P_{\vect{\Theta}}(\vect{\theta})}$} ++(0,4);
			\draw[thick] (0,4) -- (6,0);
			\node[left, rotate=-34, above] at (3,2) {$\supersqueezeequ\sqrt{\displaystyle \iint \KL{P_{Z|\vect{\Theta}=\vect{\theta}}}{P^{(P_{S},\beta)}_{\hat{Z}|\vect{\Theta}=\vect{\nu}} }\bigg)\mathrm{d}P_{\vect{\Theta}}(\vect{\nu})\mathrm{d}P_{\vect{\Theta}}(\vect{\theta})}$};
		\end{tikzpicture}
	\end{center}
	\caption{Geometric interpretation of Lemma~\ref{LemmaGeometricEntropy} involving only relative entropies. Note that~$ \int \KL{P_{Z|\vect{\Theta}=\vect{\theta}}}{P_Z}\mathrm{d}P_{\vect{\Theta}}(\vect{\theta}) = I\left(P_{Z|\vect{\Theta}}; P_{\vect{\Theta}} \right)$.}
	\label{FigNovember4at16h20in2024Nice}
\end{figure} 

\begin{figure}
	\begin{center}
		\begin{tikzpicture}
			\draw[thick] (0,0) -- node[below] {$\Tsupersqueezeequ\sqrt{ \displaystyle	\int \KL{P_Z}{P^{(P_{S},\beta)}_{\hat{Z}|\vect{\Theta}=\vect{\theta}} } \mathrm{d}P_{\vect{\Theta}}\left(\vect{\theta}\right)}$} ++(6,0);
			\draw[thick] (0,0) -- node[left,rotate=90, above] {$\supersqueezeequ\sqrt{\displaystyle \int \KL{P_{Z|\vect{\Theta}=\vect{\theta}}}{P_Z}\mathrm{d}P_{\vect{\Theta}}(\vect{\theta})}$} ++(0,4);
			\draw[thick] (0,4) -- (6,0);
			\node[left, rotate=-34, above] at (3,2) {$\squeezeequ\sqrt{\displaystyle\int\KL{P_{Z|\vect{\Theta}=\vect{\theta}}}{P^{(P_{S},\beta)}_{\hat{Z}|\vect{\Theta}=\vect{\theta}}}  \mathrm{d}P_{\vect{\Theta}}(\vect{\theta}) - \frac{1}{\beta}\overline{\overline{\mathsf{G}}} \left(P_{\vect{\Theta} | \vect{Z}}, P_{\vect{Z}} \right)}$};
		\end{tikzpicture}
	\end{center}
	\caption{Geometric interpretation of Lemma~\ref{LemmaGeometricEntropy} involving the generalization error and relative entropies. Note that~$ \int \KL{P_{Z|\vect{\Theta}=\vect{\theta}}}{P_Z}\mathrm{d}P_{\vect{\Theta}}(\vect{\theta}) = I\left(P_{Z|\vect{\Theta}}; P_{\vect{\Theta}} \right)$.}
	\label{FigNovember4at19h58in2024Nice}
\end{figure} 
 
\section{Algorithm-Driven Gaps}\label{SecAlgoDrivenGaps}

The purpose of this section is to provide explicit expressions for the algorithm-driven gap~$\mathsf{G}\left( \vect{z}, P_1, P_2 \right)$ in~\eqref{EqGZeta} for an arbitrary dataset~$\vect{z}$ and two arbitrary probability measures~$P_1$ and~$P_2$ in~$\triangle\left( \set{M} \right)$.
%
%Such explicit expressions are provided later by Theorem~\ref{TheoremRGgeneral}. 
%Such probability measures might be instances of possibly different algorithms.
%
In particular, these expressions are in terms of the Gibbs probability measure~$P^{\left(Q, \lambda\right)}_{\vect{\Theta}| \vect{Z} = \vect{z}}$ in~\eqref{EqGenpdf}, which represents a Gibbs algorithm. 
For the ease of presentation, the analysis of~$\mathsf{G}\left( \vect{z}, P_1, P_2 \right)$ is performed using the fact that ~$\mathsf{G}\left( \vect{z}, P_1 , P_2 \right) = \mathsf{G}\left( \vect{z}, P_1 , P^{\left(Q, \lambda\right)}_{\vect{\Theta}| \vect{Z} = \vect{z}} \right) - \mathsf{G}\left( \vect{z}, P_2 , P^{\left(Q, \lambda\right)}_{\vect{\Theta}| \vect{Z} = \vect{z}} \right)$. Thus, the focus is on the gap~$\mathsf{G}\left( \vect{z}, P , P^{\left(Q, \lambda\right)}_{\vect{\Theta}| \vect{Z} = \vect{z}} \right)$, for an arbitrary measure~$P \in \triangle\left( \set{M} \right)$.
 
\subsection{Deviations from the Gibbs Measure}
The variation or \emph{sensitivity} of the functional~$\mathsf{R}_{\vect{z}}$ in~\eqref{EqRxy} to deviations from the probability measure~$P^{\left(Q, \lambda\right)}_{\vect{\Theta}| \vect{Z} = \vect{z}}$ in~\eqref{EqGenpdf} to an alternative probability measure~$P$ is  
\begin{IEEEeqnarray}{rCl}
\label{EqDefSensitivity}
\mathsf{G}\left( \vect{z}, P , P^{\left(Q, \lambda\right)}_{\vect{\Theta}| \vect{Z} = \vect{z}}\right)   & = & \mathsf{R}_{\vect{z}}\left( P \right)  - \mathsf{R}_{\vect{z}}\left( P^{\left(Q, \lambda\right)}_{\vect{\Theta}| \vect{Z} = \vect{z}} \right),
\end{IEEEeqnarray}
where the functional~$\mathsf{G}$ is defined in~\eqref{EqGZeta}.
%
%The value~$\mathsf{G}\left( \vect{z}, P , P^{\left(Q, \lambda\right)}_{\vect{\Theta}| \vect{Z} = \vect{z}}\right)$ in~\eqref{EqDefSensitivity} is the \emph{sensitivity} of the expected empirical risk for a fixed dataset to variations from the probability measure~$P^{\left(Q, \lambda\right)}_{\vect{\Theta}| \vect{Z} = \vect{z}}$ in~\eqref{EqGenpdf} to an arbitrary probability measure~$P \in \triangle\left( \set{X} \times \set{Y} \right)$.  
%%
The following theorem introduces an explicit expression for~$\mathsf{G}\left( \vect{z}, P , P^{\left(Q, \lambda\right)}_{\vect{\Theta}| \vect{Z} = \vect{z}}\right)$.
\begin{lemma}[Theorem~$1$ in \cite{Perlaza-ISIT2023b}]\label{LemmaSensitivityEqual}
The probability measure~$P^{\left(Q, \lambda\right)}_{\vect{\Theta}| \vect{Z} = \vect{z}}$ in~\eqref{EqGenpdf} satisfies for all~$P \in \triangle_{Q}\left( \set{M} \right)$,
\begin{IEEEeqnarray}{rcl}
\nonumber
& & \mathsf{G}\left( \vect{z}, P , P^{\left(Q, \lambda\right)}_{\vect{\Theta}| \vect{Z} = \vect{z}}\right)\\
\label{EqSensitivity}
& = &  \lambda\left( \KL{P^{\left(Q, \lambda\right)}_{\vect{\Theta}| \vect{Z} = \vect{z}}}{Q} + \KL{P}{P^{\left(Q, \lambda\right)}_{\vect{\Theta}| \vect{Z} = \vect{z}}} - \KL{P}{Q} \right),\IEEEeqnarraynumspace
\end{IEEEeqnarray}
where  the functional~$\mathsf{G}$ is defined in~\eqref{EqGZeta}.
\end{lemma}
  
 An interesting observation from Lemma~\ref{LemmaSensitivityEqual} is that when the reference measure~$Q$ is a probability measure, it holds that~$\KL{P}{Q}~\geqslant~0$ \cite[Theorem~$1$]{PerlazaTIT2024},  which leads to the following corollary of Lemma~\ref{LemmaSensitivityEqual}.
\begin{corollary}\label{CorDiffERMRERNoPasaNada}
If~$Q$ is a probability measure, then the gap~$\mathsf{G}\left( \vect{z}, P , P^{\left(Q, \lambda\right)}_{\vect{\Theta}| \vect{Z} = \vect{z}}\right)$ in~\eqref{EqDefSensitivity} satisfies 
\begin{IEEEeqnarray}{rcl}
\label{EqSensitivityLasNoches}
\mathsf{G}\left( \vect{z}, P , P^{\left(Q, \lambda\right)}_{\vect{\Theta}| \vect{Z} = \vect{z}}\right) \squeezeequ
& \leqslant &  \lambda\left( \KL{P^{\left(Q, \lambda\right)}_{\vect{\Theta}| \vect{Z} = \vect{z}}}{Q} + \KL{P}{P^{\left(Q, \lambda\right)}_{\vect{\Theta}| \vect{Z} = \vect{z}}} \right)\supersqueezeequ,\IEEEeqnarraynumspace
\end{IEEEeqnarray}
with equality if~$\KL{P}{Q} = 0$.
\end{corollary}

Also from Lemma~\ref{LemmaSensitivityEqual}, under the assumption that $Q$ is a probability measure, it follows that the gap~$\mathsf{G}\left( \vect{z}, P , P^{\left(Q, \lambda\right)}_{\vect{\Theta}| \vect{Z} = \vect{z}}\right)$ in~\eqref{EqDefSensitivity} is nonnegative for all measures $P$ within a neighborhood of $Q$, as suggested in the following corollary.  

\begin{corollary}\label{Cor4JuneAt6h58in2024}
For all $P \in \triangle_{Q}\left( \set{M} \right)$, such that $\KL{P}{Q} \leqslant \KL{P^{\left(Q, \lambda\right)}_{\vect{\Theta}| \vect{Z} = \vect{z}}}{Q}$, with $P^{\left(Q, \lambda\right)}_{\vect{\Theta}| \vect{Z} = \vect{z}}$ in~\eqref{EqGenpdf}, the gap~$\mathsf{G}\left( \vect{z}, P, P^{\left(Q, \lambda\right)}_{\vect{\Theta}| \vect{Z} = \vect{z}} \right)$ in~\eqref{EqDefSensitivity}  satisfies 
\begin{IEEEeqnarray}{rCl}
\label{Eq4JuneAt6h59in2024}
\mathsf{G}\left( \vect{z}, P , P^{\left(Q, \lambda\right)}_{\vect{\Theta}| \vect{Z} = \vect{z}}\right) \squeezeequ
& \geqslant & 0\IEEEeqnarraynumspace
\end{IEEEeqnarray}
with equality if~$\KL{P}{P^{\left(Q, \lambda\right)}_{\vect{\Theta}| \vect{Z} = \vect{z}}} = 0$.
\end{corollary}

 \subsection{Priors and Posteriors}
 
From Lemma~\ref{LemmaSensitivityEqual}, under the assumption that~$Q$ is a probability measure, it follows that
\begin{IEEEeqnarray}{rcl}
\nonumber
\mathsf{R}_{\vect{z}}\left( Q  \right) - \mathsf{R}_{\vect{z}}\left( P^{\left(Q, \lambda\right)}_{\vect{\Theta}| \vect{Z} = \vect{z}} \right)  \Tsupersqueezeequ
%\label{EqNiceEarlyMorning5432}
& = & 
\lambda \left( \KL{Q}{P^{\left(Q, \lambda\right)}_{\vect{\Theta}| \vect{Z} = \vect{z}}} + \KL{P^{\left(Q, \lambda\right)}_{\vect{\Theta}| \vect{Z} = \vect{z}}}{Q}  \right),\Tsupersqueezeequ 
\end{IEEEeqnarray}
where the right-hand side is a symmetrized Kullback-Leibler divergence, also known as Jeffreys divergence~\cite{jeffreys1946invariant}, between the measures~$Q$ and~$P^{\left(Q, \lambda\right)}_{\vect{\Theta}| \vect{Z} = \vect{z}}$.
More importantly, when~$Q$ is a probability measure, it follows from \cite[Theorem~$1$]{PerlazaTIT2024} that~$\KL{P^{\left(Q, \lambda\right)}_{\vect{\Theta}| \vect{Z} = \vect{z}}}{Q}\geqslant 0$ and~$\KL{Q}{P^{\left(Q, \lambda\right)}_{\vect{\Theta}| \vect{Z} = \vect{z}}}\geqslant0$, which reveals the fact that the expected empirical risk induced by the Gibbs probability measure~$P^{\left(Q, \lambda\right)}_{\vect{\Theta}| \vect{Z} = \vect{z}}$ is smaller than or equal to  the expected empirical risk induced by the reference measure~$Q$. This observation  is formalized by the following corollary of Lemma~\ref{LemmaSensitivityEqual}.

\begin{corollary}\label{CorSensitivityIneq}
If~$Q$ is a probability measure, then  the probability measure~$P^{\left(Q, \lambda\right)}_{\vect{\Theta}| \vect{Z} = \vect{z}}$ in~\eqref{EqGenpdf} satisfies
\begin{IEEEeqnarray}{l} 
 \mathsf{R}_{\vect{z}}\left( P^{\left(Q, \lambda\right)}_{\vect{\Theta}| \vect{Z} = \vect{z}} \right) \leqslant \mathsf{R}_{\vect{z}}\left( Q \right),
\end{IEEEeqnarray}
where the function~$\mathsf{R}_{\vect{z}}$ is defined in~\eqref{EqRxy}.
\end{corollary}
Note that the reference measure~$Q$ in Corollary~\ref{CorSensitivityIneq} can be chosen  independently of the dataset~$\vect{z}$. From this perspective, the measure~$Q$ can be interpreted as a prior on the models, while the probability measure~$P^{\left(Q, \lambda\right)}_{\vect{\Theta}| \vect{Z} = \vect{z}}$ can be interpreted as a posterior once the prior~$Q$ is confronted with the training dataset~$\vect{z}$.

\subsection{A Geometric Interpretation}

In Lemma~\ref{LemmaSensitivityEqual}, note that if~$Q$ is a probability measure,  then it holds from \cite[Theorem~$1$]{PerlazaTIT2024} that~$\KL{P}{Q} \geqslant 0$,~$\KL{P}{P^{\left(Q, \lambda\right)}_{\vect{\Theta}| \vect{Z} = \vect{z}}} \geqslant 0$, and~$ \KL{P^{\left(Q, \lambda\right)}_{\vect{\Theta}| \vect{Z} = \vect{z}}}{Q} \geqslant 0$. Thus, from~\eqref{EqDefSensitivity}, it holds that  
\begin{IEEEeqnarray}{rcl}
\nonumber
 & &\mathsf{G}\left( \vect{z}, P , P^{\left(Q, \lambda\right)}_{\vect{\Theta}| \vect{Z} = \vect{z}}\right)   +  \lambda \KL{P}{Q} \\
 \label{EqLasNochesQueNoLlegan}
  & = & \mathsf{R}_{\vect{z}}\left( P \right)   +  \lambda \KL{P}{Q} - \mathsf{R}_{\vect{z}}\left( P^{\left(Q, \lambda\right)}_{\vect{\Theta}| \vect{Z} = \vect{z}} \right)\\
\label{EqLasNochesQueSeVan}
& \geqslant & \mathsf{R}_{\vect{z}}\left( P^{\left(Q, \lambda\right)}_{\vect{\Theta}| \vect{Z} = \vect{z}}  \right) +  \lambda \KL{P^{\left(Q, \lambda\right)}_{\vect{\Theta}| \vect{Z} = \vect{z}}}{Q}   - \mathsf{R}_{\vect{z}}\left( P^{\left(Q, \lambda\right)}_{\vect{\Theta}| \vect{Z} = \vect{z}} \right) \IEEEeqnarraynumspace \\
\label{EqLasNochesLargas}
& = & \lambda \KL{P^{\left(Q, \lambda\right)}_{\vect{\Theta}| \vect{Z} = \vect{z}}}{Q},
\end{IEEEeqnarray}
and moreover,
\begin{IEEEeqnarray}{rcl}
\nonumber
&&\mathsf{G}\left( \vect{z}, P , P^{\left(Q, \lambda\right)}_{\vect{\Theta}| \vect{Z} = \vect{z}}\right)    +  \lambda \KL{P}{Q}  \\
\label{EqMordisco}
& = &  \lambda\left( \KL{P^{\left(Q, \lambda\right)}_{\vect{\Theta}| \vect{Z} = \vect{z}}}{Q} + \KL{P}{P^{\left(Q, \lambda\right)}_{\vect{\Theta}| \vect{Z} = \vect{z}}}  \right)\\
\label{EqMordiscoMontpellier}
& \geqslant &  \lambda  \KL{P}{P^{\left(Q, \lambda\right)}_{\vect{\Theta}| \vect{Z} = \vect{z}}},
\end{IEEEeqnarray}
where the inequality in~\eqref{EqLasNochesQueSeVan} follows from~\eqref{EqGenpdf}; and the equality in~\eqref{EqMordisco} follows from Lemma~\ref{LemmaSensitivityEqual}.

Hence, under the assumption that the measure~$Q$ is a probability measure, the equality in~\eqref{EqSensitivity} can be written as follows:
\begin{IEEEeqnarray}{rcl}
\nonumber
& & \left( \sqrt{\frac{1}{\lambda} \mathsf{G}\left( \vect{z}, P , P^{\left(Q, \lambda\right)}_{\vect{\Theta}| \vect{Z} = \vect{z}}\right)  + \KL{P}{Q} } \right)^{2} \\ 
\label{EqJun4at7h26in2024}
& = & \left(\sqrt{\KL{P^{\left(Q, \lambda\right)}_{\vect{\Theta}| \vect{Z} = \vect{z}}}{Q}} \right)^{2}  + \left( \sqrt{\KL{P}{P^{\left(Q, \lambda\right)}_{\vect{\Theta}| \vect{Z} = \vect{z}}} } \right)^{2}, \IEEEeqnarraynumspace
\end{IEEEeqnarray}
which together with~\eqref{EqMordisco} and~\eqref{EqMordiscoMontpellier} imply that a right-angled triangle  can be constructed such that the hypotenuse exhibits length~$\sqrt{\frac{1}{\lambda} \mathsf{G}\left( \vect{z}, P , P^{\left(Q, \lambda\right)}_{\vect{\Theta}| \vect{Z} = \vect{z}}\right)  + \KL{P}{Q} }$ and the short sides exhibit lengths  
$\sqrt{\KL{P^{\left(Q, \lambda\right)}_{\vect{\Theta}| \vect{Z} = \vect{z}}}{Q}}$ and~$\sqrt{\KL{P}{P^{\left(Q, \lambda\right)}_{\vect{\Theta}| \vect{Z} = \vect{z}}} }$, respectively.
Figure~\ref{FigInterpretation} shows this interpretation of the gap~$\mathsf{G}\left( \vect{z}, P , P^{\left(Q, \lambda\right)}_{\vect{\Theta}| \vect{Z} = \vect{z}}\right)$ in~\eqref{EqDefSensitivity}.  

\begin{figure}
\begin{center}
\begin{tikzpicture}
\draw[thick] (0,0) -- node[below] {$\supersqueezeequ\sqrt{\KL{P^{\left(Q, \lambda\right)}_{\vect{\Theta}| \vect{Z} = \vect{z}}}{Q}}$} ++(6,0);
\draw[thick] (0,0) -- node[left,rotate=90, above] {$\supersqueezeequ\sqrt{  \KL{P}{P^{\left(Q, \lambda\right)}_{\vect{\Theta}| \vect{Z} = \vect{z}}} }$} ++(0,2);
\draw[thick] (0,2) -- (6,0);
\node[left, rotate=-20, above] at (3,1) {$\sqrt{\frac{1}{\lambda}\mathsf{G}\left( \vect{z}, P , P^{\left(Q, \lambda\right)}_{\vect{\Theta}| \vect{Z} = \vect{z}}\right) + \KL{P}{Q} }$};
\end{tikzpicture}
\end{center}
\caption{Geometric interpretation of the gap~$\mathsf{G}\left( \vect{z}, P , P^{\left(Q, \lambda\right)}_{\vect{\Theta}| \vect{Z} = \vect{z}}\right)$ in~\eqref{EqDefSensitivity}.}
\label{FigInterpretation}
\end{figure} 

\subsection{Sensitivity from/to Arbitrary Deviations}\label{SecSensitivityArbitraryA}

Given two probability measures~$P_1$ and~$P_2$, both in~$\triangle\left( \set{M} \right)$,  and a dataset~$\vect{z} \in \left( \set{X} \times \set{Y} \right)^n$, the focus is on the variation of the functional~$\mathsf{R}_{\vect{z}}$ in~\eqref{EqRxy} from~$P_2$ to~$P_1$. In terms of the functional~$\mathsf{G}$ in~\eqref{EqGZeta}, such a variation is~$
\mathsf{G}\left( \vect{z}, P_1, P_2\right)$.
The following theorem provides an exact expression for such a variation.
\begin{theorem}\label{TheoremRGgeneral}
If the probability measures~$P_1$ and~$P_2$ are both absolutely continuous with respect to the~$\sigma$-finite measure~$Q$ in~\eqref{EqGenpdf}, then the gap~$\mathsf{G}\left( \vect{z}, P_1, P_2 \right)$ in~\eqref{EqGZeta} satisfies
\begin{IEEEeqnarray}{rcl}
\nonumber
\mathsf{G}\left( \vect{z}, P_1 , P_2 \right)  
& = &
\lambda\bigg( \KL{P_1}{P^{\left(Q, \lambda\right)}_{\vect{\Theta}| \vect{Z} = \vect{z}}}  - \KL{P_2}{P^{\left(Q, \lambda\right)}_{\vect{\Theta}| \vect{Z} = \vect{z}}}  
\\ 
\label{EqTheoremRGgeneral}
& & + \KL{P_2}{Q} - \KL{P_1}{Q} \bigg), \Tsupersqueezeequ\IEEEeqnarraynumspace
\end{IEEEeqnarray}
where the probability measure~$P^{\left(Q, \lambda\right)}_{\vect{\Theta}| \vect{Z} = \vect{z}}$ is defined in~\eqref{EqGenpdf}.
\end{theorem}
\begin{IEEEproof}
The proof follows from Lemma~\ref{LemmaSensitivityEqual} and by noticing that~$\mathsf{G}\left( \vect{z}, P_1 , P_2 \right) = \mathsf{G}\left( \vect{z}, P_1 , P^{\left(Q, \lambda\right)}_{\vect{\Theta}| \vect{Z} = \vect{z}} \right) - \mathsf{G}\left( \vect{z}, P_2 , P^{\left(Q, \lambda\right)}_{\vect{\Theta}| \vect{Z} = \vect{z}} \right)$.
\end{IEEEproof}

In Theorem~\ref{TheoremRGgeneral}, as long as the probability measures~$P_1$ and~$P_2$ are both absolutely continuous with respect to the~$\sigma$-finite measure~$Q$, the reference measure~$Q$ can be arbitrarily chosen.  Similarly, the parameter~$\lambda$ can be arbitrarily chosen within the set 
$\set{K}_{Q,\vect{z}}$ in~\eqref{EqSetKxy}. Some notions of optimality for the choice of~$\lambda$ have been studied in \cite{Bu-ISIT2024}.
This flexibility in the choice of these parameters is essentially because only the right-hand side of~\eqref{EqTheoremRGgeneral} depends on them. 
Another interesting observation is that none of the terms in the right-hand side of~\eqref{EqTheoremRGgeneral} depends simultaneously on both~$P_1$ and~$P_2$.  
These observations highlight the significant flexibility of the expression in~\eqref{EqTheoremRGgeneral} to construct closed-form expressions for the gap~$\mathsf{G}\left( \vect{z}, P_1 , P_2 \right)$. 

\subsection{Choice of the Reference Measure}\label{SecChoiceOfQ}

The choice of~$Q$ in~\eqref{EqTheoremRGgeneral} is subject to the constraint that both probability measures~$P_1$ and~$P_{2}$ must be absolutely continuous with respect to~$Q$.
From this perspective, the measure~$Q$ can be seen as a free parameter, which can be strategically chosen.

\subsubsection{Probability Measures}
Two choices of~$Q$, as a probability measure, for which the expression in the right-hand side of~\eqref{EqTheoremRGgeneral} significantly simplifies are~$Q = P_1$ and~$Q = P_2$, which leads to the following corollary of Theorem~\ref{TheoremRGgeneral}.

\begin{corollary}\label{CorollaryNovember19at20h29in2024Nice}
If~$P_1$ is absolutely continuous with respect to~$P_2$, then~$\mathsf{G}\left( \vect{z}, P_1 , P_2 \right)$ in~\eqref{EqTheoremRGgeneral} satisfies: 
\begin{IEEEeqnarray}{rcl}
\nonumber
&& \mathsf{G}\left( \vect{z}, P_1 , P_2 \right) \\
 & = & 
\lambda \bigg( \KL{P_1}{P^{\left(P_2, \lambda\right)}_{\vect{\Theta}| \vect{Z} = \vect{z}}} - \KL{P_1}{P_2}  
 - \KL{P_2}{P^{\left(P_2, \lambda \right)}_{\vect{\Theta}| \vect{Z} = \vect{z}}} \bigg).\squeezeequ \IEEEeqnarraynumspace
\end{IEEEeqnarray}
Alternatively, if~$P_2$ is absolutely continuous with respect to~$P_1$ then 
\begin{IEEEeqnarray}{rcl}
\nonumber
&& \mathsf{G}\left( \vect{z}, P_1 , P_2 \right)\\
& = & 
\lambda \bigg( \KL{P_1}{P^{\left(P_1, \lambda\right)}_{\vect{\Theta}| \vect{Z} = \vect{z}}} + \KL{P_2}{P_1}  
 - \KL{P_2}{P^{\left(P_1, \lambda \right)}_{\vect{\Theta}| \vect{Z} = \vect{z}}} \bigg),\squeezeequ  \IEEEeqnarraynumspace
\end{IEEEeqnarray}
where for all~$i \inCountTwo$, the probability measure~$P^{\left(P_i, \lambda\right)}_{\vect{\Theta}| \vect{Z} = \vect{z}}$ satisfies~\eqref{EqGenpdf} under the assumption that~$Q = P_i$.
\end{corollary}

In the case in which none of the measures is absolutely continuous with respect to its counterpart, choosing~$Q$ as a convex combination of~$P_1$ and~$P_2$ guarantees an explicit expression for~$\mathsf{G}\left( \vect{z}, P_1 , P_2 \right)$ in~\eqref{EqTheoremRGgeneral}. The following corollary formalizes this observation.
\begin{corollary}\label{CorRGspecificb}
For all~$\alpha \in \left(0, 1\right)$,~$\mathsf{G}\left( \vect{z}, P_1 , P_2 \right)$ in~\eqref{EqTheoremRGgeneral} satisfies
\begin{IEEEeqnarray}{rcl}
\nonumber
&& \mathsf{G}\left( \vect{z}, P_1 , P_2 \right)\\
\nonumber
 & = & 
\lambda \bigg( \KL{P_1}{P^{\left(\alpha P_1 + (1 - \alpha)P_2, \lambda\right)}_{\vect{\Theta}| \vect{Z} = \vect{z}}} 
 - \KL{P_2}{P^{\left(\alpha P_1 + (1 - \alpha)P_2, \lambda \right)}_{\vect{\Theta}| \vect{Z} = \vect{z}}}  \\
 & & + \KL{P_2}{\alpha P_1 + (1 - \alpha)P_2} - \KL{P_1}{\alpha P_1 + (1 - \alpha)P_2} \bigg),\middlesqueezeequ \IEEEeqnarraynumspace
\end{IEEEeqnarray}
where the probability measure~$P^{\left(\alpha P_1 + (1 - \alpha)P_2, \lambda\right)}_{\vect{\Theta}| \vect{Z} = \vect{z}}$ satisfies~\eqref{EqGenpdf} under the assumption that~$Q = \alpha P_1 + (1 - \alpha)P_2$.
\end{corollary}
 
The measure~$Q$ in~\eqref{EqGenpdf} can also be chosen as a~$\sigma$-finite measure, e.g., the counting measure or the Lebesgue measure.
\subsubsection{The Counting Measure}
Assume for  instance that the set of models~$\set{M}$ is countable and the reference measure~$Q$ is the counting measure.
Under this assumption, any probability measure~$P \in \triangle\left( \set{M} \right)$ is absolutely continuous with respect to the counting measure~$Q$. 
The Radon-Nikodym derivative of~$P$ with respect to~$Q$ is the probability mass function of~$P$, denoted by~$p: \set{M} \to [0,1]$. Interestingly, under the current assumption,~$\KL{P}{Q}$ is equal to negative Shannon's discrete entropy of~$P$. That is,~$\KL{P}{Q} = \sum_{\vect{\theta} \in \set{M}} p(\vect{\theta}) \log\left( p(\vect{\theta}) \right)\triangleq - H\left( P \right)$ .
This observation leads to the following corollary of Theorem~\ref{TheoremRGgeneral}.

\begin{corollary}\label{CorRGspecificc}
Let the set~$\set{M}$ be countable and let~$Q$ in~\eqref{EqGenpdf} be the corresponding counting measure. Then, the gap~$\mathsf{G}\left( \vect{z}, P_1 , P_2 \right)$ in~\eqref{EqGZeta} satisfies   
\begin{IEEEeqnarray}{rcl}
\label{EqTaconesRojos}
&& \mathsf{G}\left( \vect{z}, P_1 , P_2 \right) \\
\nonumber
 & =  & 
\lambda \bigg( \KL{P_1}{P^{\left(Q, \lambda\right)}_{\vect{\Theta}| \vect{Z} = \vect{z}}} 
 - \KL{P_2}{P^{\left(Q, \lambda \right)}_{\vect{\Theta}| \vect{Z} = \vect{z}}}    - H\left( P_2\right) + H\left( P_1 \right)  \bigg),\Dsupersqueezeequ %\IEEEeqnarraynumspace
\end{IEEEeqnarray}
where the probability measure~$P^{\left(Q, \lambda\right)}_{\vect{\Theta}| \vect{Z} = \vect{z}}$ is defined in~\eqref{EqGenpdf}; and the terms~$H\left( P_1\right)$ and~$H\left( P_2\right)$ represent Shannon's discrete entropy of the probability measures~$P_1$ and~$P_2$, respectively.
\end{corollary}
From \cite[Theorem~$3$]{PerlazaTIT2024}, it follows that the probability mass function of the measure~$P^{\left(Q, \lambda\right)}_{\vect{\Theta}| \vect{Z} = \vect{z}}$ in~\eqref{EqTaconesRojos}, denoted~$p_{\lambda, \vect{z}}: \set{M} \to [0,1]$,  satisfies 
\begin{IEEEeqnarray}{rcl}
p_{\lambda, \vect{z}} (\vect{\theta}) & = & \frac{\mathrm{d}P^{\left(Q, \lambda\right)}_{\vect{\Theta}| \vect{Z} = \vect{z}}}{\mathrm{d}Q} \left( \vect{\theta} \right)\\
& = & \frac{ \exp\left( - \frac{1}{\lambda}  \mathsf{L}\left(\vect{z}, \vect{\theta} \right) \right)}{\sum_{\vect{\nu} \in \set{M}} \exp\left( - \frac{1}{\lambda}  \mathsf{L}\left(\vect{z}, \vect{\nu} \right) \right) },
\end{IEEEeqnarray}
where the function~$\frac{\mathrm{d}P^{\left(Q, \lambda\right)}_{\vect{\Theta}| \vect{Z} = \vect{z}}}{\mathrm{d}Q}$ is in~\eqref{EqGenpdf}.

An important observation from Corollary~\ref{CorRGspecificc} is that  for all~$i \in \lbrace 1, 2 \rbrace$,  while the term~$\KL{P_i}{P^{\left(Q, \lambda\right)}_{\vect{\Theta}| \vect{Z} = \vect{z}}}$ is convex in~$P_i$, the term~$H\left( P_i \right)$ is concave in~$P_i$ \cite[Theorem~$2$]{PerlazaTIT2024}. Interestingly, the term~$H(P_i)$ is independent of the dataset~$\vect{z}$, the choice of the loss function~$\ell$ in~\eqref{EqEll}, and the regularization factor~$\lambda$. Moreover, while the term~$\KL{P_i}{P^{\left(Q, \lambda\right)}_{\vect{\Theta}| \vect{Z} = \vect{z}}}$ is minimized when~$P_i = P^{\left(Q, \lambda\right)}_{\vect{\Theta}| \vect{Z} = \vect{z}}$, the term~$H(P_i)$ is minimized when~$P_i$ is concentrated on any singleton of~$\set{M}$. This shows the trade-off arising in model selection between randomly choosing the models by sampling the probability measure~$P^{\left(Q, \lambda\right)}_{\vect{\Theta}| \vect{Z} = \vect{z}}$ and choosing a model with probability one.

\subsubsection{The Lebesgue Measure}
Assume that the set~$\set{M}$ is a subset of~$\reals^{d}$, and the reference measure~$Q$ is the corresponding Lebesgue measure. 
Under these assumptions, the Radon-Nikodym derivative of a given absolutely continuous probability measure~$P$ with respect to~$Q$ is the probability density function of~$P$, denoted by~$p: \set{M} \to [0,1]$. Under the current assumption,~$\KL{P}{Q}$ is equal to negative Shannon's differential entropy of~$P$. That is,~$\KL{P}{Q} = \int p(\vect{\theta}) \log\left( p(\vect{\theta}) \right) \mathrm{d} \vect{\theta} \triangleq - h(P)$.
This observation leads to the following corollary of Theorem~\ref{TheoremRGgeneral}.
\begin{corollary}\label{CorRGspecificD}
Let the set~$\set{M}$ be a subset of~$\reals^{d}$, and let~$Q$ in~\eqref{EqGenpdf} be the corresponding Lebesgue measure. Then, the gap~$\mathsf{G}\left( \vect{z}, P_1 , P_2 \right)$ in~\eqref{EqGZeta} satisfies   
\begin{IEEEeqnarray}{rcl}
\label{EqTaconesRojosYTanguitasNegras}
&& \mathsf{G}\left( \vect{z}, P_1 , P_2 \right) \\
\nonumber
 & =  & 
\lambda \bigg( \KL{P_1}{P^{\left(Q, \lambda\right)}_{\vect{\Theta}| \vect{Z} = \vect{z}}}  
 - \KL{P_2}{P^{\left(Q, \lambda \right)}_{\vect{\Theta}| \vect{Z} = \vect{z}}}    - h\left( P_2\right) + h\left( P_1 \right)  \bigg),\Dsupersqueezeequ %\IEEEeqnarraynumspace
\end{IEEEeqnarray}
where the probability measure~$P^{\left(Q, \lambda\right)}_{\vect{\Theta}| \vect{Z} = \vect{z}}$ is defined in~\eqref{EqGenpdf}; and the terms~$h\left( P_1\right)$ and~$h\left( P_2\right)$ represent Shannon's differential entropy of the probability measures~$P_1$ and~$P_2$, respectively.
\end{corollary}

From \cite[Theorem~$3$]{PerlazaTIT2024}, it follows that the probability density function of the measure~$P^{\left(Q, \lambda\right)}_{\vect{\Theta}| \vect{Z} = \vect{z}}$ in~\eqref{EqTaconesRojosYTanguitasNegras}, denoted~$p_{\lambda, \vect{z}}: \set{M} \to [0,1]$,  satisfies 
\begin{IEEEeqnarray}{rcl}
p_{\lambda, \vect{z}} (\vect{\theta}) & = & \frac{\mathrm{d}P^{\left(Q, \lambda\right)}_{\vect{\Theta}| \vect{Z} = \vect{z}}}{\mathrm{d}Q} \left( \vect{\theta} \right)\\
& = & \frac{ \exp\left( - \frac{1}{\lambda}  \mathsf{L}\left(\vect{z}, \vect{\theta} \right) \right)}{\int \exp\left( - \frac{1}{\lambda}  \mathsf{L}\left(\vect{z}, \vect{\nu} \right) \right)\mathrm{d} \vect{\nu} },
\end{IEEEeqnarray}
where the function~$\frac{\mathrm{d}P^{\left(Q, \lambda\right)}_{\vect{\Theta}| \vect{Z} = \vect{z}}}{\mathrm{d}Q}$ is in~\eqref{EqGenpdf}.
Despite the fact that for all~$i \in \lbrace 1,2 \rbrace$, the convexity of~$\KL{P_i}{P^{\left(Q, \lambda \right)}_{\vect{\Theta}| \vect{Z} = \vect{z}}}$ and the concavity of~$h\left( P_i\right)$ with respect to~$P_i$ hold as in the previous case, these terms do not lead to useful interpretations as, for instance, the differential entropy~$h\left( P_i\right)$ might be negative and unboundedly increase or decrease.

\section{Data-Driven Gaps}\label{SecDataDrivenGaps}
The purpose of this section is to provide explicit expressions for the data-driven gap~$\mathsf{G}\left( \vect{\theta}, Q_1, Q_2 \right)$ in~\eqref{EqGTheta} for an arbitrary model~$\vect{\theta} \in \set{M}$ and two arbitrary probability measures~$Q_1$ and~$Q_2$ in~$\triangle \left(\set{X}\times \set{Y}\right)$. Interestingly, these expressions are in terms of the WCDG probability measure~$P_{\hat{Z} | \vect{\Theta} = \vect{\theta}}^{\left( P_{S}, \beta \right)}~$ in~\eqref{Eqtheworstgeneral}.

\subsection{Deviations from the WCDG probability measure}
The variation or \emph{sensitivity} of the functional~$\mathsf{R}_{\vect{\theta}}$ in~\eqref{EqRModel} to deviations from the WCDG probability measure~$P_{\hat{Z} | \vect{\Theta} = \vect{\theta}}^{\left( P_{S}, \beta \right)}~$ in~\eqref{Eqtheworstgeneral} to an alternative probability measure~$P \in \triangle\left(\set{X} \times \set{Y} \right)$ is 
\begin{IEEEeqnarray}{rCl}
	\label{EqdefGfromtheworst}
\mathsf{G}\left(\vect{\theta}, P, P_{\hat{Z} | \vect{\Theta} = \vect{\theta}}^{\left( P_{S}, \beta \right)} \right) &=& \mathsf{R}_{\vect{\theta}} \left(P\right) -\mathsf{R}_{\vect{\theta}} \left(P_{\hat{Z} | \vect{\Theta} = \vect{\theta}}^{\left( P_{S}, \beta \right)} \right),
\end{IEEEeqnarray}where the functional~$\mathsf{G}$ is defined in~\eqref{EqGTheta}. The following lemma introduces an explicit expression for~$\mathsf{G}\left(\vect{\theta}, P, P_{\hat{Z} | \vect{\Theta} = \vect{\theta}}^{\left( P_{S}, \beta \right)} \right)$.

\begin{lemma}[{\cite[Theorem 6]{zouJSAIT2024}}]	\label{LemmaDeviatefromWorst}
The probability measure $P_{\hat{Z} | \vect{\Theta} = \vect{\theta}}^{\left( P_{S}, \beta \right)}~$ in ~\eqref{Eqtheworstgeneral} satisfies, for all~$P \in \Delta_{P_{S}} \left(\set{X}\times\set{Y}\right)$,
\begin{IEEEeqnarray}{rCl}
	\nonumber
	&&\mathsf{G}\left(\vect{\theta}, P, P_{\hat{Z} | \vect{\Theta} = \vect{\theta}}^{\left( P_{S}, \beta \right)} \right)\\
	\label{Eqdeviationworst}
	&&=\beta\left(\KL{P}{P_{S}} - \KL{P}{P_{\hat{Z} | \vect{\Theta} = \vect{\theta}}^{\left( P_{S}, \beta \right)} } -  \KL{P_{\hat{Z} | \vect{\Theta} = \vect{\theta}}^{\left( P_{S}, \beta \right)} }{P_{S}} \right)\squeezeequ,\spnum
\end{IEEEeqnarray}
where the functional~$\mathsf{G}$ is defined in~\eqref{EqGTheta}.
\end{lemma}
\subsection{Sensitivity from/to Arbitrary Deviations}
Given two probability measures~$P_1$ and~$P_2$, both in~$\triangle\left(\set{X}\times \set{Y}\right)$, and a model~$\vect{\theta}\in \set{M}$, the focus is on the variation of the function~$\mathsf{R}_{\vect{\theta}}$ in~\eqref{EqRModel} from~$P_2$ to~$P_1$. In terms of the functional~$\mathsf{G}$ in~\eqref{EqGTheta}, such a variation is~$\mathsf{G}\left(\vect{\theta},P_1,P_2\right)$. The following theorem provides an exact expression for such a variation.
\begin{theorem}[{\cite[Theorem~$8$]{zouJSAIT2024}\label{theoremgengapexpression}}]
	For all~$P_1 \in \Delta_{P_S} \left(\set{X}\times\set{Y}\right)$ and~$P_2 \in \Delta_{P_S} \left(\set{X}\times\set{Y}\right)$, and for all~$\vect{\theta}\in \set{M}$, 
	\begin{IEEEeqnarray}{rCl}
		\nonumber
		\mathsf{G}(\vect{\theta},P_1,P_2) 
		&=& \beta\Bigl(\KL{P_2}{P_{\hat{Z} | \vect{\Theta} = \vect{\theta}}^{\left( P_{S}, \beta \right)} }-\KL{P_1}{P_{\hat{Z} | \vect{\Theta} = \vect{\theta}}^{\left( P_{S}, \beta \right)} } \\
		\label{EqExpressionGenGap}
		&&- \KL{P_2}{P_S} + \KL{P_1}{P_S}\Bigl),
	\end{IEEEeqnarray}
	where the functional~$\mathsf{G}$ is defined in~\eqref{EqGTheta}; and the parameter~$\beta$, the model~$\vect{\theta}$, and the measures~$P_S$ and~$P_{\hat{Z} | \vect{\Theta} = \vect{\theta}}^{\left( P_{S}, \beta \right)}~$ jointly satisfy~\eqref{Eqtheworstgeneral}.
\end{theorem}

\subsection{Choice of the Reference Measure}
The choice of the reference measure~$P_S$ in~\eqref{EqExpressionGenGap} is subject to the constraint that both probability measures~$P_1$ and~$P_2$ must be absolutely continuous with respect to~$P_S$. Despite this constraint, the reference measure~$P_S$ can be seen as a free parameter, which can be strategically chosen.
Two choices of~$P_S$, for which the expression in the right-hand side of~\eqref{EqExpressionGenGap} significantly simplifies are~$P_S = P_1$ and~$P_S = P_2$, which leads to the following corollary of Theorem~\ref{theoremgengapexpression}.

\begin{corollary}[{\cite[Corollary 3]{zou2024Generalization}}]\label{CorollaryNovember10at20h29in2024Nices}
	If~$P_1$ is absolutely continuous with respect to~$P_2$, then the gap~$\mathsf{G}(\vect{\theta}, P_1, P_2)$ in~\eqref{EqGTheta} satisfies: 
	\begin{IEEEeqnarray}{rcl}
		\nonumber
		&& \mathsf{G}(\vect{\theta}, P_1, P_2) \\
		\label{EqTheKeyEquality2024}			
		&=&\beta\left(\KL{P_2}{P_{\hat{Z} | \vect{\Theta} = \vect{\theta}}^{\left( P_{2}, \beta \right)} }-\KL{P_1}{P_{\hat{Z} | \vect{\Theta} = \vect{\theta}}^{\left( P_{2}, \beta \right)} }   + \KL{P_1}{P_2} \right)\supersqueezeequ. \IEEEeqnarraynumspace
	\end{IEEEeqnarray}
	Alternatively, if~$P_2$ is absolutely continuous with respect to~$P_1$ then,  
	\begin{IEEEeqnarray}{rcl}
		\nonumber
		&&  \mathsf{G}(\vect{\theta}, P_1, P_2) \\ 	
		&=&\beta\left(\KL{P_2}{P_{\hat{Z} | \vect{\Theta} = \vect{\theta}}^{\left( P_{1}, \beta \right)} }-\KL{P_1}{P_{\hat{Z} | \vect{\Theta} = \vect{\theta}}^{\left( P_{1}, \beta \right)} } -  \KL{P_2}{P_1} \right)\supersqueezeequ, \IEEEeqnarraynumspace
	\end{IEEEeqnarray}
	where for all~$i \inCountTwo$, the probability measure~$P_{Z|\vect{\Theta} = \vect{\theta}}^{\left( P_i, \beta \right)}$ satisfies~\eqref{Eqtheworstgeneral} under the assumption that~$P_S = P_i$.
\end{corollary}

Interestingly, absolute continuity of~$P_1$ with respect to~$P_2$ or~$P_2$ with respect to~$P_1$ is not necessary for obtaining an expression for the gap~$\mathsf{G}(\vect{\theta}, P_1, P_2)$ in~\eqref{EqGTheta}. In the case in which none of the measures is absolutely continuous with respect to its counterpart,  choosing~$P_S$ as a convex combination of~$P_1$ and~$P_2$ always guarantees an explicit expression for the gap~$\mathsf{G}(\vect{\theta}, P_1, P_2)$.
The following corollary formalizes this observation.
\begin{corollary}
	For all~$P_1 \in \Delta_{P_S} \left(\set{X}\times\set{Y}\right)$ and~$P_2 \in \Delta_{P_S} \left(\set{X}\times\set{Y}\right)$, for all~$\vect{\theta}\in \set{M}$, and for all~$\alpha \in (0,1)$, 
	\begin{IEEEeqnarray}{rCl}
\nonumber
& &		\mathsf{G}(\vect{\theta},P_1,P_2)  \\
\nonumber
&=& \beta\Bigl(\KL{P_2}{P_{\hat{Z} | \vect{\Theta} = \vect{\theta}}^{\left( \alpha P_{1} + (1 - \alpha) P_{2}, \beta \right)} }-\KL{P_1}{P_{\hat{Z} | \vect{\Theta} = \vect{\theta}}^{\left( \alpha P_{1} + (1 - \alpha) P_{2}, \beta \right)} } \middlesqueezeequ\\
		&&- \KL{P_2}{\alpha P_{1} + (1 - \alpha) P_{2}} + \KL{P_1}{\alpha P_{1} + (1 - \alpha) P_{2}}\Bigl),\middlesqueezeequ \IEEEeqnarraynumspace
	\end{IEEEeqnarray}
	where the functional~$\mathsf{G}$ is defined in~\eqref{EqGTheta}; and the parameter~$\beta$, the model~$\vect{\theta}$, and the measures~$\alpha P_{1} + (1 - \alpha) P_{2}$ and~$P_{\hat{Z} | \vect{\Theta} = \vect{\theta}}^{\left( \alpha P_{1} + (1 - \alpha) P_{2}, \beta \right)}~$ jointly satisfy~\eqref{Eqtheworstgeneral}.
\end{corollary}

\subsection{A Geometric Interpretation}

\begin{figure}
	\begin{center}
		\begin{tikzpicture}
			\draw[thick] (0,0) -- node[below] {$\sqrt{\KL{P}{P_{\hat{Z} | \vect{\Theta} = \vect{\theta}}^{\left( P_{S}, \beta \right)} }}$} ++(6,0);
			\draw[thick] (0,0) -- node[left,rotate=90, above] {$\sqrt{\KL{P_{\hat{Z} | \vect{\Theta} = \vect{\theta}}^{\left( P_{S}, \beta \right)} }{P_S}}$} ++(0,2);
			\draw[thick] (0,2) -- (6,0);
			\node[left, rotate=-20, above] at (3,1) {$\sqrt{\KL{P}{P_S} - \frac{1}{\beta}\mathsf{G}\left(\vect{\theta},P,P_{\hat{Z} | \vect{\Theta} = \vect{\theta}}^{\left( P_{S}, \beta \right)} \right)}$};
		\end{tikzpicture}
	\end{center}
	\caption{Geometric interpretation of the gap~$\mathsf{G}\left(\vect{\theta},P, P_{\hat{Z} | \vect{\Theta} = \vect{\theta}}^{\left( P_{S}, \beta \right)} \right)$ in~\eqref{Eqdeviationworst}.}
	\label{FigInterpretation2}
\end{figure} 

In Lemma~\ref{LemmaDeviatefromWorst}, the reference measure~$P_S$ is a probability measure. Then, from \cite[Theorem~$1$]{PerlazaTIT2024}, it follows  that~$\KL{P}{P_S} \ge~0, \KL{P}{P_{\hat{Z} | \vect{\Theta} = \vect{\theta}}^{\left( P_{S}, \beta \right)} } \ge~0$ and~$\KL{P_{\hat{Z} | \vect{\Theta} = \vect{\theta}}^{\left( P_{S}, \beta \right)} }{P_S} \ge~0$. 
This observation together with~\eqref{Eqdeviationworst} lead to the following inequality:
\begin{IEEEeqnarray}{rCl}
	\nonumber
	0 &\leqslant &\KL{P}{P_{\hat{Z} | \vect{\Theta} = \vect{\theta}}^{\left( P_{S}, \beta \right)} } +  \KL{P_{\hat{Z} | \vect{\Theta} = \vect{\theta}}^{\left( P_{S}, \beta \right)} }{P_{S}} \\
	&&= \KL{P}{P_{S}} - \frac{1}{\beta}\mathsf{G}\left(\vect{\theta}, P, P_{\hat{Z} | \vect{\Theta} = \vect{\theta}}^{\left( P_{S}, \beta \right)} \right).
	 \squeezeequ \spnum
\end{IEEEeqnarray}
Moreover, 
\begin{IEEEeqnarray}{rCl}	
\nonumber
&& \beta \KL{P}{P_{S}} - \mathsf{G}\left(\vect{\theta}, P, P_{\hat{Z} | \vect{\Theta} = \vect{\theta}}^{\left( P_{S}, \beta \right)} \right)\\ 
\nonumber
&=& - \left(  \mathsf{R}_{\vect{\theta}}\left(P\right) -	\beta\KL{P}{P_S} \right)  + \mathsf{R}_{\vect{\theta}}\left(P_{\hat{Z} | \vect{\Theta} = \vect{\theta}}^{\left( P_{S}, \beta \right)} \right) \\ 
\nonumber 
&\geqslant& - \left(  \mathsf{R}_{\vect{\theta}}\left(P_{\hat{Z} | \vect{\Theta} = \vect{\theta}}^{\left( P_{S}, \beta \right)} \right) -	\beta\KL{P_{\hat{Z} | \vect{\Theta} = \vect{\theta}}^{\left( P_{S}, \beta \right)} }{P_S} \right)  \\
\label{EqOctober21at18h05in2024Agadir}
& &+ \mathsf{R}_{\vect{\theta}}\left(P_{\hat{Z} | \vect{\Theta} = \vect{\theta}}^{\left( P_{S}, \beta \right)} \right)\\
\label{EqOctober21at18h35in2024Agadir}
&=& \beta\KL{P_{\hat{Z} | \vect{\Theta} = \vect{\theta}}^{\left( P_{S}, \beta \right)} }{P_S},
\end{IEEEeqnarray}
where the equality in~\eqref{EqOctober21at18h05in2024Agadir} follows from \cite[Lemma~$3$]{zouJSAIT2024}.
Similarly,
\begin{IEEEeqnarray}{rCl}
\nonumber
&& \beta \KL{P}{P_{S}} - \mathsf{G}\left(\vect{\theta}, P, P_{\hat{Z} | \vect{\Theta} = \vect{\theta}}^{\left( P_{S}, \beta \right)} \right)\\ 
\label{EqOctober21at18h30in2024Agadir}
& = & \beta  \KL{P}{P_{\hat{Z} | \vect{\Theta} = \vect{\theta}}^{\left( P_{S}, \beta \right)} } +  \beta \KL{P_{\hat{Z} | \vect{\Theta} = \vect{\theta}}^{\left( P_{S}, \beta \right)} }{P_{S}}\\
\label{EqOctober21at18h36in2024Agadir}
& \geqslant&   \beta  \KL{P}{P_{\hat{Z} | \vect{\Theta} = \vect{\theta}}^{\left( P_{S}, \beta \right)} },
 \end{IEEEeqnarray}
where~\eqref{EqOctober21at18h30in2024Agadir} follows from Lemma~\ref{LemmaDeviatefromWorst}.

From these observations,  the equality in~\eqref{Eqdeviationworst} can be written as follows:
\begin{IEEEeqnarray}{rCl}
	\nonumber
&&{	\left( \sqrt{\KL{P}{P_S} - \frac{1}{\beta}\mathsf{G}\left(\vect{\theta},P,P_{\hat{Z} | \vect{\Theta} = \vect{\theta}}^{\left( P_{S}, \beta \right)} \right)}\right)}^2\\
\label{Eqtraingleequality}
&=& {\left(\sqrt{\KL{P}{P_{\hat{Z} | \vect{\Theta} = \vect{\theta}}^{\left( P_{S}, \beta \right)} }}\right)}^2 + {\left(\sqrt{\KL{P_{\hat{Z} | \vect{\Theta} = \vect{\theta}}^{\left( P_{S}, \beta \right)} }{P_S}}\right)}^2. \spnum
\end{IEEEeqnarray}
The equality in~\eqref{Eqtraingleequality} together with the inequalities in~\eqref{EqOctober21at18h35in2024Agadir} and~\eqref{EqOctober21at18h36in2024Agadir} imply that a right-angled triangle can be constructed such that the hypotenuse exhibits length~$\sqrt{\KL{P}{P_S} - \frac{1}{\beta}\mathsf{G}\left(\vect{\theta},P,P_{\hat{Z} | \vect{\Theta} = \vect{\theta}}^{\left( P_{S}, \beta \right)} \right)}$ and the short sides exhibit lengths~$\sqrt{\KL{P}{P_{\hat{Z} | \vect{\Theta} = \vect{\theta}}^{\left( P_{S}, \beta \right)} }}$ and~$\sqrt{\KL{P_{\hat{Z} | \vect{\Theta} = \vect{\theta}}^{\left( P_{S}, \beta \right)} }{P_S}}$, respectively. Figure~\ref{FigInterpretation2} shows this geometric interpretation.
   
\section{Conclusions and Final Remarks}

In this work, closed-form expressions for the generalization error of supervised machine learning algorithms have been derived in terms of information-theoretic measures. Two distinct classes of expressions are identified. The first class encompasses formulations involving the Gibbs algorithm, as defined by the Gibbs probability measure in Definition~\ref{DefGibbsAlgorithm}. The second class pertains to expressions incorporating the WCDG probability measure, defined in Definition~\ref{DefWCDG}.  
While the former represents a more general framework, as it imposes no specific conditions on the probability distribution of datasets, both classes offer unique and complementary insights. These insights establish connections between the generalization error and concepts such as hypothesis testing, information measures, and Euclidean geometry. However, these connections have been only briefly explored in this work, and their full potential to enhance the understanding of generalization error remains an open avenue for further research.  

It is important to note that the derived expressions do not significantly simplify the numerical computation of the generalization error. This is unsurprising, as the primary goal of this study is to provide a deeper conceptual understanding of the generalization error from an information-theoretic perspective. Specifically, the established connections between the generalization error, the Gibbs algorithm, and the WCDG probability measure represent a significant advancement in this goal.  
Concerning the connection to the Gibbs algorithm, the expression that best unveils the relevance of these findings is the one in Theorem~\ref{TheoremJuin26at11h03in2024}. Note that such expression implies that the generalization error of an arbitrary machine learning algorithm can be understood as a comparison, via relative entropy, of all possible instances of the algorithm under study with all possible instances of the Gibbs algorithm in two scenarios: $(a)$ Both algorithms are trained upon the same training datasets drawn from the same probability distribution; and $(b)$ Both algorithms are trained upon training datasets independently drawn from the same probability distribution. 
Alternatively, concerning the connection to the WCDG probability measure, the expression that best unveils the relevance of these findings is the one in Theorem~\ref{TheoremTrainEntropyGE}. The expression therein implies that the generalization error of a given machine learning algorithm consists of a comparison, via relative entropy, of all possible likelihoods on the data points induced by the algorithm under study; and the WCDG probability measure in two scenarios: $(a)$ both probability measures, e.g., likelihood and WCDG probability measures, are conditioned on the same model drawn from the same probability distribution; and $(b)$ both probability measures are respectively conditioned upon models independently  drawn from the same probability distribution. 

Another important finding, e.g., from Theorem~\ref{TheoremAugust11at11h02in2024}, is that the generalization error is the sum of two quantities: $(a)$ A first quantity that characterizes the dependence of the models on the training data via the mutual and lautum information; and $(b)$ a second quantity that characterizes a statistical distance between the algorithm under study and a Gibbs algorithm. 
While the first quantity $(a)$ is always nonnegative, the second quantity $(b)$ might be negative, zero, or positive. Clearly, a variation in any of these quantities, $(a)$ or $(b)$, implies a variation in the other, and thus, they cannot be separately optimized for algorithm design.  
Nonetheless, these quantities place Gibbs algorithms as natural references to evaluate the generalization capabilities of supervised machine learning algorithms.

\bibliographystyle{IEEEtran}
\bibliography{references}
 
\appendices

\section{Statistical Properties of Algorithms}\label{AppendixMiscellanea}

An algorithm~$P_{\vect{\Theta} | \vect{Z}} \in  \triangle\left(\set{M}| \left( \set{X} \times \set{Y} \right)^{n} \right)$ and a probability measure~$P_{\vect{Z}} \in \triangle\left( \left( \set{X} \times \set{Y} \right)^{n} \right)$ determine two unique probability measures in~$\triangle\left( \left(\set{X}\times\set{Y}\right)^n \times \set{M} \right)$ and~$\triangle\left( \set{M} \times \left(\set{X}\times\set{Y}\right)^n \right)$, respectively. Such probability measures are respectively denoted by~$P_{\vect{Z}, \vect{\Theta} }$ and~$P_{\vect{\Theta}, \vect{\vect{Z}}}$ and for all measurable sets~$\set{A} \subset \left(\set{X}\times\set{Y}\right)^n \times \set{M}$, it follows that
\begin{IEEEeqnarray}{rCl}
\label{EqMay20at14h23in2024}
P_{\vect{Z}, \vect{\Theta} }\left( \set{A} \right) & = & \int P_{\vect{\Theta} | \vect{Z} = \vect{z}} \left( \set{A}_{\vect{z}}\right) \mathrm{d} P_{\vect{Z}} \left( \vect{z}\right),
\end{IEEEeqnarray}
where~$\set{A}_{\vect{z}}$ is the section of the set~$\set{A}$ determined by~$\vect{z}$, namely, 
\begin{IEEEeqnarray}{rCl}
\label{EqNovember15at16h57in2024InTheBusToNice}
\set{A}_{\vect{z}} \triangleq \left\lbrace \vect{\theta} \in \set{M}: ( \vect{z},\vect{\theta}) \in \set{A} \right\rbrace.
\end{IEEEeqnarray}
Alternatively,  for all measurable sets~$\set{B} \subset \set{M} \times \left(\set{X}\times\set{Y}\right)^n$, it follows that
\begin{IEEEeqnarray}{rCl}
\label{EqOctober30at7h48in2024SophiaAntipolis}
P_{ \vect{\Theta}, \vect{Z} } \left( \set{B} \right) & = & \int P_{\vect{\Theta} | \vect{Z} = \vect{z}} \left( \set{B}_{\vect{z}}\right) \mathrm{d} P_{\vect{Z}} \left( \vect{z}\right),
\end{IEEEeqnarray}
where~$\set{B}_{\vect{z}}$ is the section of the set~$\set{B}$ determined by~$\vect{z}$, namely, 
\begin{IEEEeqnarray}{rCl}
\set{B}_{\vect{z}} \triangleq \left\lbrace \vect{\theta} \in \set{M}: ( \vect{\theta},\vect{z}) \in \set{B} \right\rbrace.
\end{IEEEeqnarray}

If the sets~$\set{A}$ and~$\set{B}$ are of the form~$\set{A} = \set{C} \times \set{D}$ and~$\set{B} = \set{D} \times \set{C}$, for some~$\set{C} \subseteq \left( \set{X} \times \set{Y} \right)^n$ and~$\set{D} \subseteq \set{M}$, then, it holds that 
\begin{IEEEeqnarray}{rCl}
\label{EqMay31at11h32in2024}
P_{\vect{Z},  \vect{\Theta}}\left( \set{C} \times \set{D} \right)  =  \int_{\set{C}} P_{\vect{\Theta} | \vect{Z} = \vect{z}} \left( \set{D} \right) \mathrm{d} P_{\vect{Z}} \left( \vect{z}\right) =
P_{\vect{\Theta},\vect{Z}}\left( \set{D} \times \set{C} \right). \squeezeequ \IEEEeqnarraynumspace
\end{IEEEeqnarray}
In particular, if~$\set{C} = \left( \set{X} \times \set{Y} \right)^n$ in~\eqref{EqMay31at11h32in2024}, then, it follows that 
\begin{IEEEeqnarray}{rCl}
\label{EqOctober30at8h30in2024SophiaAntipolis}
P_{\vect{Z}, \vect{\Theta} }\left( \left( \set{X} \times \set{Y} \right)^n \times \set{D} \right) =  P_{\vect{
\Theta}, \vect{Z} }\left( \set{D} \times \left( \set{X} \times \set{Y} \right)^n  \right)   =  P_{\vect{\Theta}} \left( \set{D} \right), \squeezeequ\IEEEeqnarraynumspace
\end{IEEEeqnarray}
where the probability measure~$P_{\vect{\Theta}} \in \triangle\left( \set{M} \right)$ also satisfies~\eqref{EqJune26at16h54in2024}.

Alternatively,  if~$\set{D} = \set{M}$  in~\eqref{EqMay31at11h32in2024}, then,  it holds that
\begin{IEEEeqnarray}{rCl}
\label{EqOctober30at8h29in2024SophiaAntipolis}
P_{\vect{Z}, \vect{\Theta} }\left( \set{C} \times \set{M} \right) = P_{\vect{\Theta},\vect{Z} }\left( \set{M} 
\times \set{C}\right)  = P_{\vect{Z}} \left( \set{C} \right).
\end{IEEEeqnarray}
Hence, in the following, the measures~$P_{\vect{\Theta}}$ in~\eqref{EqOctober30at8h30in2024SophiaAntipolis} and~$P_{\vect{Z}}$ in~\eqref{EqOctober30at8h29in2024SophiaAntipolis}  are said to be the marginals of the joint probability measures~$P_{\vect{Z}, \vect{\Theta}}$  in~\eqref{EqMay20at14h23in2024} and~$P_{\vect{\Theta},\vect{Z}}$  in~\eqref{EqOctober30at7h48in2024SophiaAntipolis}.

The instance~$P_{\vect{\Theta} | \vect{Z} = \vect{z}}$ of the algorithm~$P_{\vect{\Theta} | \vect{Z}}$, for some dataset~$\vect{z} \in \left( \set{X} \times \set{Y} \right)^n$, can be interpreted as the posterior probability measure of the prior~$P_{\vect{\Theta}}$ after the observation of the dataset~$\vect{z}$. 
Let the conditional probability measure~$P_{\vect{Z} | \vect{\Theta}} \in  \triangle\left( \left( \set{X} \times \set{Y} \right)^{n} | \set{M} \right)$ be such that for all measurable sets~$\set{C} \subset  \left( \set{X} \times \set{Y} \right)^{n}$,
\begin{IEEEeqnarray}{rCl}
\label{EqMay20at15h51in2024}
P_{\vect{Z}} \left( \set{C} \right) & = & \int P_{\vect{Z} | \vect{\Theta} = \vect{\theta}} \left( \set{C}  \right) \mathrm{d} P_{\vect{\Theta}} \left( \vect{\theta} \right),
\end{IEEEeqnarray}
with~$P_{\vect{\Theta}}$ in~\eqref{EqOctober30at8h30in2024SophiaAntipolis} and~$P_{\vect{Z}}$ in~
\eqref{EqOctober30at8h29in2024SophiaAntipolis}.
Such a conditional probability measure~$P_{\vect{Z} | \vect{\Theta}}$ exists and is unique almost surely with respect to~$P_{\vect{\Theta}}$ ~\cite[Theorem~$5.3.1$]{ash2000probability}.
The probability measure~$P_{\vect{Z} | \vect{\Theta}}$ can be interpreted as the likelihood (on the datasets) induced by the posterior~$P_{\vect{\Theta} | \vect{Z} }$ and the ground-truth probability distribution of the datasets~$P_{\vect{Z}}$.
The joint probability measures~$P_{\vect{Z}, \vect{\Theta}}$ and~$P_{\vect{\Theta},\vect{Z}}$ can be described via the conditional probability measure~$P_{\vect{\Theta} | \vect{Z}}$ and the probability measure~$P_{\vect{Z}}$ as in~\eqref{EqMay20at14h23in2024} and in~\eqref{EqOctober30at7h48in2024SophiaAntipolis}; or via the conditional probability measure~$P_{\vect{Z} | \vect{\Theta}}$ and the probability measure~$P_{\vect{\Theta}}$.
More specifically, for all measurable sets~$\set{A} \in \left(\set{X}\times\set{Y}\right)^n  \times \set{M}$, it follows that
\begin{IEEEeqnarray}{rCl}
\label{EqJun3at14h31in2024}
P_{\vect{Z}, \vect{\Theta} }\left( \set{A} \right) & = & \int P_{\vect{Z} | \vect{\Theta} = \vect{\theta}} \left( \set{A}_{\vect{\theta}}\right) \mathrm{d} P_{\vect{\Theta}} \left( \vect{\theta}\right),
\end{IEEEeqnarray}
where~$\set{A}_{\vect{\theta}}$ is the section of the set~$\set{A}$ determined by~$\vect{\theta}$, namely, 
\begin{IEEEeqnarray}{rCl}
\label{EqNovember15at16h39in2024InTheBusToNice}
\set{A}_{\vect{\theta}} = \left\lbrace \vect{z} \in \left( \set{X} \times \set{Y} \right)^n: (\vect{z},\vect{\theta}) \in \set{A} \right\rbrace.
\end{IEEEeqnarray}
Alternatively,  for all measurable sets~$\set{B} \subset \set{M} \times \left(\set{X}\times\set{Y}\right)^n$, it follows that
\begin{IEEEeqnarray}{rCl}
\label{EqNovember10at20h22in2024Nice}
P_{ \vect{\Theta}, \vect{Z} } \left( \set{B} \right) & = & \int P_{\vect{Z} | \vect{\Theta} = \vect{\theta}} \left( \set{B}_{\vect{\theta}}\right) \mathrm{d} P_{\vect{\Theta}} \left( \vect{\theta}\right),
\end{IEEEeqnarray}
where the section~$\set{B}_{\vect{\theta}}$ of the set~$\set{B}$ is 
\begin{IEEEeqnarray}{rCl}
\set{B}_{\vect{\theta}} = \left\lbrace  \vect{z} \in \left(\set{X} \times \set{Y} \right)^n:( \vect{\theta},\vect{z}) \in \set{B} \right\rbrace.
\end{IEEEeqnarray}
For all measurable sets~$\set{A} \in \left(\set{X}\times\set{Y}\right)^n  \times \set{M}$, let the set~$\hat{\set{A}} \in \set{M} \times \left(\set{X}\times\set{Y}\right)^n$ be
\begin{IEEEeqnarray}{rcl}
\label{EqNovember16at18h49in2024Nice}
\hat{\set{A}}  & = & \left\lbrace (\vect{\theta}, \vect{z}) \in \set{M} \times \left(\set{X}\times\set{Y}\right)^n:  (\vect{z} , \vect{\theta}) \in \set{A} \right\rbrace,
\end{IEEEeqnarray}
then, from~\eqref{EqJun3at14h31in2024} and~\eqref{EqNovember10at20h22in2024Nice}, it holds that
\begin{IEEEeqnarray}{rcl}
P_{\vect{Z}, \vect{\Theta} }\left( \set{A} \right) & = & P_{\vect{\Theta},\vect{Z}}\left( \hat{\set{A}} \right).
\end{IEEEeqnarray}

The following lemmas, Lemma~\ref{LemmaRNDsPart1} and Lemma~\ref{LemmaRNDsPart2}, present some properties of the probability measures mentioned above, which are known, but given their central roles in this work, are presented and proved using the current notation.

\begin{lemma}[Theorem~11 in \cite{InriaRR9599}]\label{LemmaRNDsPart1}
Consider the conditional probability measures~$P_{\vect{\Theta} | \vect{Z}}$ and~$P_{\vect{Z} | \vect{\Theta}}$;  and the probability measures~$P_{\vect{\Theta}}$ and~$P_{\vect{Z}}$ that jointly satisfy~\eqref{EqJune26at16h54in2024} and~\eqref{EqMay20at15h51in2024}.
 Assume that:
\begin{itemize}
\item[$(a)$] For all~$\vect{z} \in \left( \set{X} \times \set{Y}\right)^n$, the probability measure~$P_{\vect{\Theta} | \vect{Z} = \vect{z}}$ is absolutely continuous with respect to~$P_{\vect{\Theta}}$; and 
\item[$(b)$] For all~$\vect{\theta} \in \set{M}$, the probability measure~$P_{\vect{Z} | \vect{\Theta} = \vect{\theta}}$ is absolutely continuous with respect to~$P_{\vect{Z}}$.
\end{itemize}
Then, for all~$\left( \vect{z} , \vect{\theta} \right) \in \left( \set{X} \times \set{Y}\right)^n \times \set{M}$,  
\begin{IEEEeqnarray}{rCl}
\label{EqMay20at16h50in2024a}
 \frac{\mathrm{d} P_{\vect{Z} | \vect{\Theta} = \vect{\theta}}}{\mathrm{d} P_{\vect{Z}}} \left( \vect{z} \right)  & = & \frac{\mathrm{d} P_{\vect{\Theta} | \vect{Z} = \vect{z}}}{\mathrm{d} P_{\vect{\Theta}}}\left( \vect{\theta} \right),
\end{IEEEeqnarray}
almost surely with respect to the product measure~$P_{\vect{Z}}P_{\vect{\Theta}} \in \triangle\left( \left(\set{X} \times \set{Y}  \right)^n \times \set{M} \right)$.
\end{lemma}
\begin{lemma}[Theorem~12 in \cite{InriaRR9599}]\label{LemmaRNDsPart2}
Consider the conditional probability measures~$P_{\vect{\Theta} | \vect{Z}}$ and~$P_{\vect{Z} | \vect{\Theta}}$;  and the probability measures~$P_{\vect{\Theta}}$ and~$P_{\vect{Z}}$ that jointly satisfy~\eqref{EqJune26at16h54in2024} and~\eqref{EqMay20at15h51in2024}.
 Assume that:
\begin{itemize}
\item[$(a)$] For all~$\vect{z} \in \left( \set{X} \times \set{Y}\right)^n$, the probability measure~$P_{\vect{\Theta}}$ is absolutely continuous with respect to~$P_{\vect{\Theta} | \vect{Z} = \vect{z}}$; and 
\item[$(b)$] For all~$\vect{\theta} \in \set{M}$, the probability measure~$P_{\vect{Z}}$ 
 is absolutely continuous with respect to~$P_{\vect{Z} | \vect{\Theta} = \vect{\theta}}$. 
\end{itemize}
Then, for all~$\left( \vect{z} , \vect{\theta} \right) \in \left( \set{X} \times \set{Y}\right)^n \times \set{M}$,  
\begin{IEEEeqnarray}{rCl}
\label{EqMay20at16h50in2024}
\frac{\mathrm{d} P_{\vect{Z}}}{\mathrm{d} P_{\vect{Z} | \vect{\Theta} = \vect{\theta}}} \left( \vect{z} \right)
& =&  \frac{\mathrm{d} P_{\vect{\Theta}}}{\mathrm{d} P_{\vect{\Theta} | \vect{Z} = \vect{z}}}\left( \vect{\theta} \right),
\end{IEEEeqnarray}
almost surely with respect to the measure~$P_{\vect{Z} , \vect{\Theta}} \in \triangle\left( \left(\set{X} \times \set{Y}  \right)^n \times \set{M} \right)$ in~\eqref{EqMay20at14h23in2024}.
\end{lemma}
% %
The following lemmas, Lemma~\ref{EqNovember15at10h08in2024SophiaAntipolis} and Lemma~\ref{EqNovember13at17h04in2024InTheBusToNice}, introduce other properties for the particular case in which~$P_{\vect{Z}}$ satisfies~\eqref{EqSaturdayMai20at13h20in2024} for some~$P_{Z} \in \triangle\left( \set{X} \times \set{Y}\right)$. 

\begin{lemma}\label{EqNovember15at10h08in2024SophiaAntipolis}
Consider  the conditional probability measure~$P_{\vect{Z} | \vect{\Theta}}$ and the probability measure~$P_{\vect{Z}}$, both in~\eqref{EqMay20at15h51in2024}. Assume that:
\begin{itemize}
\item[$(a)$] The probability measure~$P_{\vect{Z}}$ satisfies~\eqref{EqSaturdayMai20at13h20in2024} for some~$P_{Z} \in \triangle\left( \set{X} \times \set{Y}\right)$; and
\item[$(b)$] For all~$\vect{\theta} \in \set{M}$, the probability measure~$P_{\vect{Z} | \vect{\Theta} = \vect{\theta}}$ is absolutely continuous with respect to~$P_{\vect{Z}}$.
\end{itemize}
Then, for all~$\vect{\theta} \in \set{M}$, the probability measure~$P_{Z | \vect{\Theta} = \vect{\theta}}$ in~\eqref{EqOctober17at14h40in2024} is absolutely continuous with respect to~$P_{Z}$.
\end{lemma}
\begin{IEEEproof}
For all measurable sets~$\set{A} \subset \left( \set{X} \times \set{Y} \right)^n$ and for all~$\vect{\theta} \in \set{M}$, 
\begin{IEEEeqnarray}{rcl}
\nonumber
& & P_{\vect{Z} | \vect{\Theta} = \vect{\theta}} \left( \set{A} \right)\\
\label{EqNovember14at14h52in2024Nice}
 &  = & \int_{\set{A}} \mathrm{d}P_{\vect{Z} | \vect{\Theta} = \vect{\theta}} \left( \vect{z} \right) \\
\label{EqNovember14at14h52in2024NiceA}
&  = & \int_{\set{A}} \frac{\mathrm{d}P_{\vect{Z} | \vect{\Theta} = \vect{\theta}}}{P_{\vect{Z}}} \left( \vect{z} \right) \mathrm{d}P_{\vect{Z}}\left( \vect{z} \right)  \\
\label{EqNovember14at14h52in2024NiceB}
&  = & \iint_{\set{A}_{z}} \frac{\mathrm{d}P_{\vect{Z} | \vect{\Theta} = \vect{\theta}}}{P_{\vect{Z}}} \left( z, \vect{\nu} \right) \mathrm{d}P_{\vect{Z}_{-1}}\left( \vect{\nu} \right) \mathrm{d}P_{Z}\left( z \right), \spnum
\end{IEEEeqnarray}
where the equality in~\eqref{EqNovember14at14h52in2024NiceA} follows from Assumption~$(b)$ and~\cite[Theorem 2]{InriaRR9599}; 
and the equality in~\eqref{EqNovember14at14h52in2024NiceB} follows from Assumption~$(a)$,
where the set~$\set{A}_{z}$ is the section of the set~$\set{A}$ determined by~$z \in \set{X} \times \set{Y}$:
\begin{IEEEeqnarray}{rcl}
\label{EqNovember15at09h17in2024SophiaAntipolis}
\set{A}_{z} = \left\lbrace \vect{\nu} \in \left( \set{X} \times \set{Y} \right)^{n-1} : \left( z, \vect{\nu} \right) \in \set{A} \right\rbrace,
\end{IEEEeqnarray}
and the measure~$P_{\vect{Z}_{-1}}$ is a product distribution in~$\triangle\left( \left( \set{X} \times \set{Y} \right)^{n-1} \right)$ formed by~$P_{Z}$. That is, the product measure~$P_{Z}  P_{\vect{Z}_{-1}}$ is identical to the measure~$P_{\vect{Z}}$ in~\eqref{EqSaturdayMai20at13h20in2024}.

Given a measurable set~$\set{B} \subset  \set{X} \times \set{Y}$, consider the set~$\set{A}$ that satisfies:
\begin{IEEEeqnarray}{rcl}
\label{EqNovember14at14h32in2024Nice}
\set{A} & = & \set{B} \times \set{A}^{(2)} \times \set{A}^{(3)} \times \cdots \times \set{A}^{(n)},  
\end{IEEEeqnarray}
where 
\begin{IEEEeqnarray}{rCl}
\set{A}^{(2)} = \set{A}^{(3)} = \cdots = \set{A}^{(n)} & = & \set{X} \times \set{Y}.
\end{IEEEeqnarray}
 Hence, for all~$z \in \set{B}$, it follows that the section of the set~$\set{A}$ in~\eqref{EqNovember14at14h32in2024Nice} determined by~$z$ is
\begin{IEEEeqnarray}{rcl}
\label{EqNovember14at15h00in2024NiceB}
\set{A}_{z} & = & \left( \set{X} \times \set{Y} \right)^{n-1} .
\end{IEEEeqnarray}
Using such a fact, it follows that for the sets~$\set{A}$ and~$\set{B}$ in~\eqref{EqNovember14at14h32in2024Nice}, and for all~$\vect{\theta} \in \set{M}$,
\begin{IEEEeqnarray}{rcl}
\label{EqNovember14at14h45in2024NiceA}
P_{Z | \vect{\Theta} = \vect{\theta}} \left( \set{B} \right) & = & P_{\vect{Z} | \vect{\Theta} = \vect{\theta}} \left( \set{A} \right)\\
\label{EqNovember14at14h45in2024NiceB}
& = &   \iint_{\set{A}_{z}} \frac{\mathrm{d}P_{\vect{Z} | \vect{\Theta} = \vect{\theta}}}{P_{\vect{Z}}} \left( z, \vect{\nu} \right) \mathrm{d}P_{\vect{Z}_{-1}}\left( \vect{\nu} \right) \mathrm{d}P_{Z}\left( z \right) \spnum\\
\label{EqNovember14at14h45in2024NiceC}
& = &   \iint \frac{\mathrm{d}P_{\vect{Z} | \vect{\Theta} = \vect{\theta}}}{P_{\vect{Z}}} \left( z, \vect{\nu} \right) \mathrm{d}P_{\vect{Z}_{-1}}\left( \vect{\nu} \right) \mathrm{d} P_{Z}\left( z \right), \spnum
\end{IEEEeqnarray}
where the equality in~\eqref{EqNovember14at14h45in2024NiceA} follows from~\eqref{EqNovember6at10h55in2024SophiaAntipolis} and the observation that under Assumption~$(a)$, Lemma~\ref{LemmaNovember5at17h06in2024SophiaAntipolis} holds;  
the equality in~\eqref{EqNovember14at14h45in2024NiceB} follows from~\eqref{EqNovember14at14h52in2024NiceB}; and 
the equality in~\eqref{EqNovember14at14h45in2024NiceC} follows from~\eqref{EqNovember14at15h00in2024NiceB}.

Hence, given that the set~$\set{B}$ can be any measurable subset of~$\set{X} \times \set{Y}$, the equality in~\eqref{EqNovember14at14h45in2024NiceC}, together with \cite[Theorem~$2.2.3$]{lehmann2005testing}, imply that the probability measure~$P_{Z | \vect{\Theta} = \vect{\theta}}$ is absolutely continuous with respect to~$P_{Z}$.
Moreover, the Radon-Nikodym derivative of~$P_{Z | \vect{\Theta} = \vect{\theta}}$ with respect to~$P_{Z}$, denoted by~$\frac{\mathrm{d}P_{Z | \vect{\Theta} = \vect{\theta}}}{\mathrm{d}P_{Z}}$, is
\begin{IEEEeqnarray}{rcl}
\frac{\mathrm{d}P_{Z | \vect{\Theta} = \vect{\theta}}}{\mathrm{d}P_{Z}} \left( z \right)  & = &  \int\frac{\mathrm{d}P_{\vect{Z} | \vect{\Theta} = \vect{\theta}}}{P_{\vect{Z}}} \left( z, \vect{\nu} \right) \mathrm{d}P_{\vect{Z}_{-1}}\left( \vect{\nu} \right), \spnum
\end{IEEEeqnarray}
which completes the proof.
\end{IEEEproof}
\begin{lemma}\label{EqNovember13at17h04in2024InTheBusToNice}
Consider the probability measure~$P_{\vect{Z}}$ and the conditional probability measure~$P_{\vect{Z} | \vect{\Theta}}$, both in~\eqref{EqMay20at15h51in2024}. Assume that:
\begin{itemize}
\item[$(a)$] The probability measure~$P_{\vect{Z}}$ satisfies~\eqref{EqSaturdayMai20at13h20in2024} for some~$P_{Z} \in \triangle\left( \set{X} \times \set{Y}\right)$; and
\item[$(b)$] For all~$\vect{\theta} \in \set{M}$, the probability measure~$P_{\vect{Z}}$ is absolutely continuous with respect to~$P_{\vect{Z} | \vect{\Theta} = \vect{\theta}}$.
\end{itemize}
Then, for all~$\vect{\theta} \in \set{M}$, the probability measure~$P_{Z}$ is absolutely continuous with respect to~$P_{Z | \vect{\Theta} = \vect{\theta}}$ in~\eqref{EqOctober17at14h40in2024}.
\end{lemma}
\begin{IEEEproof}
The proof follows along the lines of the proof of Lemma~\ref{EqNovember15at10h08in2024SophiaAntipolis}.
For all measurable sets~$\set{A} \subset \left( \set{X} \times \set{Y} \right)^n$ and for all~$\vect{\theta} \in \set{M}$, 
\begin{IEEEeqnarray}{rcl}
\nonumber
& & P_{\vect{Z}} \left( \set{A} \right) \\
\label{EqNovember15at08h53in2024SophiaAntipolis}
&  = & \int_{\set{A}} \mathrm{d}P_{\vect{Z}} \left( \vect{z} \right) \\
\label{EqNovember15at08h53in2024SophiaAntipolisB}
&  = & \int_{\set{A}} \frac{\mathrm{d}P_{\vect{Z}}}{\mathrm{d}P_{\vect{Z} | \vect{\Theta} = \vect{\theta}}} \left( \vect{z} \right) \mathrm{d}P_{\vect{Z} | \vect{\Theta} = \vect{\theta}} \left( \vect{z} \right)  \\
\label{EqNovember15at08h53in2024SophiaAntipolisC}
&  = & \iint_{\set{A}_z} \frac{\mathrm{d}P_{\vect{Z}}}{\mathrm{d}P_{\vect{Z} | \vect{\Theta} = \vect{\theta}}} \left( z, \vect{\nu} \right) \mathrm{d}P_{\vect{Z}_{-1} | Z_{1} = z, \vect{\Theta} = \vect{\theta}} \left( \vect{\nu} \right) \mathrm{d} P_{Z_{1} |  \vect{\Theta} = \vect{\theta}}\left( z \right) \middlesqueezeequ \spnum \\
\label{EqNovember15at08h53in2024SophiaAntipolisD}
&  = & \iint_{\set{A}_z} \frac{\mathrm{d}P_{\vect{Z}}}{\mathrm{d}P_{\vect{Z} | \vect{\Theta} = \vect{\theta}}} \left( z, \vect{\nu} \right) \mathrm{d}P_{\vect{Z}_{-1} | Z_{1} = z, \vect{\Theta} = \vect{\theta}} \left( \vect{\nu} \right) \mathrm{d} P_{Z |  \vect{\Theta} = \vect{\theta}}\left( z \right),\middlesqueezeequ
\end{IEEEeqnarray}
where the equality in~\eqref{EqNovember15at08h53in2024SophiaAntipolisB} follows from Assumption~$(b)$ and \cite[Theorem~$2.2.3$]{lehmann2005testing}; and
in the equality~\eqref{EqNovember15at08h53in2024SophiaAntipolisC}, the set~$\set{A}_{z}$ is defined in~\eqref{EqNovember15at09h17in2024SophiaAntipolis} and the probability measure~$P_{\vect{Z}_{-1} | Z_{1} = z, \vect{\Theta} = \vect{\theta}} \in \triangle\left( \left( \set{X} \times \set{Y} \right)^{n-1} \right)$ is the one that satisfies for all measurable subsets~$\set{C}$ of~$\left( \set{X} \times \set{Y} \right)^{n-1}$,
\begin{IEEEeqnarray}{rcl}
\nonumber
& & P_{\vect{Z}  | \vect{\Theta} = \vect{\theta}} \left( \left(\set{X} \times \set{Y} \right) \times \set{C} \right)\\
 & = & \int  P_{\vect{Z}_{-1} | Z_{1} = z, \vect{\Theta} = \vect{\theta}} \left(\set{C} \right) \mathrm{d} P_{Z_{1} |  \vect{\Theta} = \vect{\theta}}\left( z \right),
\end{IEEEeqnarray}
where the measure~$P_{Z_{1} |  \vect{\Theta} = \vect{\theta}}$ is the marginal defined in~\eqref{EqNovember6at10h55in2024SophiaAntipolis}; and finally, the equality~\eqref{EqNovember15at08h53in2024SophiaAntipolisD} follows from Assumption~$(a)$ and Lemma~\ref{LemmaNovember5at17h06in2024SophiaAntipolis}.

The proof continues by noticing that for a given measurable subset~$\set{B}$ of~$\set{X} \times \set{Y}$, the set~$\set{A}$ in~\eqref{EqNovember14at14h32in2024Nice} satisfies for all~$\vect{\theta} \in \set{M}$,
\begin{IEEEeqnarray}{rcl}
\label{EqNovember15at09h48in2024SophiaAntipolisA}
& & P_{Z} \left( \set{B} \right) =  P_{\vect{Z}} \left( \set{A} \right)\\
\label{EqNovember15at09h48in2024SophiaAntipolisB}
&  = & \iint_{\set{A}_z} \frac{\mathrm{d}P_{\vect{Z}}}{\mathrm{d}P_{\vect{Z} | \vect{\Theta} = \vect{\theta}}} \left( z, \vect{\nu} \right) \mathrm{d}P_{\vect{Z}_{-1} | Z_{1} = z, \vect{\Theta} = \vect{\theta}} \left( \vect{\nu} \right) \mathrm{d} P_{Z |  \vect{\Theta} = \vect{\theta}}\left( z \right),\supersqueezeequ \spnum\\
\label{EqNovember15at09h48in2024SophiaAntipolisC}
&  = & \iint  \frac{\mathrm{d}P_{\vect{Z}}}{\mathrm{d}P_{\vect{Z} | \vect{\Theta} = \vect{\theta}}} \left( z, \vect{\nu} \right) \mathrm{d}P_{\vect{Z}_{-1} | Z_{1} = z, \vect{\Theta} = \vect{\theta}} \left( \vect{\nu} \right) \mathrm{d} P_{Z |  \vect{\Theta} = \vect{\theta}}\left( z \right),\middlesqueezeequ \spnum
\end{IEEEeqnarray}
where the equality in~\eqref{EqNovember15at09h48in2024SophiaAntipolisA} follows from Assumption~$(a)$;
the equality in~\eqref{EqNovember15at09h48in2024SophiaAntipolisB} follows from~\eqref{EqNovember15at08h53in2024SophiaAntipolisD}; and
the equality in~\eqref{EqNovember15at09h48in2024SophiaAntipolisC} follows from~\eqref{EqNovember14at15h00in2024NiceB}.
Hence, given that the set~$\set{B}$ can be any measurable subset of~$\set{X} \times \set{Y}$, the equality in~\eqref{EqNovember15at09h48in2024SophiaAntipolisC}, together with \cite[Theorem~$2.2.3$]{lehmann2005testing}, imply that the probability measure~$P_{Z}$ is absolutely continuous with respect to~$P_{Z | \vect{\Theta} = \vect{\theta}}$.
Moreover, the Radon-Nikodym derivative of~$P_{Z}$ with respect to~$P_{Z | \vect{\Theta} = \vect{\theta}}$, denoted by~$\frac{\mathrm{d}P_{Z}}{\mathrm{d}P_{Z | \vect{\Theta} = \vect{\theta}}}$, is
\begin{IEEEeqnarray}{rcl}
\frac{\mathrm{d}P_{Z}}{\mathrm{d}P_{Z | \vect{\Theta} = \vect{\theta}}} \left( z \right)  & = &  \int  \frac{\mathrm{d}P_{\vect{Z}}}{\mathrm{d}P_{\vect{Z} | \vect{\Theta} = \vect{\theta}}} \left( z, \vect{\nu} \right) \mathrm{d}P_{\vect{Z}_{-1} | Z_{1} = z, \vect{\Theta} = \vect{\theta}} \left( \vect{\nu} \right), \squeezeequ \spnum
\end{IEEEeqnarray}
which completes the proof.
\end{IEEEproof}

Lemma~\ref{EqNovember15at10h08in2024SophiaAntipolis}  and Lemma~\ref{EqNovember13at17h04in2024InTheBusToNice} lead to the following corollary.

\begin{corollary}\label{CorollaryNovember15at10h09in2024SophiaAntipolis}
Consider the probability measure~$P_{\vect{Z}}$ and the conditional probability measure~$P_{\vect{Z} | \vect{\Theta}}$, both in~\eqref{EqMay20at15h51in2024}. Assume that:
\begin{itemize}
\item[$(a)$] The probability measure~$P_{\vect{Z}}$ satisfies~\eqref{EqSaturdayMai20at13h20in2024} for some~$P_{Z} \in \triangle\left( \set{X} \times \set{Y}\right)$; and
\item[$(b)$] For all~$\vect{\theta} \in \set{M}$, the probability measures~$P_{\vect{Z}}$ and~$P_{\vect{Z} | \vect{\Theta} = \vect{\theta}}$ are mutually absolutely continuous.
\end{itemize}
Then, for all~$\vect{\theta} \in \set{M}$, the probability measures~$P_{Z}$ and~$P_{Z | \vect{\Theta} = \vect{\theta}}$ in~\eqref{EqOctober17at14h40in2024} are mutually absolutely continuous.  
\end{corollary}

This appendix ends by stating the following property of the functionals~$\mathsf{R}_{\vect{\theta}}$ in~\eqref{EqRModel} and~$\mathsf{R}_{\vect{z}}$ in~\eqref{EqRxy}.

\begin{lemma}\label{LemmaJune4at11h05in2024}
Consider the conditional probability measures~$P_{\vect{\Theta} | \vect{Z}}$ and~$P_{\vect{Z} | \vect{\Theta}}$;  and the probability measures~$P_{\vect{\Theta}}$ and~$P_{\vect{Z}}$ that jointly satisfy~\eqref{EqJune26at16h54in2024} and~\eqref{EqMay20at15h51in2024}.
Then, the functionals~$\mathsf{R}_{\vect{\theta}}$ in~\eqref{EqRModel} and~$\mathsf{R}_{\vect{z}}$ in~\eqref{EqRxy}, satisfy 
\begin{IEEEeqnarray}{rcl}
\label{EqNovember8at8h23in2024NiceA}
\int \mathsf{R}_{\vect{z}} \left( P_{\vect{\Theta}} \right) \mathrm{d} P_{\vect{Z}} \left( \vect{z} \right) & = & \int \mathsf{R}_{\vect{\theta}} \left( P_{\vect{Z}} \right) \mathrm{d} P_{\vect{\Theta}} \left( \vect{\theta} \right).\IEEEeqnarraynumspace
\end{IEEEeqnarray} 
Consider also the following assumptions:
\begin{itemize}
\item[$(a)$] For all~$\vect{z} \in \left( \set{X} \times \set{Y}\right)^n$, the probability measure~$P_{\vect{\Theta} | \vect{Z} = \vect{z}}$ is absolutely continuous with respect to~$P_{\vect{\Theta}}$; and 
\item[$(b)$] For all~$\vect{\theta} \in \set{M}$, the probability measure~$P_{\vect{Z} | \vect{\Theta} = \vect{\theta}}$ is absolutely continuous with respect to~$P_{\vect{Z}}$.
\end{itemize}
Then, the functionals~$\mathsf{R}_{\vect{\theta}}$ and~$\mathsf{R}_{\vect{z}}$ also satisfy:
\begin{IEEEeqnarray}{rcl}
\label{EqNovember8at8h23in2024NiceB}
\int \mathsf{R}_{\vect{z}} \left( P_{\vect{\Theta} | \vect{Z} = \vect{z}} \right) \mathrm{d} P_{\vect{Z}} \left( \vect{z} \right) \squeezeequ& = &\squeezeequ \int \mathsf{R}_{\vect{\theta}} \left( P_{\vect{Z} | \vect{\Theta} = \vect{\theta}} \right) \mathrm{d} P_{\vect{\Theta}} \left( \vect{\theta} \right). \IEEEeqnarraynumspace
\end{IEEEeqnarray} 
\end{lemma}
\begin{IEEEproof}
The proof of~\eqref{EqNovember8at8h23in2024NiceA} follows from observing that
\begin{IEEEeqnarray}{rcl}
\label{EqNovember8at10h02in2024SophiaAntipolisA}
\int \mathsf{R}_{\vect{z}} \left( P_{\vect{\Theta}} \right) \mathrm{d} P_{\vect{Z}} \left( \vect{z} \right) 
& = & \iint \mathsf{L}\left( \vect{z}, \vect{\theta}\right) \mathrm{d} P_{\vect{\Theta}} \left( \vect{\theta} \right) \mathrm{d} P_{\vect{Z}} \left( \vect{z} \right) \spnum \\
\label{EqNovember8at10h02in2024SophiaAntipolisB}
& = & \iint \mathsf{L}\left( \vect{z}, \vect{\theta}\right) \mathrm{d} P_{\vect{Z}} \left( \vect{z} \right)  \mathrm{d} P_{\vect{\Theta}} \left( \vect{\theta} \right)\\
\label{EqNovember8at10h02in2024SophiaAntipolisC}
& = &\int \mathsf{R}_{\vect{\theta}} \left( P_{\vect{Z}} \right) \mathrm{d} P_{\vect{\Theta}} \left( \vect{\theta} \right).\IEEEeqnarraynumspace
\end{IEEEeqnarray}
where the equality in~\eqref{EqNovember8at10h02in2024SophiaAntipolisA} follows from~\eqref{EqRxy};
 the equality in~\eqref{EqNovember8at10h02in2024SophiaAntipolisB}  follows by exchanging the order of the integrals \cite[Theorem~$2.6.6$]{ash2000probability};
and finally, the equality in~\eqref{EqNovember8at10h02in2024SophiaAntipolisC} follows from~\eqref{EqRModel}.
This completes the proof of~\eqref{EqNovember8at8h23in2024NiceA}.

The proof of~\eqref{EqNovember8at8h23in2024NiceB} follows from observing that
\begin{IEEEeqnarray}{rcl}
\nonumber
& & \int \mathsf{R}_{\vect{z}} \left( P_{\vect{\Theta} | \vect{Z} = \vect{z}} \right) \mathrm{d} P_{\vect{Z}} \left( \vect{z} \right) \\
\label{EqNovember8at9h36in2024SophiaAntipolisA}
& = & \iint \mathsf{L}\left( \vect{z}, \vect{\theta}\right) \mathrm{d} P_{\vect{\Theta} | \vect{Z} = \vect{z}} \left( \vect{\theta} \right) \mathrm{d} P_{\vect{Z}} \left( \vect{z} \right) \\
\label{EqNovember8at9h36in2024SophiaAntipolisB}
& = & \iint \mathsf{L}\left( \vect{z}, \vect{\theta}\right) \frac{\mathrm{d} P_{\vect{\Theta} | \vect{Z} = \vect{z}}}{\mathrm{d}P_{\vect{\Theta}}} \left( \vect{\theta} \right) \mathrm{d}P_{\vect{\Theta}} \left( \vect{\theta} \right)  \mathrm{d}P_{\vect{Z}} \left( \vect{z} \right) \\
\label{EqNovember8at9h36in2024SophiaAntipolisC}
& = & \iint \mathsf{L}\left( \vect{z}, \vect{\theta}\right) \frac{\mathrm{d} P_{\vect{\Theta} | \vect{Z} = \vect{z}}}{\mathrm{d}P_{\vect{\Theta}}} \left( \vect{\theta} \right)  \mathrm{d}P_{\vect{Z}} \left( \vect{z} \right)  \mathrm{d}P_{\vect{\Theta}} \left( \vect{\theta} \right)\\
\label{EqNovember8at9h36in2024SophiaAntipolisD}
& = & \iint \mathsf{L}\left( \vect{z}, \vect{\theta}\right)  \frac{\mathrm{d} P_{\vect{Z} | \vect{\Theta} = \vect{\theta}}}{\mathrm{d}P_{\vect{Z}}} \left( \vect{z} \right) \mathrm{d}P_{\vect{Z}} \left( \vect{z} \right)  \mathrm{d}P_{\vect{\Theta}} \left( \vect{\theta} \right)\\
\label{EqNovember8at9h36in2024SophiaAntipolisE}
& = & \iint \mathsf{L}\left( \vect{z}, \vect{\theta}\right)  \mathrm{d}P_{\vect{Z} | \vect{\Theta} = \vect{\theta}} \left( \vect{z} \right)  \mathrm{d}P_{\vect{\Theta}} \left( \vect{\theta} \right)\\
\label{EqNovember8at9h36in2024SophiaAntipolisF}
& = & \int \mathsf{R}_{\vect{\theta}}\left( P_{\vect{Z} | \vect{\Theta} = \vect{\theta}} \right)  \mathrm{d}P_{\vect{\Theta}} \left( \vect{\theta} \right),
\end{IEEEeqnarray}
where the equality in~\eqref{EqNovember8at9h36in2024SophiaAntipolisA} follows from~\eqref{EqRxy};
the equality in~\eqref{EqNovember8at9h36in2024SophiaAntipolisB} follows from the assumption that~$P_{\vect{\Theta} | \vect{Z} = \vect{z}}$ is absolutely continuous with respect to~$P_{\vect{\Theta}}$ and \cite[Theorem~2]{InriaRR9591}; 
the equality in~\eqref{EqNovember8at9h36in2024SophiaAntipolisC} follows by exchanging the order of the integrals \cite[Theorem~$2.6.6$]{ash2000probability}; 
the equality in~\eqref{EqNovember8at9h36in2024SophiaAntipolisD}  follows from Lemma~\ref{LemmaRNDsPart1}; 
the equality in~\eqref{EqNovember8at9h36in2024SophiaAntipolisE}  follows from~\cite[Theorem~2]{InriaRR9591}; 
and finally, the equality in~\eqref{EqNovember8at9h36in2024SophiaAntipolisF} follows from~\eqref{EqRModel}. 
This completes the proof of~\eqref{EqNovember8at8h23in2024NiceB}.
\end{IEEEproof}
 
\section{Proof of Lemma~\ref{LemmaJun4at9h32in2024}}\label{AppProofOfLemmaJun4at9h32in2024}

The proof is essentially algebraic and follows from Definition~\ref{DefDEGG}. More specifically, 
\begin{IEEEeqnarray}{rCl}
\nonumber
& & \overline{\overline{\mathsf{G}}} \left(P_{\vect{\Theta} | \vect{Z}}, P_{\vect{Z}} \right) \\
\label{EqJun4at9h52in2024a}
& = & \iint  \left(  
\mathsf{R}_{\vect{u}} \left( P_{\vect{\Theta} | \vect{Z} = \vect{z}} \right) - 
\mathsf{R}_{\vect{z}}\left( P_{\vect{\Theta} | \vect{Z} = \vect{z}} \right)
 \right) \mathrm{d}P_{\vect{Z}} \left( \vect{u} \right) \mathrm{d}P_{\vect{Z}} \left( \vect{z} \right)\Dsupersqueezeequ\IEEEeqnarraynumspace \\
\nonumber
& = & \iint \int \mathsf{L} \left(\vect{u}, \vect{\theta} \right) \mathrm{d} P_{\vect{\Theta} | \vect{Z} = \vect{z}} \left( \vect{\theta} \right) \mathrm{d}P_{\vect{Z}} \left( \vect{u} \right) \mathrm{d}P_{\vect{Z}} \left( \vect{z} \right)\\ 
 \label{EqJun4at9h52in2024b}
 & & -  \int \mathsf{R}_{\vect{z}}\left( P_{\vect{\Theta} | \vect{Z} = \vect{z}} \right) \mathrm{d}P_{\vect{Z}} \left( \vect{z} \right), \Dsupersqueezeequ\IEEEeqnarraynumspace 
\end{IEEEeqnarray}
where the equality in~\eqref{EqJun4at9h52in2024b} follows from~\eqref{EqRxy}.
The proof follows by noticing that
\begin{IEEEeqnarray}{rCl}
\nonumber
&  & \iint \int \mathsf{L} \left(\vect{u}, \vect{\theta} \right) \mathrm{d} P_{\vect{\Theta} | \vect{Z} = \vect{z}} \left( \vect{\theta} \right) \mathrm{d}P_{\vect{Z}} \left( \vect{u} \right) \mathrm{d}P_{\vect{Z}} \left( \vect{z} \right)\\ 
 \label{EqJun4at10h12in2024a}
 & =  &  \iint \left( \int \mathsf{L} \left(\vect{u}, \vect{\theta} \right)  \mathrm{d}P_{\vect{Z}} \left( \vect{u} \right)  \right) \mathrm{d} P_{\vect{\Theta} | \vect{Z} = \vect{z}} \left( \vect{\theta} \right) \mathrm{d}P_{\vect{Z}} \left( \vect{z} \right) \IEEEeqnarraynumspace\\ 
  \label{EqJun4at10h12in2024aa}
 & =  &  \iint\mathsf{R}_{\vect{\theta}}\left(P_{\vect{Z}}  \right)\mathrm{d} P_{\vect{\Theta} | \vect{Z} = \vect{z}} \left( \vect{\theta} \right) \mathrm{d}P_{\vect{Z}} \left( \vect{z} \right) \IEEEeqnarraynumspace\\ 
 \label{EqJun4at10h12in2024b}
 & =  &  \iint\mathsf{R}_{\vect{\theta}}\left(P_{\vect{Z}}  \right)\frac{\mathrm{d} P_{\vect{\Theta} | \vect{Z} = \vect{z}}}{\mathrm{d}P_{\vect{\Theta}}} \left( \vect{\theta} \right) \mathrm{d}P_{\vect{\Theta}} \left( \vect{\theta} \right)  \mathrm{d}P_{\vect{Z}} \left( \vect{z} \right)  \IEEEeqnarraynumspace\\ 
 \label{EqJun4at10h12in2024c}
 & =  &  \int\mathsf{R}_{\vect{\theta}}\left(P_{\vect{Z}}  \right)\left( \int \frac{\mathrm{d} P_{\vect{\Theta} | \vect{Z} = \vect{z}}}{\mathrm{d}P_{\vect{\Theta}}} \left( \vect{\theta} \right)  \mathrm{d}P_{\vect{Z}} \left( \vect{z} \right) \right) \mathrm{d}P_{\vect{\Theta}} \left( \vect{\theta} \right)  \IEEEeqnarraynumspace\\ 
 \label{EqJun4at10h12in2024ddd}
 & =  &  \int \mathsf{R}_{\vect{\theta}}\left(P_{\vect{Z}}  \right) \mathrm{d}P_{\vect{\Theta}} \left( \vect{\theta} \right)  \IEEEeqnarraynumspace\\ 
 \label{EqJun4at10h12in2024e}
 & =  &  \int \mathsf{R}_{\vect{z}}  \left( P_{\vect{\Theta}} \right)  \mathrm{d}P_{\vect{Z}} \left( \vect{z} \right), 
\end{IEEEeqnarray}
where the equality in~\eqref{EqJun4at10h12in2024a} follows by exchanging the order of the integrals \cite[Theorem~$2.6.6$]{ash2000probability}; 
the equality in~\eqref{EqJun4at10h12in2024aa} follows from~\eqref{EqRModel}; 
the equality in~\eqref{EqJun4at10h12in2024b} follows from the assumption that~$P_{\vect{\Theta} | \vect{Z} = \vect{z}}$ is absolutely continuous with respect to~$P_{\vect{\Theta}}$ and \cite[Theorem~2]{InriaRR9591}; 
the equality in~\eqref{EqJun4at10h12in2024c} follows by exchanging the order of the integrals \cite[Theorem~$2.6.6$]{ash2000probability};
the equality in~\eqref{EqJun4at10h12in2024ddd} follows from \cite[Theorem~10]{InriaRR9591};
 and finally,  the equality in~\eqref{EqJun4at10h12in2024e} follows from Lemma~\ref{LemmaJune4at11h05in2024} (in Appendix \ref{AppendixMiscellanea}).

Plugging the equality~\eqref{EqJun4at10h12in2024e} in~\eqref{EqJun4at9h52in2024b} yields
\begin{IEEEeqnarray}{rCl}
\nonumber
& & \overline{\overline{\mathsf{G}}} \left(P_{\vect{\Theta} | \vect{Z}}, P_{\vect{Z}} \right) \\
\label{EqJun4at11h08in2024a}
& = &  \int \mathsf{R}_{\vect{z}}  \left( P_{\vect{\Theta}} \right)  \mathrm{d}P_{\vect{Z}} \left( \vect{z} \right) - 
\int \mathsf{R}_{\vect{z}}\left( P_{\vect{\Theta} | \vect{Z} = \vect{z}} \right) \mathrm{d}P_{\vect{Z}} \left( \vect{z} \right) \IEEEeqnarraynumspace \\
\label{EqJun4at11h08in2024b}
& = &  \int \left(  \mathsf{R}_{\vect{z}}  \left( P_{\vect{\Theta}} \right)  - 
\mathsf{R}_{\vect{z}}\left( P_{\vect{\Theta} | \vect{Z} = \vect{z}} \right) \right) \mathrm{d}P_{\vect{Z}} \left( \vect{z} \right) \IEEEeqnarraynumspace\\
\label{EqJun4at11h08in2024c}
& = &  \int \mathsf{G}\left( \vect{z}, P_{\vect{\Theta}} , P_{\vect{\Theta} | \vect{Z} = \vect{z}} \right)  \mathrm{d} P_{\vect{Z}} \left( \vect{z} \right),
\end{IEEEeqnarray}
which completes the proof.

\section{Proof of Lemma~\ref{LemmaOctober17at14h42in2024}}\label{AppProofOfLemmaOctober17at14h42in2024}

The proof is essentially algebraic and follows from Definition~\ref{DefDEGG}. More specifically, 
\begin{IEEEeqnarray}{rcl}
\nonumber
& & \overline{\overline{\mathsf{G}}} \left(P_{\vect{\Theta} | \vect{Z}}, P_{\vect{Z}} \right) \\
\label{EqNovember8at14h22in2024SophiaAntipolisA}
& = & \iint  \left(  
\mathsf{R}_{\vect{u}} \left( P_{\vect{\Theta} | \vect{Z} = \vect{z}} \right) - 
\mathsf{R}_{\vect{z}}\left( P_{\vect{\Theta} | \vect{Z} = \vect{z}} \right)
 \right) \mathrm{d}P_{\vect{Z}} \left( \vect{u} \right) \mathrm{d}P_{\vect{Z}} \left( \vect{z} \right) \squeezeequ \IEEEeqnarraynumspace\\
 \label{EqNovember8at14h22in2024SophiaAntipolisB} 
& = & \iint \int  \left(  
\mathsf{L} \left(\vect{u}, \vect{\theta}\right) - 
\mathsf{L}\left(\vect{z}, \vect{\theta}\right)
 \right)  \mathrm{d}P_{\vect{\Theta} | \vect{Z} = \vect{z}}  \left( \vect{\theta} \right)  \mathrm{d}P_{\vect{Z}} \left( \vect{u} \right) \mathrm{d}P_{\vect{Z}} \left( \vect{z} \right) \squeezeequ \IEEEeqnarraynumspace\\
 \nonumber
& = & \iint \left( \int 
\mathsf{L} \left(\vect{u}, \vect{\theta}\right)    \mathrm{d}P_{\vect{Z}} \left( \vect{u} \right) \right)   \mathrm{d}P_{\vect{\Theta} | \vect{Z} = \vect{z}}  \left( \vect{\theta} \right)\mathrm{d}P_{\vect{Z}} \left( \vect{z} \right) \\
\label{EqNovember8at14h22in2024SophiaAntipolisC}  
& & -  \iint   \mathsf{L}\left(\vect{z}, \vect{\theta} \right)  \mathrm{d}P_{\vect{\Theta} | \vect{Z} = \vect{z}}  \left( \vect{\theta} \right)   \mathrm{d}P_{\vect{Z}} \left( \vect{z} \right),
\squeezeequ \IEEEeqnarraynumspace\\
\label{EqNovember8at14h22in2024SophiaAntipolisD}  
& = & \iint \left( \mathsf{R}_{\vect{\theta}} \left( P_{\vect{Z}}\right) -  \mathsf{L}\left(\vect{z}, \vect{\theta} \right)  \right) \mathrm{d}P_{\vect{\Theta} | \vect{Z} = \vect{z}}  \left( \vect{\theta} \right)\mathrm{d}P_{\vect{Z}} \left( \vect{z} \right) ,
\end{IEEEeqnarray}
where the equality in~\eqref{EqNovember8at14h22in2024SophiaAntipolisB} follows from~\eqref{EqRxy}; 
the equality in~\eqref{EqNovember8at14h22in2024SophiaAntipolisC} follows by exchanging the order of the integrals \cite[Theorem~$2.6.6$]{ash2000probability}; and
the equality in~\eqref{EqNovember8at14h22in2024SophiaAntipolisD} follows from~\eqref{EqRModel}.  

The proof continues by noticing that: 
\begin{IEEEeqnarray}{rcl}
\nonumber
& & \iint  \mathsf{R}_{\vect{\theta}} \left( P_{\vect{Z}}\right)  \mathrm{d}P_{\vect{\Theta} | \vect{Z} = \vect{z}}  \left( \vect{\theta} \right)\mathrm{d}P_{\vect{Z}} \left( \vect{z} \right) 
\squeezeequ \IEEEeqnarraynumspace\\ 
\label{EqNovember8at15h23in2024SophiaAntipolisB}  
& = & \iint  \mathsf{R}_{\vect{\theta}} \left( P_{\vect{Z}}\right)  \frac{\mathrm{d}P_{\vect{\Theta} | \vect{Z} = \vect{z}}}{\mathrm{d} P_{\vect{\Theta}}}  \left( \vect{\theta} \right) \mathrm{d} P_{\vect{\Theta}} \left( \vect{\theta} \right) \mathrm{d}P_{\vect{Z}} \left( \vect{z} \right) 
\squeezeequ \IEEEeqnarraynumspace\\
\label{EqNovember8at15h23in2024SophiaAntipolisC}  
& = & \int   \mathsf{R}_{\vect{\theta}} \left( P_{\vect{Z}}\right)  \left( \int \frac{\mathrm{d}P_{\vect{\Theta} | \vect{Z} = \vect{z}}}{\mathrm{d} P_{\vect{\Theta}}}  \left( \vect{\theta} \right)  \mathrm{d}P_{\vect{Z}} \left( \vect{z} \right) \right)   \mathrm{d} P_{\vect{\Theta}} \left( \vect{\theta} \right) \squeezeequ \IEEEeqnarraynumspace \\
\label{EqNovember8at15h23in2024SophiaAntipolisD}  
& = & \int   \mathsf{R}_{\vect{\theta}} \left( P_{\vect{Z}}\right)      \mathrm{d} P_{\vect{\Theta}} \left( \vect{\theta} \right) \squeezeequ \IEEEeqnarraynumspace \\
\label{EqNovember8at15h23in2024SophiaAntipolisE}  
& = & \iint   \left( \frac{1}{n}\sum_{t =1}^{n} \ell\left( x_t, y_t, \vect{\theta}\right) \right) \mathrm{d} P_{\vect{Z}}\left( \vect{z} \right)  \mathrm{d} P_{\vect{\Theta}} \left( \vect{\theta} \right)\\
\label{EqNovember8at15h23in2024SophiaAntipolisF}  
& = & \frac{1}{n}\sum_{t =1}^{n}  \iint   \left( \ell\left( x, y, \vect{\theta}\right) \right) \mathrm{d} P_{Z}\left( x,y \right)  \mathrm{d} P_{\vect{\Theta}} \left( \vect{\theta} \right)\\
\label{EqNovember8at15h23in2024SophiaAntipolisG}  
& = & \iint   \left( \ell\left( x, y, \vect{\theta}\right) \right) \mathrm{d} P_{Z}\left( x,y \right)  \mathrm{d} P_{\vect{\Theta}} \left( \vect{\theta} \right),
\end{IEEEeqnarray}
where the equality~\eqref{EqNovember8at15h23in2024SophiaAntipolisB} follows from Assumption~$(b)$  and \cite[Theorem~2]{InriaRR9591};
 the equality in~\eqref{EqNovember8at15h23in2024SophiaAntipolisC}  follows by exchanging the order of the integrals \cite[Theorem~$2.6.6$]{ash2000probability};
 the equality in~\eqref{EqNovember8at15h23in2024SophiaAntipolisD} follows from \cite[Theorem~$10$]{InriaRR9591};
the equality in~\eqref{EqNovember8at15h23in2024SophiaAntipolisE} follows from~\eqref{EqLxy} and~\eqref{EqRModel}; and finally, the equality in~\eqref{EqNovember8at15h23in2024SophiaAntipolisF} follows from~\cite[Theorem~$1.6.3$]{ash2000probability}.

Finally, note that
\begin{IEEEeqnarray}{rcl}
\nonumber
& &  \iint \mathsf{L}\left(\vect{z}, \vect{\theta} \right) \mathrm{d}P_{\vect{\Theta} | \vect{Z} = \vect{z}}  \left( \vect{\theta} \right)\mathrm{d}P_{\vect{Z}} \left( \vect{z} \right) \\
\label{EqNovember10at13h45in2024NiceB}  
& = &  \iint \mathsf{L}\left(\vect{z}, \vect{\theta} \right)  \frac{\mathrm{d}P_{\vect{\Theta} | \vect{Z} = \vect{z}}}{\mathrm{d} P_{\vect{\Theta}}}  \left( \vect{\theta} \right) \mathrm{d} P_{\vect{\Theta}} \left( \vect{\theta} \right) \mathrm{d}P_{\vect{Z}} \left( \vect{z} \right) 
 \IEEEeqnarraynumspace\\
\label{EqNovember10at13h45in2024NiceC}  
& = &  \iint \mathsf{L}\left(\vect{z}, \vect{\theta} \right)  \frac{\mathrm{d}P_{\vect{\Theta} | \vect{Z} = \vect{z}}}{\mathrm{d} P_{\vect{\Theta}}}  \left( \vect{\theta} \right) \mathrm{d}P_{\vect{Z}} \left( \vect{z} \right)  \mathrm{d} P_{\vect{\Theta}} \left( \vect{\theta} \right)
 \IEEEeqnarraynumspace\\
\nonumber
& = &  \iint \mathsf{L}\left(\vect{z}, \vect{\theta} \right)  \frac{\mathrm{d}P_{\vect{\Theta} | \vect{Z} = \vect{z}}}{\mathrm{d} P_{\vect{\Theta}}}  \left( \vect{\theta} \right) \\
\label{EqNovember10at13h45in2024NiceD}  
& &\frac{\mathrm{d}P_{\vect{Z}}}{\mathrm{d} P_{\vect{Z} | \vect{\Theta} = \vect{\theta}}}  \left( \vect{z} \right) \mathrm{d}P_{\vect{Z} | \vect{\Theta} = \vect{\theta}}\left( \vect{z} \right)  \mathrm{d} P_{\vect{\Theta}} \left( \vect{\theta} \right)
 \IEEEeqnarraynumspace\\
\label{EqNovember10at13h45in2024NiceE}  
& = &  \iint \mathsf{L}\left(\vect{z}, \vect{\theta} \right)  \mathrm{d}P_{\vect{Z} | \vect{\Theta} = \vect{\theta}}\left( \vect{z} \right)  \mathrm{d} P_{\vect{\Theta}} \left( \vect{\theta} \right)
 \IEEEeqnarraynumspace\\
\label{EqNovember10at13h45in2024NiceF}  
& = &  \iint\left( \frac{1}{n}\sum_{t =1}^{n} \ell\left( x_t, y_t, \vect{\theta}\right) \right) \mathrm{d} P_{\vect{Z} | \vect{\Theta} = \vect{\theta}}\left( \vect{z} \right)  \mathrm{d} P_{\vect{\Theta}}\left( \vect{\theta}\right) \IEEEeqnarraynumspace\\
\label{EqNovember10at13h45in2024NiceG}  
& = &  \frac{1}{n}\sum_{t =1}^{n} \iint   \ell\left( x, y,  \vect{\theta} \right)  \mathrm{d} P_{Z_t | \vect{\Theta} = \vect{\theta}}\left( x,y \right)  \mathrm{d} P_{\vect{\Theta}}\left( \vect{\theta}\right)\\
\label{EqNovember10at13h45in2024NiceH}  
& = & \iint   \ell\left( x, y,  \vect{\theta} \right)  \mathrm{d} P_{Z | \vect{\Theta} = \vect{\theta}}\left( x,y \right)  \mathrm{d} P_{\vect{\Theta}}\left( \vect{\theta}\right),
\end{IEEEeqnarray}
where 
the equality~\eqref{EqNovember10at13h45in2024NiceB} follows from Assumption~$(b)$ and \cite[Theorem~2]{InriaRR9591}; 
the equality in~\eqref{EqNovember10at13h45in2024NiceC}  follows by exchanging the order of the integrals \cite[Theorem~$2.6.6$]{ash2000probability}; 
the equality~\eqref{EqNovember10at13h45in2024NiceD} follows from Assumption~$(c)$ and \cite[Theorem~2]{InriaRR9591}; 
the equality in~\eqref{EqNovember10at13h45in2024NiceE} follows from Lemma~\ref{LemmaRNDsPart1} and Lemma~\ref{LemmaRNDsPart2} (in Appendix~\ref{AppendixMiscellanea}); and \cite[Theorem~5]{InriaRR9591}, which imply that for all~$\left( \vect{z} , \vect{\theta} \right) \in \left( \set{X} \times \set{Y} \right)^n \times \set{M}$,
\begin{IEEEeqnarray}{rcl}
\frac{\mathrm{d}P_{\vect{\Theta}|\vect{Z} = \vect{z}}}{\mathrm{d}P_{\vect{\Theta}}}\left(\vect{\theta}\right)\frac{\mathrm{d}P_{\vect{Z}}}{\mathrm{d}P_{\vect{Z}|\vect{\Theta}=\vect{\theta}}}\left(\vect{z}\right)  & = & 1;
\end{IEEEeqnarray}
the equality in~\eqref{EqNovember10at13h45in2024NiceF} follows from~\eqref{EqLxy}; 
the equality in~\eqref{EqNovember10at13h45in2024NiceG} follows from~\cite[Theorem~$1.6.3$]{ash2000probability};
and finally, the equality in~\eqref{EqNovember10at13h45in2024NiceH} follows from Lemma~\ref{LemmaNovember5at17h06in2024SophiaAntipolis}, which holds under Assumption~$(a)$.

The proof ends by using~\eqref{EqNovember8at15h23in2024SophiaAntipolisG} and~\eqref{EqNovember10at13h45in2024NiceH} in~\eqref{EqNovember8at14h22in2024SophiaAntipolisD}, which yields
\begin{IEEEeqnarray}{rcl}
\nonumber
 \overline{\overline{\mathsf{G}}} \left(P_{\vect{\Theta} | \vect{Z}}, P_{\vect{Z}} \right)  
& = & \iint   \ell\left( x, y, \vect{\theta}\right)  \mathrm{d} P_{Z}\left( x,y \right) \mathrm{d} P_{\vect{\Theta}}\left( \vect{\theta}\right) \\
& & -  \iint  \ell\left( x, y,  \vect{\theta} \right)  \mathrm{d} P_{Z | \vect{\Theta} = \vect{\theta}}\left( x,y \right) \mathrm{d} P_{\vect{\Theta}}\left( \vect{\theta}\right) \squeezeequ\spnum\\
\label{EqNovember10at16h19in2024Nice}
& = & \int \mathsf{G} \left( \theta,  P_{Z} ,  P_{Z | \vect{\Theta} = \vect{\theta}} \right) \mathrm{d} P_{\vect{\Theta}}\left( \vect{\theta}\right),
\end{IEEEeqnarray}
where the equality in~\eqref{EqNovember10at16h19in2024Nice} follows from~\eqref{EqGTheta}.
This completes the proof.

\section{Proof of Theorem~\ref{TheoNovember29at13h20in2025HomeNice}}
\label{AppProofOfTheoNovember29at13h20in2025HomeNice}
The proof follows along the lines of the proof of \cite[Theorem~18]{InriaRR9599}.
From Theorem~\ref{TheoAugust9at11h15in2024}, it follows that the generalization error~$\overline{\overline{\mathsf{G}}} \left(P_{\vect{\Theta} | \vect{Z}}, P_{\vect{Z}} \right)$ in~\eqref{EqJun4at9h02in2024} satisfies:
\begin{IEEEeqnarray}{rCl}
\nonumber
& & \overline{\overline{\mathsf{G}}} \left(P_{\vect{\Theta} | \vect{Z}}, P_{\vect{Z}} \right) \\
\nonumber
& = & \lambda\int \left( \log \frac{\mathrm{d} P^{\left(Q, \lambda\right)}_{\vect{\Theta}| \vect{Z} = \vect{z}}}{\mathrm{d} Q}\left( \vect{\theta} \right)  \right) \mathrm{d}P_{\vect{\Theta} | \vect{Z} }P_{\vect{Z}} \left( \vect{\theta} , \vect{z} \right) \\
& & - \lambda \int\left( \log \frac{\mathrm{d} P^{\left(Q, \lambda\right)}_{\vect{\Theta}| \vect{Z} = \vect{z}}}{\mathrm{d} Q}\left( \vect{\theta} \right)  \right) \mathrm{d}P_{\vect{\Theta}} P_{\vect{Z}}\left( \vect{\theta}, \vect{z} \right)\IEEEeqnarraynumspace\\
\nonumber
& = & \lambda\int_{\set{A}_{\gamma}} \left( \log \frac{\mathrm{d} P^{\left(Q, \lambda\right)}_{\vect{\Theta}| \vect{Z} = \vect{z}}}{\mathrm{d} Q}\left( \vect{\theta} \right)  \right) \mathrm{d}P_{\vect{\Theta} | \vect{Z} }P_{\vect{Z}} \left( \vect{\theta} , \vect{z} \right) \\
\nonumber
& &+  \lambda\int_{\set{A}_{\gamma}^{\sfc}} \left( \log \frac{\mathrm{d} P^{\left(Q, \lambda\right)}_{\vect{\Theta}| \vect{Z} = \vect{z}}}{\mathrm{d} Q}\left( \vect{\theta} \right)  \right) \mathrm{d}P_{\vect{\Theta} | \vect{Z} }P_{\vect{Z}} \left( \vect{\theta} , \vect{z} \right) \\
\nonumber
& & - \lambda \int_{\set{A}_{\gamma}}\left( \log \frac{\mathrm{d} P^{\left(Q, \lambda\right)}_{\vect{\Theta}| \vect{Z} = \vect{z}}}{\mathrm{d} Q}\left( \vect{\theta} \right)  \right) \mathrm{d}P_{\vect{\Theta}} P_{\vect{Z}}\left( \vect{\theta}, \vect{z} \right)\IEEEeqnarraynumspace\\
\label{EqNovember29at21h05in2025HomeNiceA}
& & - \lambda \int_{\set{A}_{\gamma}^{\sfc}}\left( \log \frac{\mathrm{d} P^{\left(Q, \lambda\right)}_{\vect{\Theta}| \vect{Z} = \vect{z}}}{\mathrm{d} Q}\left( \vect{\theta} \right)  \right) \mathrm{d}P_{\vect{\Theta}} P_{\vect{Z}}\left( \vect{\theta}, \vect{z} \right)\IEEEeqnarraynumspace\\
& \geqslant  & \lambda \gamma P_{\vect{\Theta} | \vect{Z} }P_{\vect{Z}} \left( \set{A}_{\gamma} \right)  + \lambda \underline{\gamma} P_{\vect{\Theta} | \vect{Z} }P_{\vect{Z}} \left( \set{A}_{\gamma}^{\sfc} \right) \\
\label{EqNovember29at21h05in2025HomeNiceB}
& & - \lambda \bar{\gamma}  P_{\vect{\Theta}}P_{\vect{Z}} \left( \set{A}_{\gamma} \right) -\lambda
\gamma P_{\vect{\Theta}}P_{\vect{Z}} \left( \set{A}_{\gamma}^{\sfc} \right)  \\
\label{EqNovember29at21h05in2025HomeNiceC}
& =  & \lambda \left(  \underline{\gamma}  - \gamma \right) P_{\vect{\Theta} | \vect{Z} }P_{\vect{Z}} \left( \set{A}_{\gamma}^{\sfc} \right)  + \lambda \left( \gamma - \bar{\gamma} \right) P_{\vect{\Theta}  }P_{\vect{Z}} \left( \set{A}_{\gamma} \right),   \spnum
\end{IEEEeqnarray}
where the inequality in~\eqref{EqNovember29at21h05in2025HomeNiceB} follows from the definition of the set $\set{A}_{\gamma}$ in~\eqref{EqNovember20at13h44in2025HomeNice}; and the extrema in~\eqref{EqNovember30at13h35in2025HomeNiceA} and~\eqref{EqNovember30at13h35in2025HomeNiceB}. Finally, 
the equality in~\eqref{EqNovember29at21h05in2025HomeNiceC} follows from~\eqref{EqNovember30at13h38in2025HomeNiceA} and~\eqref{EqNovember30at13h38in2025HomeNiceB}.
This completes the proof of the lower bound. 

The proof of the upper bound follows a similar argument. From the equality in~\eqref{EqNovember29at21h05in2025HomeNiceA}, the following holds,
\begin{IEEEeqnarray}{rCl}
\nonumber
& & \overline{\overline{\mathsf{G}}} \left(P_{\vect{\Theta} | \vect{Z}}, P_{\vect{Z}} \right) \\
\nonumber
& \leqslant  & \lambda \bar{\gamma} P_{\vect{\Theta} | \vect{Z} }P_{\vect{Z}} \left( \set{A}_{\gamma} \right)  + \lambda \gamma P_{\vect{\Theta} | \vect{Z} }P_{\vect{Z}} \left( \set{A}_{\gamma}^{\sfc} \right) \\
\label{EqNovember30at13h51in2025HomeNiceA}
& & - \lambda \gamma  P_{\vect{\Theta}}P_{\vect{Z}} \left( \set{A}_{\gamma} \right) -\lambda
\underline{\gamma} P_{\vect{\Theta}}P_{\vect{Z}} \left( \set{A}_{\gamma}^{\sfc} \right)  \\
\label{EqNovember30at13h51in2025HomeNiceB}
& =  & \lambda \left(  \bar{\gamma} - \gamma  \right) P_{\vect{\Theta} | \vect{Z} }P_{\vect{Z}} \left( \set{A}_{\gamma} \right)  + \lambda \left( \gamma - \underline{\gamma} \right) P_{\vect{\Theta}  }P_{\vect{Z}} \left( \set{A}_{\gamma}^{\sfc} \right),   \spnum
\end{IEEEeqnarray}
where the inequality in~\eqref{EqNovember30at13h51in2025HomeNiceA} follows from the definition of the set $\set{A}_{\gamma}$ in~\eqref{EqNovember20at13h44in2025HomeNice}; and the extrema in~\eqref{EqNovember30at13h35in2025HomeNiceA} and~\eqref{EqNovember30at13h35in2025HomeNiceB}. Finally, 
the equality in~\eqref{EqNovember30at13h51in2025HomeNiceB} follows from~\eqref{EqNovember30at13h38in2025HomeNiceA} and~\eqref{EqNovember30at13h38in2025HomeNiceB}.
This completes the proof of the upper bound.

\section{Proof of Theorem~\ref{TheoDecember21at7h07in2025HomeNice}}\label{AppProofOfTheoDecember21at7h07in2025HomeNice}

The proof follows along the lines of the proof of \cite[Theorem~18]{InriaRR9599}.
From Theorem~\ref{TheoremAugust11at11h02in2024}, it follows that the difference $\overline{\overline{\mathsf{G}}} \left(P_{\vect{\Theta} | \vect{Z}}, P_{\vect{Z}} \right) -  \lambda \left( I\left( P_{\vect{\Theta}| \vect{Z}}; P_{\vect{Z}} \right) + L\left( P_{\vect{\Theta}| \vect{Z}}; P_{\vect{Z}} \right) \right)$ satisfies:
\begin{IEEEeqnarray}{rCl}
\nonumber
& & \overline{\overline{\mathsf{G}}} \left(P_{\vect{\Theta} | \vect{Z}}, P_{\vect{Z}} \right) -  \lambda \left( I\left( P_{\vect{\Theta}| \vect{Z}}; P_{\vect{Z}} \right) + L\left( P_{\vect{\Theta}| \vect{Z}}; P_{\vect{Z}} \right) \right) \\
& = & \lambda\int \left( \log \frac{\mathrm{d} P^{\left(Q, \lambda\right)}_{\vect{\Theta}| \vect{Z} = \vect{z}}}{\mathrm{d} P_{\vect{\Theta}| \vect{Z} = \vect{z}}}\left( \vect{\theta} \right)  \right) \mathrm{d}P_{\vect{\Theta} | \vect{Z} }P_{\vect{Z}} \left( \vect{\theta} , \vect{z} \right) \\
& & - \lambda \int\left( \log \frac{\mathrm{d} P^{\left(Q, \lambda\right)}_{\vect{\Theta}| \vect{Z} = \vect{z}}}{\mathrm{d} P_{\vect{\Theta}| \vect{Z} = \vect{z}}}\left( \vect{\theta} \right)  \right) \mathrm{d}P_{\vect{\Theta}} P_{\vect{Z}}\left( \vect{\theta}, \vect{z} \right)\IEEEeqnarraynumspace\\
\nonumber
& = & \lambda\int_{\set{B}_{\gamma}} \left( \log \frac{\mathrm{d} P^{\left(Q, \lambda\right)}_{\vect{\Theta}| \vect{Z} = \vect{z}}}{\mathrm{d} P_{\vect{\Theta}| \vect{Z} = \vect{z}}}\left( \vect{\theta} \right)  \right) \mathrm{d}P_{\vect{\Theta} | \vect{Z} }P_{\vect{Z}} \left( \vect{\theta} , \vect{z} \right) \\
\nonumber
& &+  \lambda\int_{\set{B}_{\gamma}^{\sfc}} \left( \log \frac{\mathrm{d} P^{\left(Q, \lambda\right)}_{\vect{\Theta}| \vect{Z} = \vect{z}}}{\mathrm{d} P_{\vect{\Theta}| \vect{Z} = \vect{z}}}\left( \vect{\theta} \right)  \right) \mathrm{d}P_{\vect{\Theta} | \vect{Z} }P_{\vect{Z}} \left( \vect{\theta} , \vect{z} \right) \\
\nonumber
& & - \lambda \int_{\set{B}_{\gamma}}\left( \log \frac{\mathrm{d} P^{\left(Q, \lambda\right)}_{\vect{\Theta}| \vect{Z} = \vect{z}}}{\mathrm{d} P_{\vect{\Theta}| \vect{Z} = \vect{z}}}\left( \vect{\theta} \right)  \right) \mathrm{d}P_{\vect{\Theta}} P_{\vect{Z}}\left( \vect{\theta}, \vect{z} \right)\IEEEeqnarraynumspace\\
\label{EqDecember21at8h28in2025HomeNiceA}
& & - \lambda \int_{\set{B}_{\gamma}^{\sfc}}\left( \log \frac{\mathrm{d} P^{\left(Q, \lambda\right)}_{\vect{\Theta}| \vect{Z} = \vect{z}}}{\mathrm{d} P_{\vect{\Theta}| \vect{Z} = \vect{z}}}\left( \vect{\theta} \right)  \right) \mathrm{d}P_{\vect{\Theta}} P_{\vect{Z}}\left( \vect{\theta}, \vect{z} \right)\IEEEeqnarraynumspace\\
& \geqslant  & \lambda \gamma P_{\vect{\Theta} | \vect{Z} }P_{\vect{Z}} \left( \set{B}_{\gamma} \right)  + \lambda \underline{\gamma} P_{\vect{\Theta} | \vect{Z} }P_{\vect{Z}} \left( \set{B}_{\gamma}^{\sfc} \right) \\
\label{EqDecember21at8h28in2025HomeNiceB}
& & - \lambda \bar{\gamma}  P_{\vect{\Theta}}P_{\vect{Z}} \left( \set{B}_{\gamma} \right) -\lambda
\gamma P_{\vect{\Theta}}P_{\vect{Z}} \left( \set{B}_{\gamma}^{\sfc} \right)  \\
\label{EqDecember21at8h28in2025HomeNiceC}
& =  & \lambda \left(  \underline{\gamma}  - \gamma \right) P_{\vect{\Theta} | \vect{Z} }P_{\vect{Z}} \left( \set{B}_{\gamma}^{\sfc} \right)  + \lambda \left( \gamma - \bar{\gamma} \right) P_{\vect{\Theta}  }P_{\vect{Z}} \left( \set{B}_{\gamma} \right),   \spnum
\end{IEEEeqnarray}
where the inequality in~\eqref{EqDecember21at8h28in2025HomeNiceB} follows from the definition of the set $\set{B}_{\gamma}$ in~\eqref{EqDecember20at21h49in2025HomeNice}; and the extrema in~\eqref{EqDecember20at22h15in2025HomeNiceA} and~\eqref{EqDecember20at22h15in2025HomeNiceB}. Finally, 
the equality in~\eqref{EqDecember21at8h28in2025HomeNiceC} follows from~\eqref{EqDecember20at21h55in2025HomeNiceA} and~\eqref{EqDecember20at21h55in2025HomeNiceB}.
This completes the proof of the lower bound. 
The proof of the upper bound follows a similar argument. From the equality in~\eqref{EqDecember21at8h28in2025HomeNiceA}, the following holds,
\begin{IEEEeqnarray}{rCl}
\nonumber
& & \overline{\overline{\mathsf{G}}} \left(P_{\vect{\Theta} | \vect{Z}}, P_{\vect{Z}} \right) \\
\nonumber
& \leqslant  & \lambda \bar{\gamma} P_{\vect{\Theta} | \vect{Z} }P_{\vect{Z}} \left( \set{B}_{\gamma} \right)  + \lambda \gamma P_{\vect{\Theta} | \vect{Z} }P_{\vect{Z}} \left( \set{B}_{\gamma}^{\sfc} \right) \\
\label{EqNovember30at13h51in2025HomeNiceC}
& & - \lambda \gamma  P_{\vect{\Theta}}P_{\vect{Z}} \left( \set{B}_{\gamma} \right) -\lambda
\underline{\gamma} P_{\vect{\Theta}}P_{\vect{Z}} \left( \set{B}_{\gamma}^{\sfc} \right)  \\
\label{EqNovember30at13h51in2025HomeNiceD}
& =  & \lambda \left(  \bar{\gamma} - \gamma  \right) P_{\vect{\Theta} | \vect{Z} }P_{\vect{Z}} \left( \set{B}_{\gamma} \right)  + \lambda \left( \gamma - \underline{\gamma} \right) P_{\vect{\Theta}  }P_{\vect{Z}} \left( \set{B}_{\gamma}^{\sfc} \right),   \spnum
\end{IEEEeqnarray}
where the inequality in~\eqref{EqNovember30at13h51in2025HomeNiceC} follows from the definition of the set $\set{B}_{\gamma}$ in~\eqref{EqDecember20at21h49in2025HomeNice}; and the extrema in~\eqref{EqDecember20at22h15in2025HomeNiceA} and~\eqref{EqDecember20at22h15in2025HomeNiceB}. Finally, 
the equality in~\eqref{EqNovember30at13h51in2025HomeNiceD} follows from~\eqref{EqDecember20at21h55in2025HomeNiceA} and~\eqref{EqDecember20at21h55in2025HomeNiceB}.
This completes the proof of the upper bound.

\end{document}